\documentclass[a4paper,fleqn]{cas-dc}

\usepackage[authoryear,longnamesfirst]{natbib}
\usepackage{array}
\usepackage{amsmath}
\usepackage{float}
\usepackage{caption}
\def\tsc#1{\csdef{#1}{\textsc{\lowercase{#1}}\xspace}}
\tsc{WGM}
\tsc{QE}

\begin{document}
\let\WriteBookmarks\relax
\def\floatpagepagefraction{1}
\def\textpagefraction{.001}

\shorttitle{HierarchicalDAEW for Spatial Gene Expression Prediction}    
\shortauthors{Chattopadhyay et al.}  

\title[mode = title]{HierarchicalDAEW: Domain-Aware Edge-Weighted Graph Convolution with Evidential Uncertainty for Multi-Section Spatial Gene Expression Prediction from H\&E Histology}  

\tnotemark[1] 
\tnotetext[1]{} 
\author[1]{Kritanu Chattopadhyay}[orcid=0009-0008-1912-6531]
\cormark[1]
\fnmark[1]
\ead{kritanuchattopahdyaywork@gmail.com}
\ead{kc.23me8010@nitdgp.ac.in}
\ead[url]{}
\credit{Conceptualization, Methodology, Software, Formal analysis, 
Investigation, Data curation, Visualization, Writing -- original draft}
\affiliation[1]{organization={Department of Mechanical Engineering, 
            National Institute of Technology Durgapur},
            addressline={},
            city={Durgapur},
            postcode={713209},
            state={West Bengal},
            country={India}}

\author[5]{Soumya Chatterjee}[orcid=0000-0002-4162-4472]
\fnmark[2]
\ead{chapeshwar@gmail.com}
\ead[url]{}
\credit{Methodology, Validation, Writing -- review and editing}
\affiliation[5]{organization={Department of Electrical Engineering, 
            National Institute of Technology Durgapur},
            addressline={},
            city={Durgapur},
            postcode={713209},
            state={West Bengal},
            country={India}}

\author[2,4]{Ondrej Krejcar}[orcid=0000-0002-5992-2574]
\fnmark[3]
\ead{ondrej.krejcar@uhk.cz}
\ead[url]{}
\credit{Supervision, Funding acquisition, Writing -- review and editing}
\affiliation[2]{organization={Research Center, Skoda Auto University},
            addressline={Na Karmeli 1457},
            city={Mlada Boleslav},
            postcode={293 01},
            state={},
            country={Czech Republic}}
\affiliation[4]{organization={Center for Basic and Applied Research, 
            Faculty of Informatics and Management, 
            University of Hradec Kralove},
            addressline={Rokitanskeho 62},
            city={Hradec Kralove},
            postcode={500 03},
            state={},
            country={Czech Republic}}

\author[3,4]{Debotosh Bhattacharjee}[orcid=0000-0002-1163-6413]
\fnmark[4]
\ead{debotoshbh@gmail.com}
\ead[url]{}
\credit{Supervision, Project administration, Writing -- review and editing}
\affiliation[3]{organization={Department of Computer Science and Engineering, 
            Jadavpur University},
            addressline={},
            city={Kolkata},
            postcode={700032},
            state={},
            country={India}}

\cortext[1]{Corresponding author}
\fntext[1]{}

\begin{abstract}
Spatial transcriptomics assays remain costly and technically demanding, 
restricting transcriptome-wide profiling to specialist settings and 
preventing routine clinical deployment. Predicting spatially resolved 
gene expression from H\&E histology could close this gap, yet current 
methods largely ignore the underlying tissue architecture and rarely 
quantify how their predictions can be trusted. We introduce 
\textit{\textbf{HierarchicalDAEW}}, a dual-graph architecture that 
addresses both gaps. On the spot graph, a Domain-Aware Edge-Weighted 
convolutional operator learns separate projections for inter-domain, 
intra-domain, and boundary edges derived from Leiden clustering, 
allowing the model to treat tissue heterogeneity as an explicit 
structural signal rather than an implicit one. A second gene-level 
graph then fuses protein-protein interaction priors from STRING-DB with 
tissue-specific co-expression through learned attention gating, 
propagating predictions from a landmark gene set to a broader gene 
panel. Reliability is handled through evidential uncertainty estimation, 
which produces far better calibrated confidence intervals than Monte 
Carlo dropout under identical conditions. Across six human Visium 
sections spanning breast, colorectal, prostate, and cerebellar tissue, 
and against thirteen published baselines, HierarchicalDAEW achieves the 
strongest correlation with ground-truth expression, with gains that hold 
up under multi-seed reproducibility checks and negative controls that 
rule out positional shortcuts. Ablations further confirm that both the 
domain-aware edge typing and the hierarchical depth are necessary to 
this improvement, and calibrated uncertainty estimates identify 
low-confidence predictions for pathologist review before clinical action.
\end{abstract}

\begin{highlights}
\item Domain-aware edge-weighted convolution learns separate projections per tissue domain edge type, explicitly encoding tissue heterogeneity into graph message passing.
\item Hierarchical DomainGCN and CrossScaleGate fuse spot- and domain-level representations, yielding +0.044 PCC over a flat graph baseline.
\item A gene graph decoder propagates landmark predictions to a broader gene panel via STRING-DB and co-expression edges (mean PCC 0.8314 on held-out genes).
\item Evidential uncertainty estimation achieves near-exact 90\% empirical coverage (0.903), confirmed by distribution-free conformal prediction guarantees.
\item HierarchicalDAEW outperforms thirteen published baselines across 
six human Visium sections, with calibrated uncertainty estimates that 
flag low-confidence predictions for pathologist review, supporting 
deployment as a clinical decision-support tool rather than a black-box 
classifier.
\end{highlights}

\date{}

\maketitle

\section{Introduction}\label{Introduction}

Spatial transcriptomics technologies such as 10x Visium \cite{staahl2016visualization} have transformed our ability to study tissue biology by preserving the spatial context of gene expression measurements, enabling researchers to relate transcriptional programs to tissue architecture, cell-cell interactions, and disease progression \cite{rao2021exploring,longo2021integrating}. However, the cost, technical complexity, and limited throughput of spatial transcriptomics assays restrict their use at the scale needed for large cohort studies or routine clinical deployment. Hematoxylin and eosin stained histology images, in contrast, are inexpensive, ubiquitous in clinical pathology workflows, and already routinely digitized. This has motivated a growing body of work on predicting spatially resolved gene expression directly from H\&E histology \cite{he2020integrating,zahedi2024deep,chelebian2025combining}, effectively using morphology as a proxy for the underlying transcriptional state of tissue.

Early approaches to this problem treated each tissue spot independently, mapping local histological features to expression profiles using convolutional or transformer-based encoders \cite{he2020integrating,pang2021leveraging}. While effective, these methods discard the spatial relationships between neighboring spots, relationships that are known to carry substantial biological signal since gene expression varies smoothly within tissue domains but shifts sharply across domain boundaries. Subsequent work addressed this by incorporating graph neural networks to model spot-to-spot spatial dependencies \cite{zeng2022spatial,mejia2023sepal,yang2024spatial}, and more recent architectures have introduced hierarchical or multi-resolution graph structures to better capture tissue heterogeneity at multiple scales \cite{ganguly2025merge,chung2024accurate}. Despite this progress, two important limitations persist across the current state of the art.

First, most existing graph-based methods treat all spatial edges identically, applying the same convolutional weights irrespective of whether two connected spots lie within the same tissue domain, across a domain boundary, or in structurally distinct regions altogether. This is a strong and often biologically implausible assumption: gene expression at a domain boundary is governed by different local processes than expression well within a homogeneous domain \cite{hu2021spagcn,dong2022deciphering}, and a model that shares weights across both settings is poorly positioned to capture the distinction. Second, the field has largely overlooked prediction reliability. Gene expression prediction from histology is an inherently uncertain task, since morphology is at best an imperfect surrogate for transcriptional state, yet the overwhelming majority of published methods report only point predictions with no accompanying measure of confidence. This is particularly consequential given that these models are ultimately intended to support downstream biological interpretation, where knowing when a prediction should not be trusted is as important as the prediction itself \cite{dolezal2022uncertainty,lambert2024trustworthy}.

In this work, we introduce \textbf{\textit{HierarchicalDAEW}}, a dual-graph architecture designed to overcome both of these limitations. At the spot level, we develop a domain-aware edge-weighted convolution that classifies edges as intra-domain, inter-domain, or boundary connections using Leiden-derived tissue domains. Instead of applying the same transformation to every neighboring spot, the model learns separate projections for each edge type following the spirit of relational graph convolutions \cite{schlichtkrull2018modeling}, allowing message passing to adapt to the underlying tissue structure rather than relying solely on spatial proximity.

At the gene level, we construct a second, structurally distinct graph and introduce a decoder that combines protein--protein interaction priors with tissue-specific co-expression relationships through a learned attention-gating mechanism \cite{gilmer2017neural}. This enables information from a compact set of spatially variable genes \cite{svensson2018spatialde,chen2024evaluating} to be propagated to a broader gene panel while preserving biologically meaningful relationships. Finally, to improve prediction reliability, we incorporate evidential uncertainty estimation using a Normal--Inverse--Gamma (NIG) loss \cite{amini2020deep,sensoy2018evidential}, allowing the model to generate calibrated confidence intervals for every prediction without the computational cost of ensemble methods \cite{lakshminarayanan2017simple} or repeated stochastic sampling \cite{gal2016dropout}.

We evaluate HierarchicalDAEW against thirteen published baselines across six human Visium tissue sections spanning breast, colorectal, prostate, and cerebellar tissue. Baseline comparisons are conducted under both single-section and multi-section joint training on breast tissue \cite{niu2025spabatch}, while generalization to colorectal, prostate, and cerebellar sections is assessed separately. We further evaluate cross-section transfer through leave-one-dataset-out and few-shot adaptation, calibration against Monte Carlo dropout \cite{gal2016dropout} and conformal prediction \cite{stephen2021gentle}, and robustness via negative controls and multi-seed reproducibility \cite{tibshirani1993introduction,cohen2013statistical}. Results show that domain-aware edge typing yields consistent accuracy gains over both graph-based and non-graph baselines, while evidential uncertainty estimation delivers substantially better-calibrated confidence than existing alternatives.

Spatial transcriptomics assays cost several hundred dollars per section 
and require specialized equipment unavailable in most clinical 
laboratories, making transcriptome-wide profiling impractical at the 
scale needed for routine cancer care \cite{staahl2016visualization,
rao2021exploring}. HierarchicalDAEW predicts spatially resolved gene 
expression directly from H\&E histology slides that are already 
generated as standard of care \cite{he2020integrating}, offering a 
path toward scalable molecular profiling without additional assay cost. 
Calibrated uncertainty estimates further identify predictions that 
should not be trusted \cite{lambert2024trustworthy,dolezal2022uncertainty}, 
enabling pathologists to flag low-confidence spots for follow-up rather 
than acting on unreliable predictions. This combination of scalable 
prediction and honest uncertainty quantification positions 
HierarchicalDAEW as a decision-support tool rather than a black-box 
classifier \cite{Khawaled_2024,chereda2024stable}.

Our contributions are summarized as follows:
\begin{itemize}
    \item We propose a domain-aware edge-weighted convolution operator that explicitly types spatial edges as intra-domain, inter-domain, or boundary based on Leiden-derived tissue domains \cite{hu2021spagcn}, learning separate projections per edge type \cite{schlichtkrull2018modeling} rather than sharing weights across structurally distinct spatial relationships.
    \item We introduce a gene-level graph decoder that fuses protein-protein interaction priors with tissue-specific co-expression relationships through learned attention gating \cite{gilmer2017neural}, propagating predictions from a compact spatially variable gene set \cite{svensson2018spatialde} to a broader gene panel.
    \item We incorporate evidential uncertainty estimation via a Normal-Inverse-Gamma loss \cite{amini2020deep}, yielding calibrated per-prediction confidence intervals that substantially outperform Monte Carlo dropout \cite{gal2016dropout} and remain informative under conformal prediction \cite{stephen2021gentle} and dataset shift, without the computational cost of ensembling \cite{lakshminarayanan2017simple}.
    \item We conduct an extensive empirical evaluation against thirteen published baselines across six tissue sections, comparing single-section and multi-section joint training on breast tissue, and separately assessing cross-tissue generalization, leave-one-dataset-out transfer, negative controls, and multi-seed reproducibility, demonstrating that our gains stem from principled architectural choices rather than incidental factors.
\end{itemize}

%% ─────────────────────────────────────────────────────────────────────────────
\section{Related Work}\label{related_work}
%% ─────────────────────────────────────────────────────────────────────────────

\subsection{Spatial Transcriptomics and Gene Expression Prediction from Histology}
\label{sec:rw-st}

Spatial transcriptomics, pioneered by \citet{staahl2016visualization}, enables the measurement of gene expression while preserving the spatial organization of tissue, offering a window into how transcriptional programs relate to tissue architecture and disease progression \cite{rao2021exploring,longo2021integrating}. Reviews and benchmarks of the field have documented the rapid growth of computational methods for analysing spatially resolved data \cite{zahedi2024deep, ruiz2025completing}.

The problem of predicting spatial gene expression from H\&E histology was introduced by \citet{he2020integrating} with STNet, a DenseNet-based encoder that maps local image patches to spot-level expression. \citet{pang2021leveraging} extended this with HisToGene, incorporating a Vision Transformer \cite{dosovitskiy2020image} to capture contextual relationships between patches. Hist2ST \cite{zeng2022spatial} combined transformer-based image features with graph neural networks, constructing a spatial graph over spots and jointly learning from visual and positional context. More recently, BLEEP \cite{xie2023spatially} introduced bi-modal contrastive learning between image patches and gene expression profiles, while EGGN \cite{yang2023exemplar,yang2024spatial} proposed exemplar-guided retrieval to augment local predictions with information from visually similar spots. TRIPLEX \cite{chung2024accurate} applied multi-resolution feature fusion to capture structure at multiple spatial scales, and SEPAL \cite{mejia2023sepal} framed prediction as a local graph regression problem. THItoGene \cite{jia2024thitogene} combined dynamic convolution and capsule networks with graph-based aggregation, while MERGE \cite{ganguly2025merge} introduced a multi-faceted hierarchical GNN explicitly designed for whole-slide histopathology images. Boundary-guided approaches \cite{qu2024boundary} have further explored how tissue boundary structure can improve spatial predictions. Comprehensive benchmarking \cite{ruiz2025completing} has highlighted the importance of consistent evaluation protocols across these methods, and cross-modal mask reconstruction with contrastive learning \cite{liu2025spatialtranscriptomicsexpressionprediction} offers complementary directions for self-supervised spatial prediction.

More recent work has moved toward foundation model-scale approaches, with multimodal models integrating histology and transcriptomics at scale \cite{xiang2026multimodal,lin2024st} and adapting single-cell foundation models for spatial generation \cite{fang2026adapting}. Our work is positioned within this continuum: rather than scaling to larger foundation models, we focus on explicitly exploiting tissue domain structure within the graph message-passing mechanism itself, an inductive bias that complements both retrieval-based and contrastive approaches.

\subsection{Graph Neural Networks for Spatial Biology}
\label{sec:rw-gnn}

Graph neural networks (GNNs) have become a foundational tool for learning on structured spatial and biological data \cite{gilmer2017neural,zhang2021graph}. Spectral graph convolutions \cite{kipf2016semi} and attention-based aggregation \cite{velivckovic2017graph} provide the core message-passing mechanisms underlying most subsequent work, while inductive representations \cite{hamilton2017inductive} generalize to unseen nodes without requiring the full graph during training. The PyTorch Geometric library \cite{fey2019fast} has made large-scale GNN experimentation practical.

In spatial biology, GNNs have been applied primarily to tissue domain identification and cell clustering. SpaGCN \cite{hu2021spagcn} pioneered the integration of histology, spatial coordinates, and expression for domain detection, while STAGATE \cite{dong2022deciphering} introduced adaptive graph attention autoencoders for finer domain boundaries. GraphST \cite{long2023spatially} demonstrated spatially informed clustering and deconvolution using graph-structured self-supervised learning. Benchmarks of these clustering methods \cite{yuan2024benchmarking} and surveys of GNN approaches to spatial transcriptomics \cite{liu2023comprehensive,li2022cell} have characterized the landscape of available techniques. Graph convolutional networks have further demonstrated strong performance in cancer biomarker discovery directly from molecular network structure \cite{chereda2024stable}, 
motivating their extension to spatially resolved biological graphs 
as pursued in this work. Multi-slice integration via graph neural networks \cite{niu2025spabatch} has extended these ideas to three-dimensional tissue reconstruction.

The treatment of edge heterogeneity in spatial graphs has received less attention. Relational graph convolutional networks \cite{schlichtkrull2018modeling} introduced the idea of separate projections per relation type in knowledge graphs; we adapt this principle to the spatial biology setting, where edge types are defined by biologically motivated Leiden domain clustering rather than a predefined ontology. Unlike prior spatial GNN methods that treat all spot-to-spot edges identically, our DAEWConv explicitly differentiates intra-domain, inter-domain, and boundary edges through per-type learned projections and learnable scalar gates, providing a direct inductive bias for the known biological discontinuity at tissue domain boundaries.

\subsection{Uncertainty Quantification in Deep Learning}
\label{sec:rw-uq}

Quantifying the uncertainty of neural network predictions has been an active research area since the introduction of Monte Carlo dropout as a Bayesian approximation \cite{gal2016dropout} and the demonstration that deep ensembles provide strong calibration with minimal architectural overhead \cite{lakshminarayanan2017simple}. Calibration of neural network confidence estimates was studied systematically by \citet{kuleshov2018accurate}, who proposed regression calibration metrics including the expected normalized calibration error used in our evaluation. Reviews of uncertainty quantification in medical image analysis \cite{lambert2024trustworthy} have documented the importance of reliable confidence estimates for clinical deployment.

In computational pathology and medical imaging more broadly, 
uncertainty-informed deep learning has been shown to enable 
high-confidence predictions for digital histopathology classification 
\cite{dolezal2022uncertainty} and reliable reconstruction from 
undersampled medical imaging data \cite{Khawaled_2024}, establishing 
principled uncertainty quantification as a practical requirement for 
clinical deployment rather than a theoretical nicety. Evidential deep learning \cite{amini2020deep,sensoy2018evidential} provides an alternative to sampling-based approaches, directly modelling higher-order uncertainty through a Normal-Inverse-Gamma distribution over the predictive distribution, obtaining calibrated intervals from a single forward pass. Conformal prediction \cite{stephen2021gentle} offers a complementary, distribution-free approach, providing marginal coverage guarantees under only exchangeability, without assuming any particular form of the predictive distribution.

Our work applies evidential regression \cite{amini2020deep} to gene expression prediction, a regression setting where the high dimensionality of the output (512 genes per spot) and the heterogeneity of expression patterns across tissue types makes reliable uncertainty quantification particularly challenging. We compare directly against Monte Carlo dropout \cite{gal2016dropout} and augment NIG-based intervals with conformal prediction \cite{stephen2021gentle} to provide distribution-free coverage guarantees, a combination not previously demonstrated in the spatial transcriptomics prediction literature.

\section{Methodology}\label{methodology}

\subsection{Problem Formulation} \label{sec:problem-formulation}

Let a spatial transcriptomics section consist of $N$ tissue spots indexed by $i \in \{1,\dots, N\}$, each associated with a histology patch centered at coordinate $p_i \in \mathbb{R}^2$, a $d$-dimensional image-derived feature vector $x_i \in \mathbb{R}^d$ extracted by a pretrained encoder (Section~\ref{sec:feature-extraction}), and an observed expression vector $y_i \in \mathbb{R}^G$ over a panel of spatially variable genes (Section~\ref{sec:svg-selection}).

We construct a spot graph $\mathcal{G}_s= (V,E_s)$, with node set $V=\{1,2,\dots,N\}$ and edge set $E_s$, using $k$-nearest neighbors in physical coordinate space. Edges are weighted by inverse Euclidean distance,
\begin{equation}
    w_{ij} = \frac{1}{\lVert p_i - p_j \rVert_2 +\epsilon}, \hspace{1cm} (i,j) \in E_s,
\end{equation}
where $\epsilon$ is a small constant preventing division by zero. Each edge is further assigned a discrete type $r_{ij} \in \{0,1,2\}$, indicating whether it connects spots within the same tissue domain, across two distinct domains, or along a domain boundary, based on Leiden clustering of the expression data (Section~\ref{sec:leiden}). This typing lets the model treat spatial proximity differently depending on tissue context, rather than assuming one relationship for all neighboring spots.

In parallel, we define a gene graph $\mathcal{G}_g =(V_g, E_g)$, where the node set $V_g = \{1,\dots,G\}$ corresponds to the same gene panel as $y_i$, and edges combine curated protein-protein interaction priors with empirical co-expression correlations computed from training data (Section~\ref{sec:genegraph}). This graph refines per-spot expression estimates after an initial prediction from the spot graph alone.

Given $\mathcal{G}_s, \mathcal{G}_g$ and learnable parameters $\theta$, our objective is to learn a function
\begin{equation}
    f_\theta : (\mathcal{G}_s, \mathcal{G}_g) \longmapsto \{(\hat{y}_i, \hat{\sigma}_i^2)\}_{i=1}^{N},
\end{equation}
predicting, for every spot, an expression estimate $\hat{y}_i \in \mathbb{R}^G$ together with a calibrated per-gene uncertainty $\hat{\sigma}_i^2 \in \mathbb{R}_{+}^G$. Sections~\ref{sec:feature-extraction} to \ref{sec:multisection} describe each component of $f_\theta$ in turn.
\begin{figure*}[t]
\centering
\includegraphics[width=\textwidth, keepaspectratio]{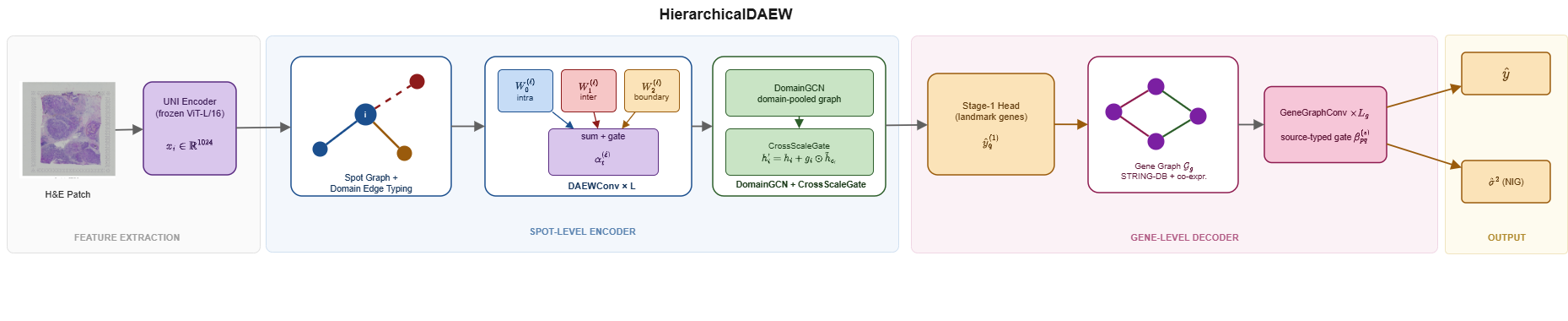}
\caption{Overview of HierarchicalDAEW. A frozen UNI encoder extracts spot features, which are processed by a domain-aware spot-level encoder (DAEWConv, DomainGCN, CrossScaleGate) to produce Stage-1 landmark predictions, propagated by a gene graph decoder to a broader panel with calibrated evidential uncertainty.}
\label{fig:architecture}
\end{figure*}
\subsection{Histology Feature Extraction} \label{sec:feature-extraction}

For each tissue spot, we extract a fixed-size feature vector from the histology image patch centered at its spatial coordinate. A patch of radius $r$ pixels is cropped from the whole-slide H\&E image, resized to $224 \times 224$, and normalized using standard ImageNet channel statistics with mean [0.485, 0.456, 0.406] and standard deviation of [0.229, 0.224, 0.225]. The normalized patch is passed through UNI \cite{chen2024towards}, a vision transformer (ViT-L/16) \cite{dosovitskiy2020image} pretrained on large-scale histopathology data via self-supervised learning, producing a 1024-dimensional embedding $x_i \in \mathbb{R}^{1024}$ for spot $i$. Unlike encoders pretrained on natural images such as ResNet-50 \cite{he2016deep}, UNI is trained specifically on histology, and its representations are expected to better reflect tissue morphology and cellular structure relevant to gene expression prediction. We also evaluated Prov-GigaPath \cite{xu2024whole} as an alternative encoder; UNI outperformed it in ablation experiments despite GigaPath's larger parameter count, consistent with the observation that histopathology-specific pretraining matters more than encoder scale in this setting.

The encoder is used strictly as a frozen feature extractor. Its weights are not updated during training, and only extracted embeddings are consumed by the downstream graph network. The resulting feature vector $x_i$ serves as the initial node representation in the spot graph $\mathcal{G}_s$ and remains fixed throughout training, decoupling representation learning from the graph-based prediction stage.

\subsection{Spatially Variable Gene (SVG) Selection via Moran's I} \label{sec:svg-selection}

Whole-transcriptome expression panels are dominated by genes with low or spatially unstructured expression, contributing little biological signal while destabilizing training \cite{svensson2018spatialde,chen2024evaluating,li2023benchmarking}. We therefore restrict prediction to a panel of $G$ spatially variable genes, selected using Moran's I statistic \cite{chen2024evaluating}, a standard measure of spatial autocorrelation. For a spatial neighborhood graph constructed over the six nearest spots, with edge weights $w_{ij}$ indicating adjacency between spots $i$ and $j$, Moran's I for gene $g$ is defined as
\begin{equation}
I_g = \frac{N}{\sum_{i,j} w_{ij}} \cdot \frac{\sum_{i,j} w_{ij} (y_{ig} - \bar{y}_g)(y_{jg} - \bar{y}_g)}{\sum_{i} (y_{ig} - \bar{y}_g)^2},
\label{eq:moran}
\end{equation}
where $N$ is the number of spots, $y_{ig}$ is the expression of gene $g$ at spot $i$, and $\bar{y}_g$ is its mean expression across all spots. Values of $I_g$ near $1$ indicate strong positive spatial autocorrelation, while values near $0$ indicate spatially random expression. Statistical significance is assessed via a permutation test with 100 permutations, in which spot labels are randomly shuffled to construct a null distribution of $I_g$, yielding an empirical $p$-value $p_g$ for each gene. Genes with $p_g \geq 0.05$ are discarded, and the remaining genes are ranked by $I_g$. The top $G=512$ genes are retained as the prediction target panel. When spatial autocorrelation cannot be computed for a section, we fall back to ranking genes by normalized dispersion. Gene selection is performed independently within each training fold using only training-split spots, ensuring that spatial structure present in held-out test spots never influences panel composition.

\subsection{Leakage-Free Leiden Domain Assignment}
\label{sec:leiden}

To identify coherent tissue domains, we apply Leiden community detection \cite{hu2021spagcn,dong2022deciphering,long2023spatially} to the expression profiles of each section. A shared nearest-neighbor graph is built over the top 30 principal components of the expression data, using 15 neighbors per spot, and Leiden clustering is applied at a resolution of $0.7$, selected via ablation over the range $\{0.3, 0.5, 0.7, 1.0, 1.5\}$ and applied uniformly across all sections, yielding between 8 and 20 domains depending on the underlying tissue heterogeneity of each section. This produces a domain label $c_i \in \{1, \dots, C\}$ for every spot $i$.

Since domain labels are later used to construct graph edge types for training (Section~\ref{sec:daew}), computing them on the full dataset would allow information from test spots to influence the structure seen during training, a subtle form of label leakage. We therefore restrict Leiden clustering to training spots only within each cross-validation fold. Domain labels for held-out test spots are then assigned by nearest-centroid matching in principal component space,
\begin{equation}
c_i = \operatorname*{arg\,min}_{c \in \{1,\dots,C\}} \left\lVert z_i - \mu_c \right\rVert_2, \qquad i \notin \mathcal{D}_{\text{train}},
\label{eq:centroid}
\end{equation}
where $z_i$ is the principal-component representation of test spot $i$ and $\mu_c$ is the centroid of training spots assigned to domain $c$, ensuring that no test-spot expression contributes to domain formation while still providing every spot with a domain label.

Given the resulting domain labels, we define a spot $i$ as a boundary spot if $\mathbb{1}[\exists\, j \in \mathcal{N}(i) : c_j \neq c_i] = 1$, where $\mathcal{N}(i)$ denotes its spatial $k$-nearest neighbors. Each edge $(i,j)$ in the spot graph is then assigned a type
\begin{equation}
r_{ij} =
\begin{cases}
0 & \text{if } c_i = c_j \text{ and } i \text{ is not a boundary spot (intra-domain)}, \\
1 & \text{if } c_i \neq c_j \text{ (inter-domain)}, \\
2 & \text{if } c_i = c_j \text{ and } i \text{ is a boundary spot (boundary)}.
\end{cases}
\label{eq:edgetype}
\end{equation}

This typing forms the structural input consumed by the domain-aware convolution described next.

\subsection{Domain-Aware Edge-Weighted Graph Convolution (DAEWConv)}
\label{sec:daew}

Standard graph convolutional layers \cite{kipf2016semi} aggregate information from neighboring nodes using a single shared transformation, implicitly treating all spatial relationships as equivalent. In the spot graph, however, edges of different types $r_{ij} \in \{0, 1, 2\}$ (intra-domain, inter-domain, boundary) reflect qualitatively different local tissue contexts, and a single shared weight matrix is poorly suited to capturing this distinction. We address this with DAEWConv, a domain-aware edge-weighted convolution that learns a separate linear projection for each edge type, allowing message passing to adapt to the underlying tissue structure rather than treating spatial proximity as uniform, following the relational message-passing framework of \citet{schlichtkrull2018modeling} adapted for biologically typed spatial graphs.

Let $h_i^{(\ell)} \in \mathbb{R}^{d_\ell}$ denote the representation of spot $i$ at layer $\ell$, with $h_i^{(0)} = x_i$ given by the histology feature. For each edge type $t \in \{0,1,2\}$, DAEWConv maintains an independent projection matrix $W_t^{(\ell)} \in \mathbb{R}^{d_{\ell+1} \times d_\ell}$. A message from spot $j$ to spot $i$ along an edge of type $r_{ij} = t$ is computed as
\begin{equation}
m_{i \leftarrow j}^{(\ell)} = \alpha_t^{(\ell)} \cdot \tilde{d}_i^{-1/2} \tilde{d}_j^{-1/2} \, W_t^{(\ell)} h_j^{(\ell)},
\label{eq:daew_message}
\end{equation}
where $\tilde{d}_i$ is the degree of spot $i$ (clamped to a minimum of $1$), $\tilde{d}_i^{-1/2}\tilde{d}_j^{-1/2}$ is the symmetric degree normalization standard in spectral graph convolutions \cite{kipf2016semi}, and $\alpha_t^{(\ell)}$ is a learnable, layer-specific gate controlling the contribution of edge type $t$. Messages are aggregated over neighbors, partitioned by edge type:
\begin{equation}
h_i^{(\ell+1)} = \sigma \left( \mathrm{BN}\left( \sum_{t=0}^{2} \sum_{j \in \mathcal{N}_t(i)} m_{i \leftarrow j}^{(\ell)} + b^{(\ell)} \right) + W_{\mathrm{res}}^{(\ell)} h_i^{(\ell)} \right),
\label{eq:daew_update}
\end{equation}
where $\mathcal{N}_t(i)$ denotes neighbors connected via edge type $t$, $\mathrm{BN}(\cdot)$ is batch normalization, $b^{(\ell)}$ is a learnable bias, $\sigma$ is the ReLU nonlinearity, and $W_{\mathrm{res}}^{(\ell)}$ is a residual projection (the identity when $d_\ell = d_{\ell+1}$) that stabilizes gradient flow across layers. Dropout is applied to the layer output during training.

The per-type projections $\{W_t^{(\ell)}\}_{t=0}^{2}$ allow the model to learn distinct aggregation behavior for each structural context: intra-domain edges ($t{=}0$) can emphasize smooth propagation of expression signal within a homogeneous region, while boundary edges ($t{=}2$) can learn a sharper transformation reflecting the discontinuity typically observed at domain interfaces \cite{dong2022deciphering}. The learnable gates $\alpha_t^{(\ell)}$ further allow the network to modulate the relative importance of each edge type per layer. Unlike relation-specific convolutions that assign fixed weight matrices to a predefined taxonomy of relations, our edge types are derived dynamically from unsupervised domain clustering computed per section, allowing DAEWConv to adapt its structural inductive bias to the specific tissue architecture of each sample.

\begin{center}
\includegraphics[width=0.95\linewidth]{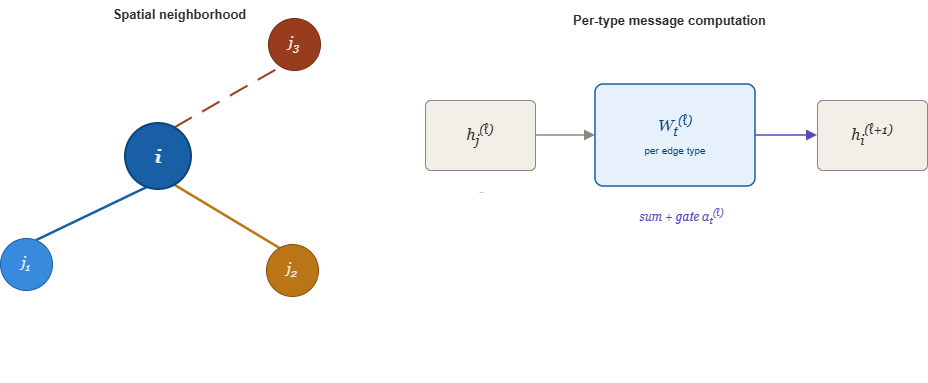}
\captionof{figure}{DAEWConv edge typing and per-type message passing. Intra-domain, inter-domain, and boundary edges are each routed through a distinct learned projection before a gated aggregation step.}
\label{fig:daewconv}
\end{center}

\subsection{Hierarchical Architecture: DomainGCN and CrossScaleGate}\label{hierarchical}

DAEWConv operates locally, propagating information only across immediate spatial neighbors. Many biologically relevant patterns, however, span an entire tissue domain rather than a small neighborhood \cite{hu2021spagcn,long2023spatially}, motivating a second, coarser level of representation that reasons over domains as whole units. After $L$ stacked DAEWConv layers produce spot-level representations $h_i \in \mathbb{R}^d$, we construct the domain-level representation by mean-pooling spot features within each domain,
\begin{equation}
    h_c = \frac{1}{|\{i:c_i=c\}|}\sum_{i:c_i=c}h_i, \hspace{1cm} c = 1,\dots,C,
\end{equation}
where $c_i$ is the domain label of spot $i$. This yields one representation per domain, summarizing the aggregate state of that region.

To allow domains to exchange information with one another, particularly along regions of physical contact, we construct a domain-level graph $\mathcal{G}_d$ in which two domains $c$ and $c'$ are connected if at least one boundary edge in the spot graph links a spot in domain $c$ to a spot in domain $c'$. A single graph convolutional layer \cite{kipf2016semi}, DomainGCN, is applied over $\mathcal{G}_d$:
\begin{equation}
    \tilde{h}_c = \sigma(\mathrm{BN}(\mathrm{GCNConv}(h_c, \mathcal{G}_d))),
\end{equation}
producing domain representations that incorporate information from adjacent tissue regions. When no valid inter-domain boundary connections exist for a section, $\mathcal{G}_d$ falls back to self-loops on each domain, so DomainGCN still applies a learned transformation to $h_c$ without requiring inter-domain messages.

The refined domain representation $\tilde{h}_c$ is then fused back into every spot belonging to domain $c$ through a CrossScaleGate, a learned gating mechanism that lets each spot decide how much domain-level context to incorporate:
\begin{equation}
    g_i = \sigma_{\mathrm{sig}}(W_g h_i), \hspace{1cm} h_i' = h_i + g_i \odot (W_c\tilde{h}_{c_i}),
\end{equation}
where $\sigma_{\mathrm{sig}}$ denotes the sigmoid function, $W_g \in \mathbb{R}^{1 \times d}$ and $W_c \in \mathbb{R}^{d \times d}$ are learned parameters, and $\odot$ is elementwise multiplication. This additive, gated fusion allows spots with strong, well-supported local signal to rely primarily on $h_i$, while spots in more ambiguous or boundary-adjacent regions can draw more heavily on domain-level context $\tilde{h}_{c_i}$. The resulting spot representations $h_i'$ are passed to the expression and uncertainty heads described next.

\subsection{Gene Graph Decoder: Source-Typed Message Passing and Landmark Propagation}
\label{sec:genegraph}

The spot-level representations produced by the hierarchical spot encoder yield predictions only for a compact set of $G$ landmark genes selected in the gene-selection stage. To extend these predictions to a broader gene panel while respecting known biological relationships between genes \cite{gilmer2017neural}, we introduce a second-stage decoder that operates over the gene graph $\mathcal{G}_g$, whose nodes correspond to genes rather than spots.

Edges in $\mathcal{G}_g$ are drawn from two structurally distinct sources: curated protein--protein interaction priors from STRING-DB \cite{zhang2021graph}, and empirical co-expression correlations computed from training spots only. Each edge $(p,q)$ is annotated with a source label $s_{pq} \in \{0,1\}$ (STRING or co-expression) and a scalar weight $w_{pq}$ given by the corresponding interaction confidence or absolute correlation magnitude; self-loops are added for genes without any incoming edge, and both sources may be combined for the same gene pair when evidence overlaps. Every gene $q$ is assigned a learnable embedding $e_q \in \mathbb{R}^{h}$, and landmark genes additionally receive a residual signal projected from their Stage-1 prediction $\hat{y}_q^{(1)}$, giving the initial gene representation
\begin{equation}
    h_q^{(0)} = e_q + \mathbb{1}[q \text{ is landmark}] \cdot W_{\mathrm{lm}}\, \hat{y}_q^{(1)}.
\end{equation}

Representations are then propagated through $L_g$ layers of GeneGraphConv, a source-typed message-passing operator \cite{gilmer2017neural} analogous in spirit to DAEWConv but applied over the gene graph. For source $s \in \{0,1\}$, a message from gene $q$ to gene $p$ is gated by both the fixed edge weight $w_{pq}$ and a learned, content-dependent attention score:
\begin{equation}
    \beta_{pq}^{(s)} = \sigma_{\mathrm{sig}}\Big(W_{\mathrm{gate}}^{(s)}\,\big[h_p^{(\ell)} \,\|\, W_s^{(\ell)} h_q^{(\ell)}\big]\Big) \cdot \min\big(|w_{pq}|,1\big),
\end{equation}
\begin{equation}
    h_p^{(\ell+1)}=\sigma\Big(\mathrm{BN}\Big(\sum_{s=0}^1 \sum_{q\,:\,s_{pq}=s} \beta_{pq}^{(s)} W_s^{(\ell)} h_q^{(\ell)}\Big)+W_{\mathrm{res}}^{(\ell)} h_p^{(\ell)}\Big),
\end{equation}
where $[\cdot\|\cdot]$ denotes concatenation and $\sigma_{\mathrm{sig}}$ the sigmoid function. This allows the decoder to learn distinct propagation behavior for physical protein interactions versus statistical co-expression, while the attention gate further modulates each message by how relevant it is to the receiving gene's current state, rather than relying on the fixed edge weight alone.

After $L_g$ layers, a shared feed-forward readout maps each gene's final representation $h_q^{(L_g)}$ to a scalar expression prediction $\hat{y}_q^{(2)}$. Landmark genes therefore retain a direct residual path to their Stage-1 estimate while still being refined through graph propagation, whereas non-landmark genes are inferred purely from their position in the gene graph relative to genes with observed Stage-1 predictions.
\begin{center}
\includegraphics[width=0.95\linewidth]{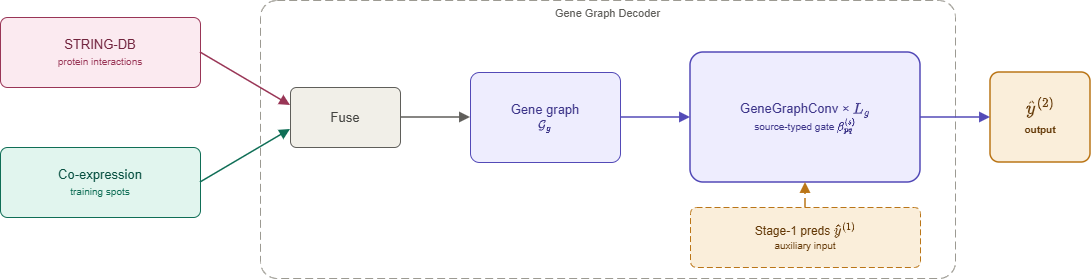}
\captionof{figure}{Gene graph decoder construction. STRING-DB and co-expression edges are fused into a single gene graph, propagated by GeneGraphConv alongside a residual signal from Stage-1 landmark predictions.}
\label{fig:genegraph}
\end{center}
\subsection{Training Objectives: MSE, NIG Evidential Loss and Domain Contrastive Loss}
\label{sec:objectives}

HierarchicalDAEW is trained end-to-end with a combined objective that supervises expression accuracy, predictive uncertainty, and the quality of domain-structured representations.

The primary supervision signal is a standard mean squared error between predicted and observed expression, $\mathcal{L}_{\mathrm{mse}} = \frac{1}{NG}\sum_{i,g} (\hat{y}_{ig} - y_{ig})^2$. Rather than predicting a single point estimate, the expression head instead outputs the parameters of a Normal-Inverse-Gamma distribution $(\mu_{ig}, \nu_{ig}, \alpha_{ig}, \beta_{ig})$ per spot-gene pair \cite{amini2020deep}, from which the point prediction $\hat{y}_{ig} = \mu_{ig}$ and an aleatoric-plus-epistemic uncertainty estimate are jointly derived. These parameters are supervised using the evidential regression loss \cite{amini2020deep,sensoy2018evidential}, which combines a negative log-likelihood term under the predictive Student-$t$ distribution induced by the NIG prior with a regularizer that penalizes high model evidence at large residuals:
\begin{equation}
\begin{split}
\mathcal{L}_{\mathrm{NIG}} = \;& \tfrac{1}{2}\log\frac{\pi}{\nu} - \alpha \log\big(2\beta(1+\nu)\big) \\
& + \Big(\alpha+\tfrac{1}{2}\Big)\log\Big(\nu(y-\mu)^2 + 2\beta(1+\nu)\Big) \\
& + \log\Gamma(\alpha) - \log\Gamma\Big(\alpha+\tfrac{1}{2}\Big) \\
& + \lambda_{\mathrm{ev}}\,|y-\mu|\,(2\nu+\alpha),
\end{split}
\label{eq:nig}
\end{equation}
where the regularization weight $\lambda_{\mathrm{ev}}$ controls the trade-off between fit and evidential calibration; larger values more strongly discourage the model from expressing confidence in regions of large error.

To further encourage representations that respect the tissue domain structure introduced earlier, we add a supervised contrastive loss \cite{xie2023spatially} over spot embeddings $z_i$ (L2-normalized pre-readout representations). For each intra-domain or boundary edge $(i,j)$ whose endpoints have highly similar observed expression profiles (cosine similarity above $0.8$), the pair is treated as a positive; spot pairs connected by an inter-domain edge serve as negatives. Following the InfoNCE formulation, the loss for a sampled positive pair $(a,p)$ against a batch of negatives $\{n\}$ is
\begin{equation}
    \mathcal{L}_{\mathrm{ctr}} = -\log \frac{\exp(z_a^\top z_p / \tau)}{\exp(z_a^\top z_p / \tau) + \sum_{n} \exp(z_a^\top z_n / \tau)},
\end{equation}
using a temperature of $\tau = 0.1$, selected via ablation over the range $\{0.01, 0.05, 0.07, 0.1, 0.3\}$, encouraging spots with similar expression within the same domain to have similar embeddings while pushing apart embeddings across domain boundaries.

The overall training objective is a weighted sum,
\begin{equation}
\mathcal{L} = \mathcal{L}_{\mathrm{mse}} + \lambda_{\mathrm{NIG}}\, \mathcal{L}_{\mathrm{NIG}} + \lambda_{\mathrm{ctr}}\, \mathcal{L}_{\mathrm{ctr}},
\end{equation}
with $\lambda_{\mathrm{NIG}}$ and $\lambda_{\mathrm{ctr}}$ selected via the hyperparameter search described later. All three terms are computed on Stage-1 landmark predictions; the Stage-2 decoder is supervised by MSE alone against its own expanded gene panel.

\subsection{Multi-Section Joint Training with Disconnected Section Graphs}
\label{sec:multisection}

Training on a single tissue section limits the diversity of morphological and expression patterns the model observes and provides no signal about whether learned representations generalize across independently prepared samples \cite{niu2025spabatch}. We therefore also train HierarchicalDAEW jointly across multiple sections of the same tissue type, treating each section as an independent subgraph within a single training batch.

For a set of $K$ sections, we first restrict the gene panel to the intersection of genes measured across all sections, and independently select spatially variable genes and Leiden tissue domains within each section, exactly as described earlier, so that no cross-section information leaks into either gene selection or domain assignment. Each section $k$ therefore yields its own spot graph $\mathcal{G}_s^{(k)}$ with its own node features, domain labels, and edge types.

These per-section graphs are combined into a single disconnected graph $\mathcal{G}_s^{\mathrm{joint}} = \bigcup_{k=1}^{K} \mathcal{G}_s^{(k)}$ by concatenating node features and offsetting node indices in each section's edge list by the cumulative spot count of preceding sections,
\begin{equation}
    E_s^{\mathrm{joint}} = \bigcup_{k=1}^{K} \left\{ (i + o_k,\, j + o_k) \mid (i,j) \in E_s^{(k)} \right\}, \;
    o_k = \sum_{k'<k} N_{k'}
    \label{eq:joint_graph}
\end{equation}
where $N_{k'}$ is the number of spots in section $k'$. Crucially, no edges are added between spots belonging to different sections, since spatial adjacency and tissue domain structure are only physically meaningful within a single tissue slide. DAEWConv and the domain-level pooling described earlier therefore operate correctly on $\mathcal{G}_s^{\mathrm{joint}}$ without modification, as message passing naturally respects the block-diagonal connectivity structure: representations for spots in section $k$ are influenced only by other spots in section $k$, while all sections contribute simultaneously to a single set of shared model parameters through backpropagation.

This joint training scheme allows the domain-aware convolution and gene graph decoder to learn parameters that are consistent across independently collected sections, rather than overfitting to the specific morphological or technical characteristics of a single sample, and forms the basis of our primary evaluation setting.

\section{Experimental Setup}
\label{sec:expsetup}

\subsection{Datasets}
\label{sec:datasets}

We evaluate HierarchicalDAEW on six publicly available 10x Genomics Visium sections \cite{staahl2016visualization}, obtained through the Squidpy dataset repository, spanning four tissue types and two distinct sample preservation protocols. All sections are profiled using the standard Visium platform, in which gene expression is captured on a $6.5 \times 6.5$ mm array of barcoded spots, each $55\,\mu\text{m}$ in diameter with a center-to-center spacing of $100\,\mu\text{m}$, such that each spot aggregates transcripts from several adjacent cells rather than resolving single-cell expression directly.

Breast tissue is represented by three independently processed sections. Breast S1 and Breast S2 are fresh-frozen sections derived from the same invasive ductal carcinoma tumor block, cryosectioned and processed under the standard Visium tissue preparation protocol, and represent biological and technical replicates of one another. Breast FFPE is a formalin-fixed, paraffin-embedded section from a separate breast cancer sample, prepared under a distinct fixation protocol known to degrade RNA quality and alter capture efficiency relative to fresh-frozen tissue. Together, these three sections form the basis of our multi-section joint training experiments, allowing us to assess robustness to both biological replicate variation and cross-protocol differences within a single cancer type.

To evaluate generalization beyond breast cancer, we additionally include a colorectal cancer section, a prostate acinar cell carcinoma FFPE section, and a human cerebellum section representing healthy, non-cancerous neural tissue. The cerebellum section provides an important contrast to the five cancer-derived sections, as normal tissue exhibits substantially different structural organization and, as observed in our quality control statistics (Section~\ref{sec:qc}), higher baseline mitochondrial transcript content. Together, the six sections span both malignant and non-malignant tissue, fresh-frozen and FFPE preservation protocols, and four anatomical sites, providing a reasonably diverse testbed for evaluating whether architectural improvements generalize beyond a single tissue type or cancer indication. Table~\ref{tbl:datasets} summarizes the raw composition of each section as obtained from the repository, prior to quality control.

\begin{table}[t]
\caption{Raw dataset composition.}
\label{tbl:datasets}
\small
\centering
\resizebox{\columnwidth}{!}{%
\begin{tabular}{lllrr}
\toprule
Section & Tissue Type & Prep. & Spots & Genes \\
\midrule
Breast S1        & Ductal carcinoma   & FF   & 3,798 & 36,601 \\
Breast S2        & Ductal carcinoma   & FF   & 3,987 & 36,601 \\
Breast FFPE      & Breast cancer      & FFPE & 2,518 & 17,943 \\
Colorectal       & Colorectal cancer  & FF   & 3,138 & 36,601 \\
Prostate FFPE    & Prostate carcinoma & FFPE & 3,043 & 17,943 \\
Human Cerebellum & Cerebellum         & FF   & 4,992 & 36,601 \\
\bottomrule
\end{tabular}%
}
\end{table}
\subsection{Preprocessing and Quality Control}
\label{sec:qc}

Each section undergoes standard quality control prior to modeling \cite{hu2021spagcn,long2023spatially}. Spots with fewer than 200 detected genes and genes detected in fewer than 10 spots are removed to exclude low-quality spots and rarely expressed, uninformative genes. Spots with a mitochondrial read fraction above $20\%$ are additionally excluded, as high mitochondrial content typically indicates cellular stress or tissue damage; this threshold is relaxed for the cerebellum section, which exhibits substantially higher baseline mitochondrial activity than the cancerous sections and would otherwise lose a disproportionate fraction of valid spots. Following spot- and gene-level filtering, expression counts are library-size normalized to 10,000 counts per spot and log-transformed, after which the top 10,000 highly variable genes are retained using the Seurat v3 flavor. This filtered, normalized expression matrix serves as the basis for all subsequent spatially variable gene selection, Leiden domain assignment, and model training. Table~\ref{tbl:qc} reports the resulting section sizes, gene counts, and median mitochondrial read percentage after this pipeline. FFPE sections show negligible mitochondrial content, consistent with RNA degradation during formalin fixation, while the cerebellum section retains markedly higher mitochondrial content than the fresh-frozen cancer sections, reflecting known differences in baseline mitochondrial activity between neural and tumor tissue.

\begin{table}[width=\columnwidth,cols=4,pos=t]
\caption{Post-QC dataset statistics.}\label{tbl:qc}
\small
\begin{tabular*}{\columnwidth}{@{\extracolsep{\fill}}p{2.1cm}p{1.3cm}p{1.3cm}p{1.3cm}@{}}
\toprule
Section & Spots & Genes & Median MT\% \\
\midrule
Breast S1 & 3,798 & 19,690 & 3.5\% \\
Breast S2 & 3,986 & 19,699 & 3.7\% \\
Breast FFPE & 2,518 & 14,963 & 0.0\% \\
Colorectal & 3,137 & 17,252 & 3.8\% \\
Prostate FFPE & 3,043 & 15,215 & 0.0\% \\
Human Cerebellum & 4,916 & 18,252 & 14.5\% \\
\bottomrule
\end{tabular*}
\end{table}

\subsection{Baselines and Implementation Details}
\label{sec:baselines}

We compare HierarchicalDAEW against thirteen published baselines spanning non-graph, transformer-based, and graph-based architectures for gene expression prediction from histology: \textit{STNet} \cite{he2020integrating}, \textit{HisToGene} \cite{pang2021leveraging}, \textit{Hist2ST} \cite{zeng2022spatial}, \textit{BLEEP} \cite{xie2023spatially}, \textit{EGGN} \cite{yang2023exemplar,yang2024spatial}, \textit{TRIPLEX} \cite{chung2024accurate}, \textit{SEPAL} \cite{mejia2023sepal}, \textit{MERGE} \cite{ganguly2025merge}, and \textit{THItoGene} \cite{jia2024thitogene}. These cover a range of design philosophies from simple feed-forward mappings to transformer-based encoders and graph-structured architectures. Alongside these, three general-purpose graph neural network baselines, \textit{GCN} \cite{kipf2016semi}, \textit{GAT} \cite{velivckovic2017graph}, and \textit{GraphSAGE} \cite{hamilton2017inductive}, together with a non-graph MLP, serve to isolate the contribution of graph structure itself, independent of any method-specific architectural choices such as exemplar retrieval, contrastive pretraining, or multi-resolution feature fusion that characterize the domain-specific baselines. All models are implemented using PyTorch Geometric \cite{fey2019fast}.

All baselines are reimplemented under a shared experimental protocol to ensure a fair and controlled comparison. Every model consumes identical UNI \cite{chen2024towards} patch embeddings as input features, is trained using the same optimizer, learning rate schedule, and early stopping criterion, and is evaluated on identical cross-validation splits. This shared protocol controls for confounds such as differing feature extractors, training budgets, or evaluation procedures, isolating architectural differences as the primary source of any observed performance variation between methods. Where a baseline's original architecture assumes graph structure, such as Hist2ST \cite{zeng2022spatial}, SEPAL \cite{mejia2023sepal}, EGGN \cite{yang2023exemplar}, TRIPLEX \cite{chung2024accurate}, and MERGE \cite{ganguly2025merge}, we retain its core structural design while substituting our shared feature extraction and training pipeline in place of the encoder or training regime described in its original publication. Where a baseline requires multi-resolution image inputs unavailable in our setting, such as TRIPLEX \cite{chung2024accurate}, we approximate the missing resolution using our single-scale UNI embeddings, preserving the architectural mechanism while adapting it to the input modality available across all sections in this study.

All models, including HierarchicalDAEW, are trained with the AdamW optimizer using a weight decay of $1 \times 10^{-4}$ and a cosine annealing learning rate schedule with a ten-epoch linear warmup, for a maximum of 500 epochs with early stopping triggered after 80 epochs without validation improvement. A learning rate of $5 \times 10^{-4}$ is applied uniformly across all baselines rather than tuned individually per model, following a learning rate sweep confirming that this value performs competitively across architectures of varying depth and parameter count, including HierarchicalDAEW itself. This uniform treatment avoids inadvertently favoring our proposed method through more extensive hyperparameter search than afforded to competing baselines, a common source of unfair comparison in empirical machine learning evaluations.

HierarchicalDAEW additionally uses a hidden dimension of 1024 throughout its spot and gene graph encoders, two stacked DAEWConv layers, and 4 nearest neighbors for spot graph construction, along with a Leiden resolution of $0.7$ and a contrastive temperature of $0.1$, all selected via ablation studies conducted prior to final evaluation (Section~\ref{sec:ablations}). The evidential and contrastive training objectives are weighted by $\lambda_{\mathrm{NIG}} = 0.01$ and $\lambda_{\mathrm{ctr}} = 0.1$ respectively. All experiments use a fixed random seed of 42 to ensure exact reproducibility of reported results, except where seed is explicitly varied to assess the robustness of our findings to initialization alone.

\subsection{Evaluation Protocol and Hyperparameter Selection}
\label{sec:evalprotocol}

We evaluated all models using $K$-fold cross-validation at the spot level, with $K=5$ for primary benchmark comparisons and $K=3$ for ablation studies and secondary analyses, reducing computational cost across the large number of ablation configurations while retaining a held-out test set in every run. Within each fold, spatially variable gene selection and Leiden domain assignment are recomputed independently using only the training split, so that no information from held-out spots influences gene selection, domain structure, or graph construction, and every reported metric is computed exclusively on spots unseen during that fold's training.

Primary accuracy is reported as the Pearson correlation coefficient (PCC) \cite{benesty2009pearson} between predicted and observed expression, computed per spot across the $G$ genes and averaged over all $N$ test spots,
\begin{equation}
    \mathrm{PCC} = \frac{1}{N} \sum_{i=1}^{N} \frac{\sum_{g=1}^{G} (\hat{y}_{ig} - \bar{\hat{y}}_i)(y_{ig} - \bar{y}_i)}{\sqrt{\sum_{g=1}^{G} (\hat{y}_{ig} - \bar{\hat{y}}_i)^2} \sqrt{\sum_{g=1}^{G} (y_{ig} - \bar{y}_i)^2}},
    \label{eq:pcc}
\end{equation}
alongside Spearman rank correlation \cite{zar2005spearman}, computed identically but over the rank-transformed expression vectors, Kendall's $\tau$ \cite{mcleod2005kendall}, and the coefficient of determination $R^2$, quantifying the proportion of variance in observed expression explained by the model's predictions. Statistical significance of pairwise comparisons against HierarchicalDAEW is assessed per fold, and effect size is reported using Cohen's $d$ \cite{cohen2013statistical}.

Calibration of the evidential uncertainty estimates is assessed using expected normalized calibration error \cite{kuleshov2018accurate},
\begin{equation}
    \mathrm{ENCE} = \frac{1}{M} \sum_{m=1}^{M} \frac{\left| \mathrm{RMSE}_m - \sqrt{\mathrm{Var}_m} \right|}{\sqrt{\mathrm{Var}_m}},
    \label{eq:ence}
\end{equation}
where predictions are grouped into $M$ bins by predicted variance, $\mathrm{RMSE}_m$ is the root mean squared error within bin $m$, and $\mathrm{Var}_m$ is the average predicted variance within that bin, together with empirical coverage at the 90\% confidence level, the fraction of observed values falling within the predicted 90\% credible interval.

Architectural and training hyperparameters, including the number of DAEWConv layers, neighborhood size $k$, hidden dimension, Leiden resolution, evidential loss weight, contrastive temperature, and STRING-DB confidence threshold, are selected via grid search on the primary breast section using 3-fold cross-validation, prior to and independent of the main 5-fold benchmark runs. Once selected, the same configuration is applied uniformly across all subsequent experiments to avoid tuning hyperparameters on data used for final reported comparisons.

\section{Results}
\label{sec:results}

\subsection{Multi-Section Joint Training}
\label{sec:multisection-results}

We first evaluate all methods under our primary experimental setting: joint training across three breast tissue sections (Breast S1, Breast S2, Breast FFPE), spanning 10,302 spots and two distinct sample preservation protocols \cite{ruiz2025completing}, using the 512 spatially variable genes common to all three sections and 3-fold cross-validation. This setting directly tests whether a model's learned representation generalizes across biological replicates and technical preparation differences within a single training run, rather than being fit to the idiosyncrasies of a single section.

Table~\ref{tbl:multisection} reports mean PCC, Spearman correlation \cite{zar2005spearman}, coefficient of determination ($R^2$), and statistical significance relative to HierarchicalDAEW from a paired test across folds. HierarchicalDAEW achieves the highest mean PCC (0.696), outperforming every baseline by a statistically significant margin ($p<0.01$ for all pairwise comparisons after Benjamini-Hochberg correction \cite{benjamini1995controlling}). The strongest baselines, THItoGene \cite{jia2024thitogene} and SEPAL \cite{mejia2023sepal}, both graph-based methods incorporating some form of spatial structure, trail HierarchicalDAEW by approximately 0.058 PCC, while non-graph baselines such as STNet \cite{he2020integrating} and the plain MLP fall further behind at approximately 0.071 PCC lower. Notably, HierarchicalDAEW also achieves the highest $R^2$ (0.592) by a wide margin, indicating that its predictions explain substantially more variance in observed expression than any competing method, and shows the lowest fold-to-fold standard deviation (0.003) among all graph-based methods, suggesting stable performance across different biological replicate compositions within each fold.

\begin{table*}[width=\textwidth,cols=5,pos=t]
\caption{Multi-section joint training results (Breast S1 + S2 + FFPE).}\label{tbl:multisection}
\small
\begin{tabular*}{\textwidth}{@{\extracolsep{\fill}}l c c c c@{}}
\toprule
Model & PCC & Spearman & $R^2$ & Sig. \\
\midrule
HierarchicalDAEW & \textbf{0.696} & \textbf{0.669} & \textbf{0.592} & -- \\
THItoGene \cite{jia2024thitogene}       & 0.638 & 0.617 & 0.509 & ** \\
SEPAL \cite{mejia2023sepal}             & 0.638 & 0.625 & 0.330 & ** \\
MERGE \cite{ganguly2025merge}           & 0.629 & 0.608 & 0.497 & ** \\
MLP                                      & 0.625 & 0.602 & 0.499 & ** \\
STNet \cite{he2020integrating}          & 0.625 & 0.602 & 0.497 & ** \\
GraphSAGE \cite{hamilton2017inductive}  & 0.622 & 0.602 & 0.493 & ** \\
Hist2ST \cite{zeng2022spatial}          & 0.620 & 0.602 & 0.371 & ** \\
TRIPLEX \cite{chung2024accurate}        & 0.618 & 0.597 & 0.496 & ** \\
GCN \cite{kipf2016semi}                 & 0.614 & 0.595 & 0.488 & ** \\
EGGN \cite{yang2024spatial}             & 0.604 & 0.590 & 0.004 & ** \\
BLEEP \cite{xie2023spatially}           & 0.599 & 0.576 & 0.467 & ** \\
HisToGene \cite{pang2021leveraging}     & 0.593 & 0.583 & 0.471 & ** \\
GAT \cite{velivckovic2017graph}         & 0.587 & 0.573 & 0.363 & ** \\
\bottomrule
\end{tabular*}
\end{table*}

\begin{center}
\includegraphics[width=\linewidth]{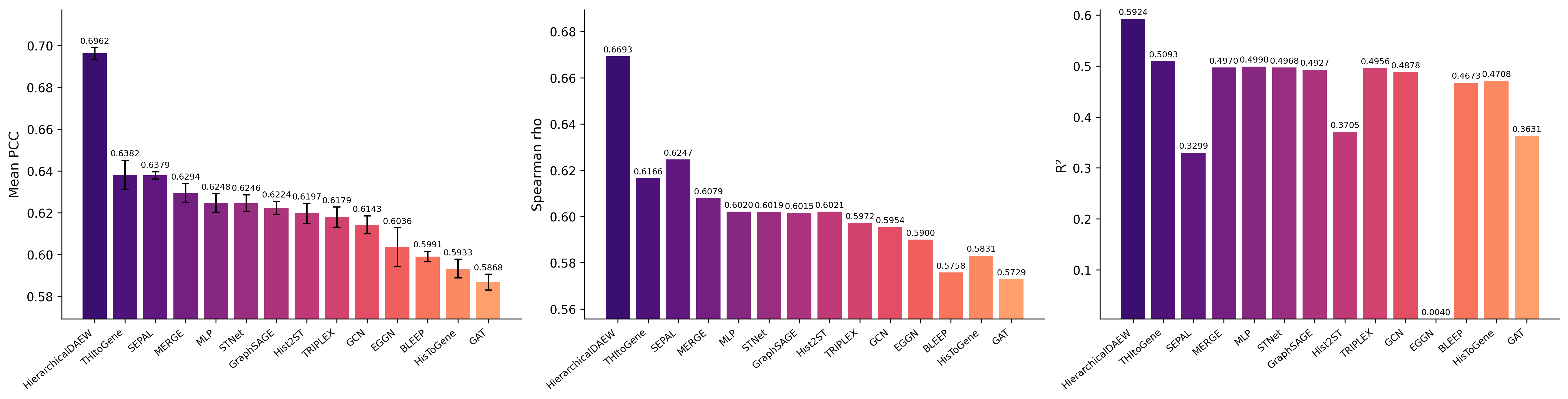}
\captionof{figure}{Multi-section joint training results on Breast S1 + S2 + FFPE (3-fold CV). HierarchicalDAEW achieves the highest mean PCC (0.696), Spearman correlation (0.669), and $R^2$ (0.592) of all fourteen methods, with error bars showing fold-to-fold standard deviation.}
\label{fig:multisection_results}
\end{center}

\subsection{Single-Section Benchmark}
\label{sec:single-section}

To directly compare against prior work under the standard single-section evaluation setting used throughout the spatial transcriptomics literature \cite{ruiz2025completing}, we additionally train and evaluate all methods on Breast S1 alone, using 5-fold cross-validation. Table~\ref{tbl:singlesection} reports the resulting correlation metrics.

HierarchicalDAEW achieves the highest mean PCC (0.704), followed closely by SEPAL \cite{mejia2023sepal} (0.699); this difference is not statistically significant under a paired test across folds ($p \geq 0.05$), unlike the multi-section setting where HierarchicalDAEW's advantage over all baselines, including SEPAL, is significant. HierarchicalDAEW and SEPAL are also the only two methods among the fourteen evaluated that are not significantly different from one another, with every other baseline trailing both by a statistically significant margin. This pattern is consistent with the intuition that SEPAL, itself a graph-based method incorporating local structure, is best positioned among prior work to compete with HierarchicalDAEW when trained on a single, relatively homogeneous section; the architectural advantage of domain-aware edge typing becomes more clearly separable once the model must additionally handle the added heterogeneity introduced by multiple sections and preparation protocols, as shown in the multi-section setting. HierarchicalDAEW nonetheless attains the highest $R^2$ (0.574) and lowest RMSE (0.529) among all methods, and shows the highest Top-20\%-Recovery (0.980), indicating that the genes with strongest expression signal are recovered most reliably by our model even where the aggregate PCC margin over SEPAL is narrow.

\begin{table*}[width=\textwidth,cols=6,pos=t]
\caption{Single-section benchmark results (Breast S1, 5-fold CV).}\label{tbl:singlesection}
\small
\begin{tabular*}{\textwidth}{@{\extracolsep{\fill}}l c c c c c@{}}
\toprule
Model & PCC & Spearman & CCC & $R^2$ & Sig. \\
\midrule
HierarchicalDAEW & \textbf{0.704} & \textbf{0.671} & 0.655 & \textbf{0.574} & -- \\
SEPAL \cite{mejia2023sepal}             & 0.699 & 0.664 & \textbf{0.656} & 0.567 & ns \\
THItoGene \cite{jia2024thitogene}       & 0.675 & 0.642 & 0.637 & 0.526 & ** \\
MERGE \cite{ganguly2025merge}           & 0.668 & 0.635 & 0.620 & 0.509 & ** \\
GraphSAGE \cite{hamilton2017inductive}  & 0.665 & 0.631 & 0.632 & 0.526 & ** \\
Hist2ST \cite{zeng2022spatial}          & 0.650 & 0.609 & 0.603 & 0.486 & ** \\
EGGN \cite{yang2024spatial}             & 0.649 & 0.607 & 0.605 & 0.483 & ** \\
GCN \cite{kipf2016semi}                 & 0.648 & 0.611 & 0.615 & 0.497 & ** \\
GAT \cite{velivckovic2017graph}         & 0.644 & 0.600 & 0.600 & 0.479 & ** \\
TRIPLEX \cite{chung2024accurate}        & 0.635 & 0.605 & 0.580 & 0.502 & ** \\
STNet \cite{he2020integrating}          & 0.626 & 0.593 & 0.588 & 0.466 & ** \\
MLP                                      & 0.625 & 0.591 & 0.586 & 0.463 & ** \\
HisToGene \cite{pang2021leveraging}     & 0.625 & 0.606 & 0.557 & --    & ** \\
BLEEP \cite{xie2023spatially}           & 0.607 & 0.575 & 0.586 & --    & ** \\
\bottomrule
\end{tabular*}
\end{table*}

\begin{center}
\includegraphics[width=\linewidth]{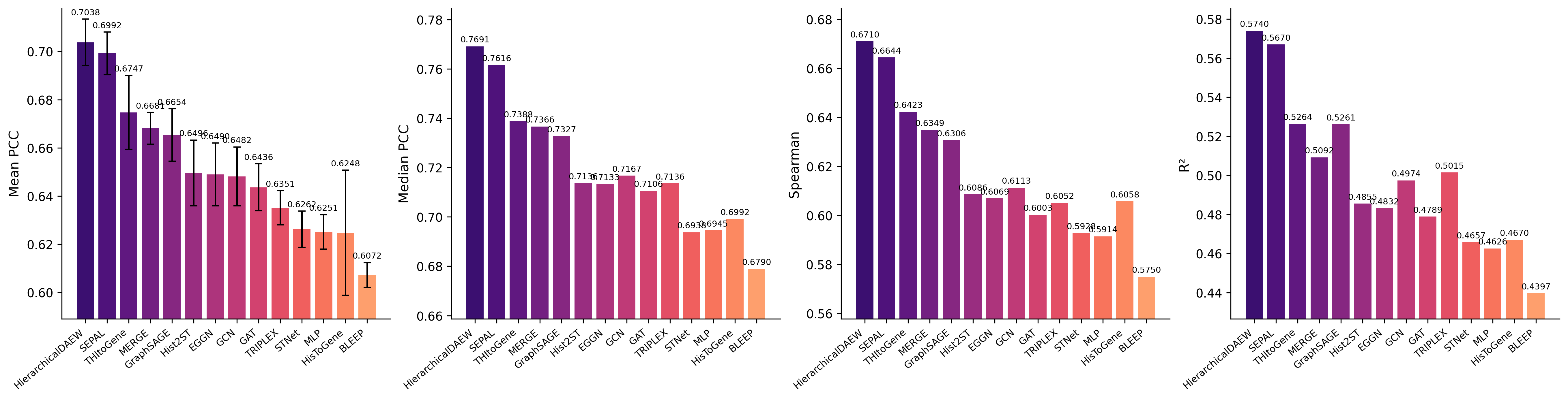}
\captionof{figure}{Single-section benchmark on Breast S1 (5-fold CV): mean PCC, median PCC, Spearman correlation \cite{zar2005spearman}, and $R^2$. HierarchicalDAEW and SEPAL \cite{mejia2023sepal} are statistically indistinguishable here, unlike in the multi-section setting.}
\label{fig:singlesection_a}
\end{center}

\begin{center}
\includegraphics[width=\linewidth]{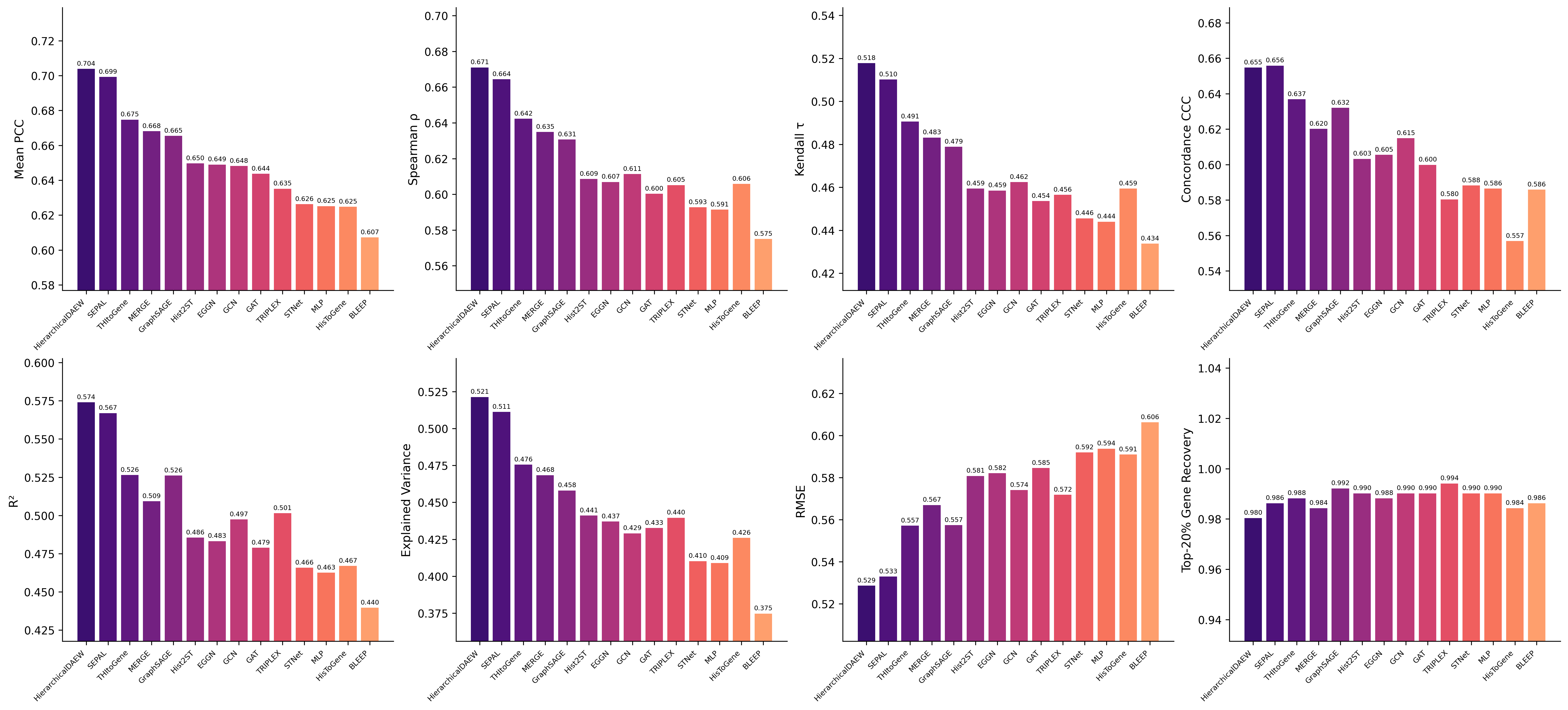}
\captionof{figure}{Extended single-section metrics on Breast S1: Spearman \cite{zar2005spearman}, Kendall's $\tau$ \cite{mcleod2005kendall}, concordance correlation coefficient, $R^2$, explained variance, RMSE, and Top-20\% gene recovery. HierarchicalDAEW attains the lowest RMSE and highest Top-20\% recovery among all baselines.}
\label{fig:singlesection_b}
\end{center}

\subsection{Cross-Tissue Generalization}
\label{sec:crosstissue}

To assess whether HierarchicalDAEW's architecture generalizes beyond breast tissue, we evaluate it, without any baseline comparison, on three additional Visium sections spanning distinct tissue types and biological contexts: colorectal cancer, prostate acinar cell carcinoma, and healthy human cerebellum, each under identical 5-fold cross-validation and hyperparameter settings selected on breast tissue alone. This experiment specifically tests generalization of the trained architecture and its hyperparameters, rather than raw single-tissue predictive ceiling, since no per-tissue retuning is performed.

HierarchicalDAEW attains a mean PCC of 0.495 on colorectal cancer, 0.488 on prostate FFPE, and 0.616 on human cerebellum. Performance on cerebellum tissue notably exceeds that observed on either cancer section, and is comparable to the breast tissue results reported earlier, despite cerebellum being structurally and biologically distinct from all three breast sections used for hyperparameter selection. Colorectal and prostate sections, by contrast, show a meaningful drop relative to breast tissue, with prostate FFPE showing the lowest mean PCC among all six sections evaluated in this work. Colorectal additionally shows the highest fold-to-fold variance observed in this study (PCC standard deviation 0.036, several times larger than any other section), coinciding with substantial variation in the number of Leiden domains detected across folds, suggesting that instability in domain structure contributes to less consistent performance on this tissue. Cerebellum, in contrast, achieves the second-lowest fold-to-fold variance (0.007) among all six sections, indicating that HierarchicalDAEW's predictions on healthy neural tissue are both accurate and highly consistent across folds despite the tissue never appearing during hyperparameter selection.

\begin{table}[width=\columnwidth,cols=2,pos=t]
\caption{Cross-tissue generalization results (HierarchicalDAEW only, 5-fold CV).}\label{tbl:crosstissue}
\small
\begin{tabular*}{\columnwidth}{@{\extracolsep{\fill}}lc@{}}
\toprule
Section & PCC (mean $\pm$ std) \\
\midrule
Colorectal        & 0.495 $\pm$ 0.036 \\
Prostate FFPE     & 0.488 $\pm$ 0.006 \\
Human Cerebellum  & 0.616 $\pm$ 0.007 \\
\bottomrule
\end{tabular*}
\end{table}
\begin{center}
\includegraphics[width=0.95\linewidth]{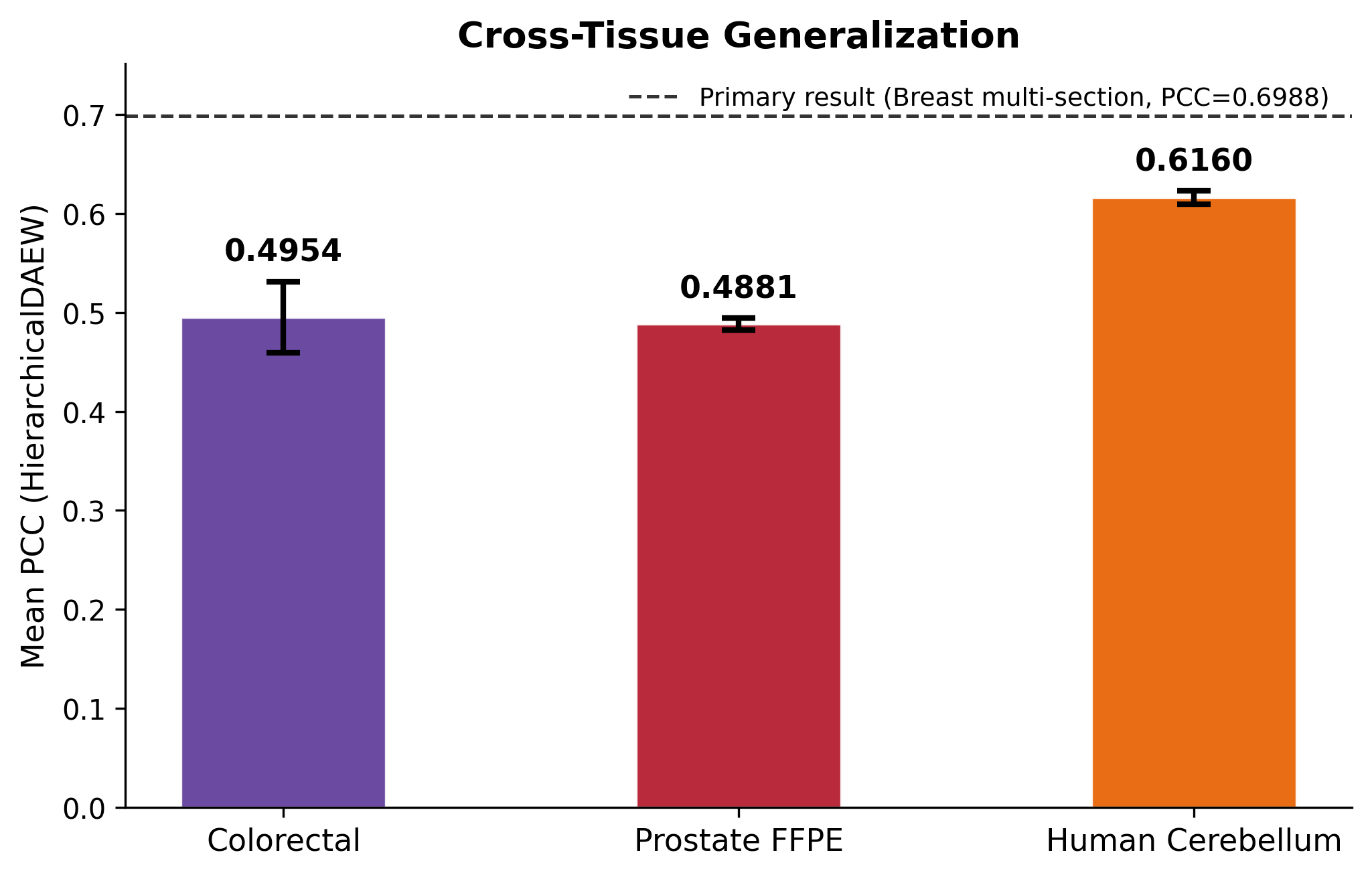}
\captionof{figure}{Cross-tissue generalization of HierarchicalDAEW, trained on breast tissue and evaluated without retuning on colorectal, prostate FFPE, and human cerebellum. Cerebellum performance approaches the in-distribution breast result despite no exposure during training.}
\label{fig:crosstissue}
\end{center}

\subsection{Leave-One-Dataset-Out and Few-Shot Adaptation}
\label{sec:lodo}

To assess whether HierarchicalDAEW's learned representations transfer to entirely unseen sections rather than merely interpolating within a training distribution, we conduct two complementary transfer experiments across the three breast sections. In the leave-one-dataset-out (LODO) setting, the model is trained jointly on two of the three sections and evaluated, without any fine-tuning, on the third, held-out section. Training on Breast S1 and FFPE and testing on Breast S2 yields a mean PCC of 0.427, while training on Breast S2 and FFPE and testing on Breast S1 yields a mean PCC of 0.481. Both results represent a substantial drop relative to the within-distribution results reported earlier, indicating that zero-shot transfer to a held-out section, even of the same tissue type and cancer indication, remains considerably more difficult than prediction on sections seen during training.

To characterize how quickly this gap closes with limited target-domain supervision, we conduct a bidirectional few-shot adaptation experiment between Breast S1 and Breast S2: a model trained on one section is fine-tuned using an increasing fraction of spots from the other, ranging from 1\% to 50\% of the target section. From S1 to S2, zero-shot PCC of 0.426 improves to 0.482 with 1\% fine-tuning data, reaching 0.629 at 25\% and plateauing near 0.630 at 50\%. From S2 to S1, zero-shot PCC of 0.498 improves more sharply, reaching 0.703 at 25\% and 0.701 at 50\%. In both directions, the majority of the achievable improvement is realized by 25\% fine-tuning data, with negligible additional gain from 25\% to 50\%, suggesting that HierarchicalDAEW's representations require only a modest amount of target-section supervision to adapt effectively, rather than needing to be retrained from scratch on each new section \cite{niu2025spabatch}.

\begin{table}[width=\columnwidth,cols=4,pos=t]
\caption{Leave-one-dataset-out results.}\label{tbl:lodo}
\small
\begin{tabular*}{\columnwidth}{@{\extracolsep{\fill}}llcc@{}}
\toprule
Train & Test & PCC & Spearman \\
\midrule
S1 + FFPE & S2 & 0.427 & 0.420 \\
S2 + FFPE & S1 & 0.481 & 0.459 \\
\bottomrule
\end{tabular*}
\end{table}

\begin{table}[width=\columnwidth,cols=3,pos=t]
\caption{Bidirectional few-shot adaptation (PCC).}\label{tbl:fewshot}
\small
\begin{tabular*}{\columnwidth}{@{\extracolsep{\fill}}lcc@{}}
\toprule
Fine-tune fraction & S1 $\to$ S2 & S2 $\to$ S1 \\
\midrule
0\% (zero-shot) & 0.426 & 0.498 \\
1\%             & 0.482 & 0.480 \\
5\%             & 0.493 & 0.617 \\
10\%            & 0.557 & 0.655 \\
25\%            & 0.629 & 0.703 \\
50\%            & 0.630 & 0.701 \\
\bottomrule
\end{tabular*}
\end{table}

\begin{center}
\includegraphics[width=0.95\linewidth]{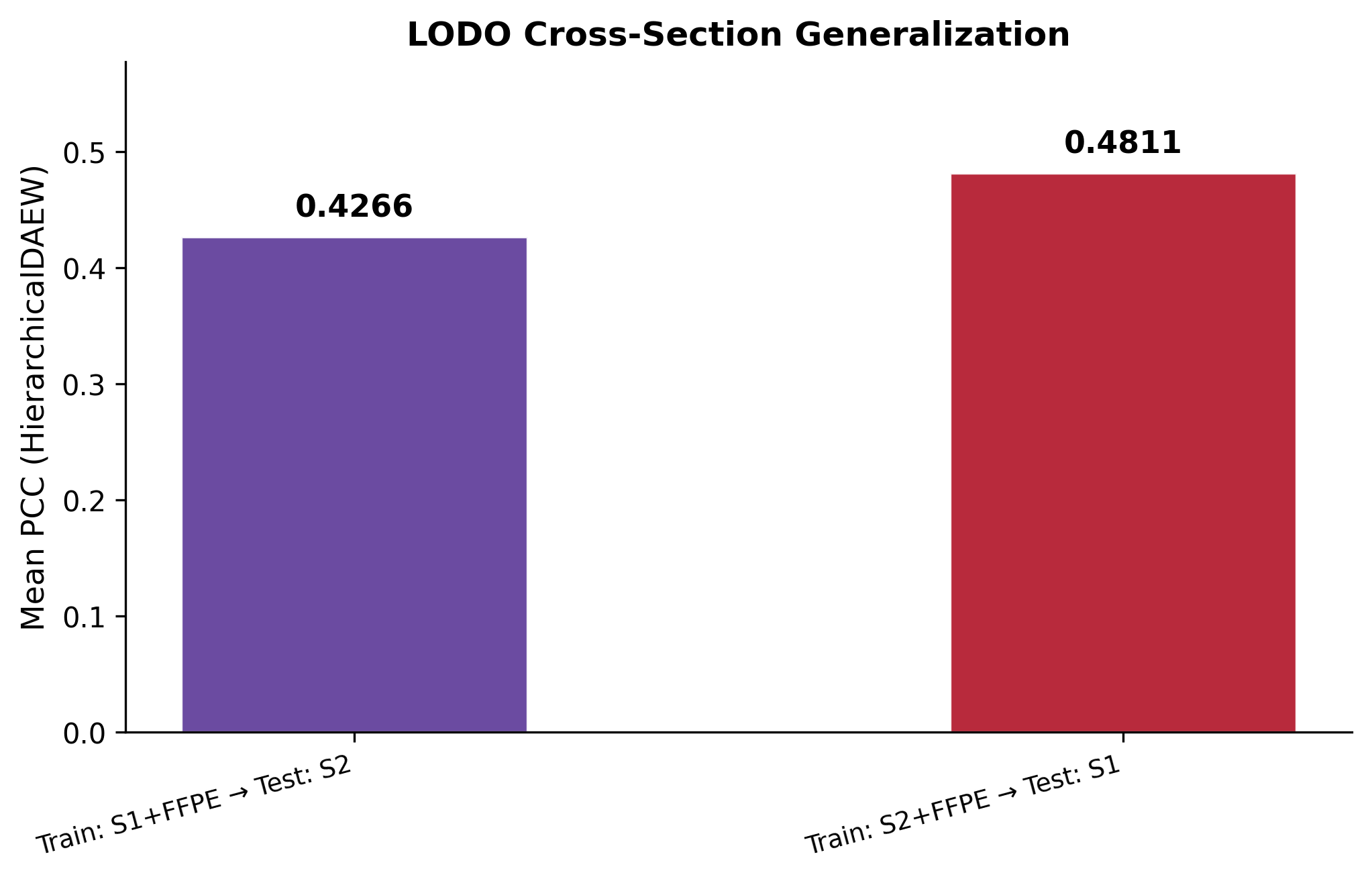}
\captionof{figure}{Leave-one-dataset-out (LODO) evaluation: mean PCC when jointly training on two breast sections and testing, without fine-tuning, on the third held-out section.}
\label{fig:lodo}
\end{center}
\begin{center}
\includegraphics[width=\linewidth]{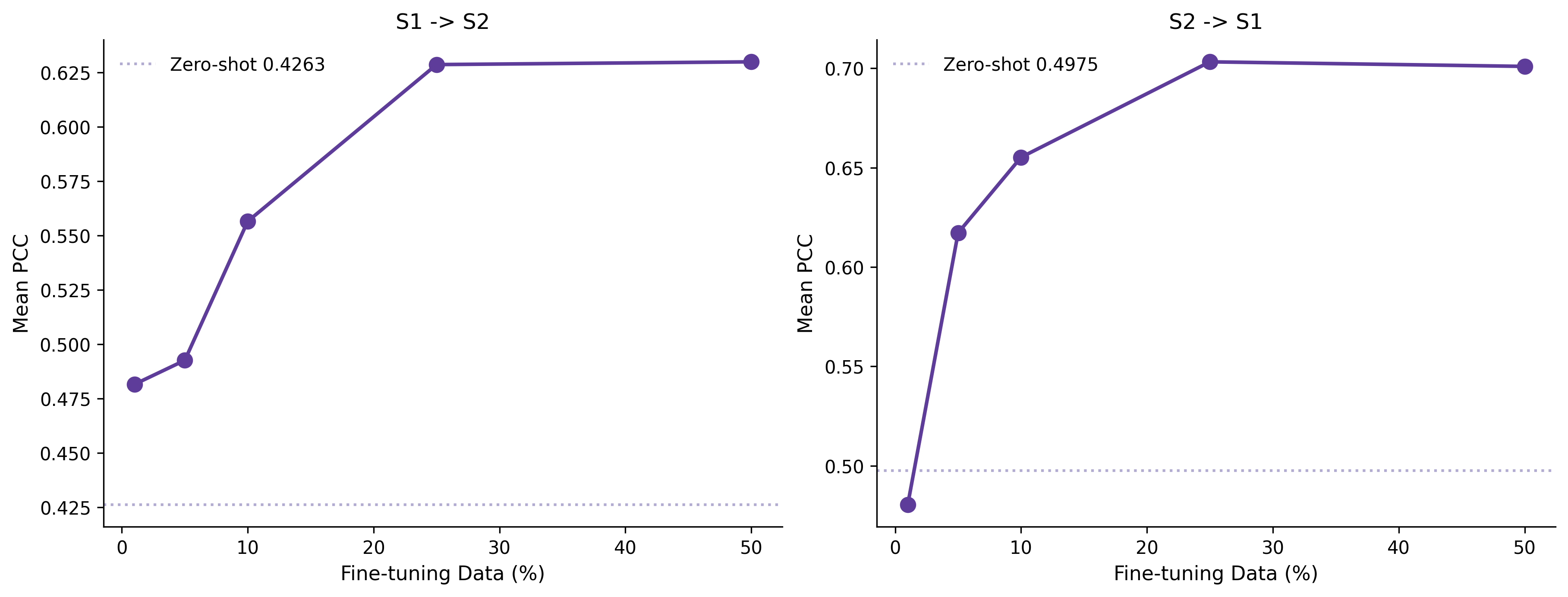}
\captionof{figure}{Bidirectional few-shot adaptation between Breast S1 and S2. Most of the achievable PCC gain over the zero-shot baseline is realized with only 25\% of target-section spots used for fine-tuning.}
\label{fig:fewshot}
\end{center}

\section{Analysis}
\label{sec:analysis}

\subsection{Ablation Studies}
\label{sec:ablations}

\subsubsection{Encoder, GNN Depth, Neighbourhood Size, Hidden Dimension}
\label{sec:ablation-arch}

We conduct a sequence of greedy ablation studies on a held-out subset of breast tissue, using a reduced 3-fold cross-validation protocol and a smaller gene panel to permit exhaustive search across a wide range of architectural choices. Each ablation is run in sequence, with the best-performing value fixed before proceeding to the next. Because this setup uses fewer genes and a shorter training budget than our main evaluation, absolute PCC values reported in this subsection should be interpreted only in relative terms, as a ranking among candidate configurations, rather than compared directly against the benchmark results reported earlier.

Among the three candidate histology encoders, UNI \cite{chen2024towards} achieves the highest mean PCC (0.743), outperforming both ResNet-50 \cite{he2016deep} (0.718) and Prov-GigaPath \cite{xu2024whole} (0.737), confirming that domain-specific histopathology pretraining provides a meaningful advantage over general-purpose ImageNet features, and that our largest foundation-model encoder does not necessarily outperform a comparatively smaller model trained specifically on histology. UNI is retained as the encoder for all subsequent experiments.

Varying the number of stacked DAEWConv layers $L \in \{1,2,3,5,7,10\}$ reveals a clear inverted-U relationship: performance peaks at $L=2$ (PCC = 0.769) and degrades steadily with additional depth, falling to 0.737 at $L=10$. This is consistent with the well-documented oversmoothing behavior of deep graph neural networks \cite{kipf2016semi}, in which repeated message passing causes node representations to converge toward indistinguishable values as depth increases; a shallow architecture is sufficient to propagate information across the relevant spatial neighborhood in this setting, while deeper stacks primarily add oversmoothing risk rather than useful receptive field.

Varying the spatial neighborhood size $K$ from the options in $\{4,6,8,10,14,18,24\}$ shows a comparatively flat response surface, with all values achieving PCC between 0.764 and 0.771; the small neighborhood $K=4$ performs marginally best (0.771), suggesting that gene expression prediction in this setting benefits from a tightly local receptive field rather than aggregating over a broad spatial neighborhood, though the narrow spread across all tested values indicates the model is not highly sensitive to this choice.

Hidden dimension shows the largest and most monotonic effect among the four ablations, with PCC increasing from 0.678 at $h=128$ to 0.782 at $h=1024$, a gain of over 0.10 PCC. Unlike depth, which exhibits a clear optimum, representational capacity continues to help up to the largest value tested; we did not extend the search beyond $h=1024$ due to computational cost, and it remains possible that larger hidden dimensions would yield further, likely diminishing, gains.

\begin{table*}[width=\textwidth,cols=4,pos=t]
\caption{Architecture ablation results (ablation subset, PCC).}\label{tbl:archablation}
\small
\begin{tabular*}{\textwidth}{@{\extracolsep{\fill}}lccl@{}}
\toprule
Ablation & Best & PCC & Range Tested \\
\midrule
Encoder      & UNI \cite{chen2024towards}  & 0.743 & ResNet \cite{he2016deep}, UNI, GigaPath \cite{xu2024whole} \\
Depth ($L$)  & 2    & 0.769 & 1, 2, 3, 5, 7, 10 \\
Neighbors ($K$) & 4 & 0.771 & 4, 6, 8, 10, 14, 18, 24 \\
Hidden dim   & 1024 & 0.782 & 128, 256, 512, 1024 \\
\bottomrule
\end{tabular*}
\end{table*}
\begin{center}
\includegraphics[width=0.7\linewidth]{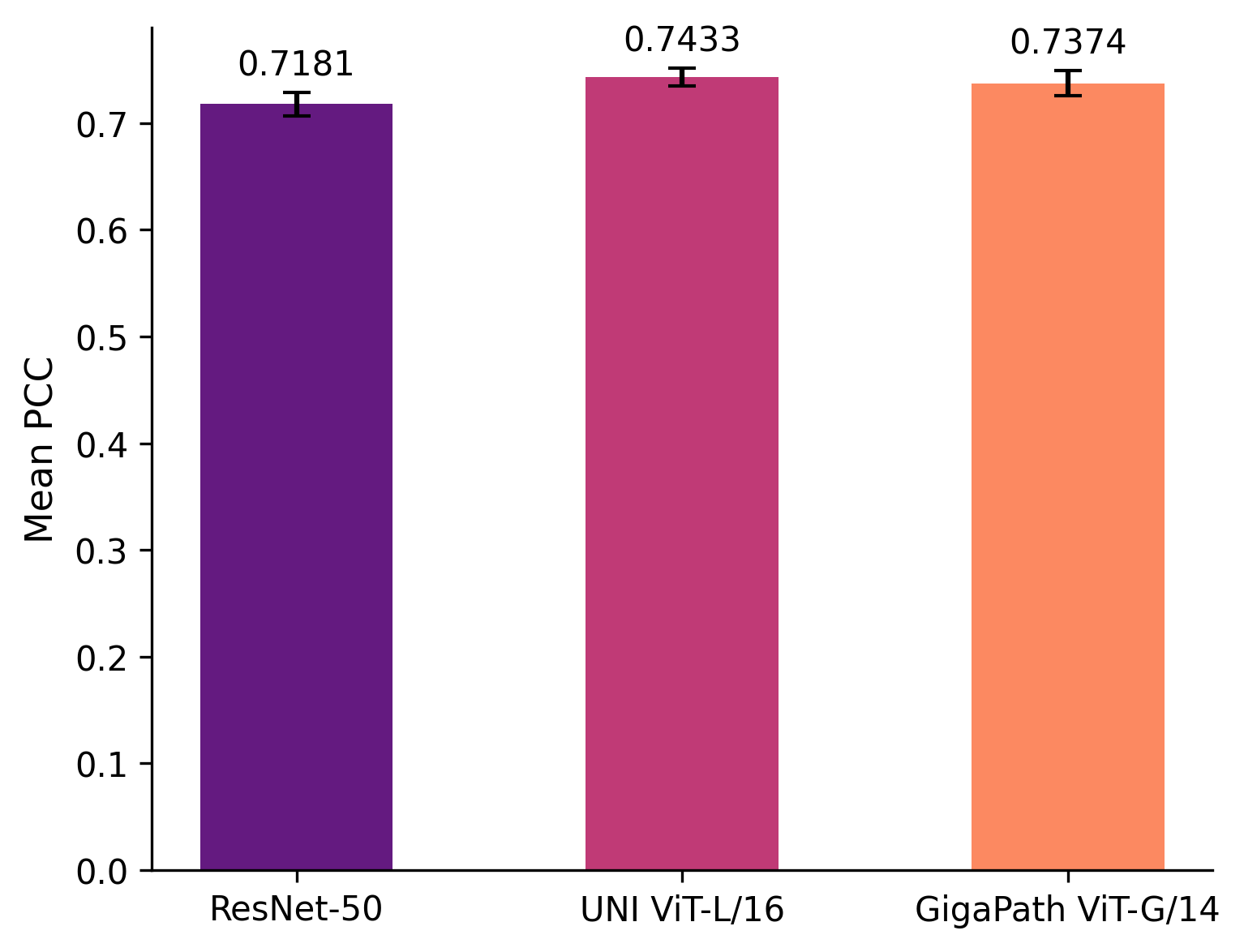}
\captionof{figure}{Histology encoder ablation. UNI \cite{chen2024towards} outperforms both ResNet-50 \cite{he2016deep} and Prov-GigaPath \cite{xu2024whole}, confirming that histopathology-specific pretraining matters more than encoder scale.}
\label{fig:ablation_encoder}
\end{center}
\begin{center}
\includegraphics[width=0.7\linewidth]{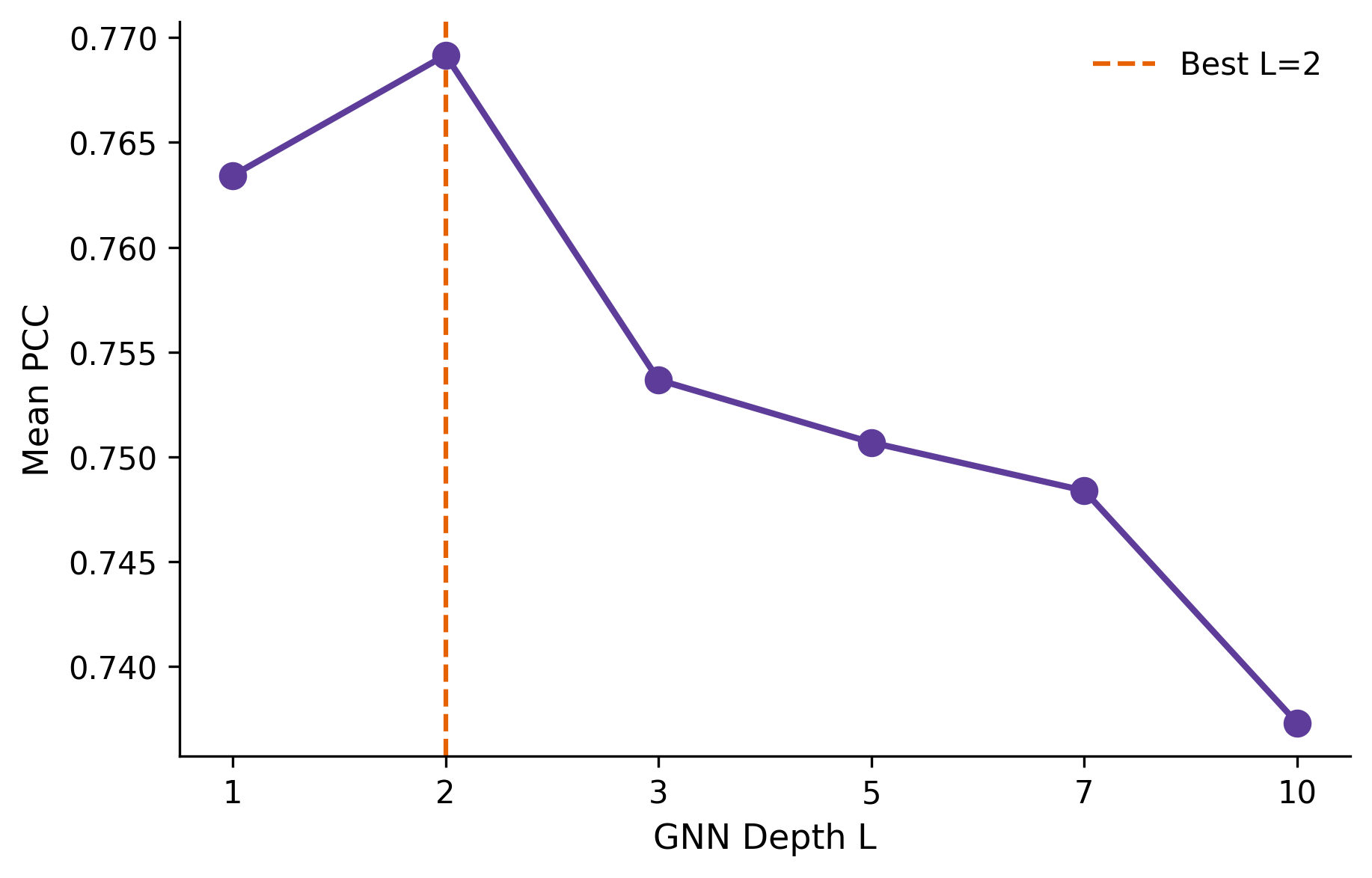}
\captionof{figure}{GNN depth ablation. Performance peaks at $L{=}2$ DAEWConv layers and degrades steadily with additional depth, consistent with oversmoothing in deeper graph networks \cite{kipf2016semi}.}
\label{fig:ablation_depth}
\end{center}
\begin{center}
\includegraphics[width=0.7\linewidth]{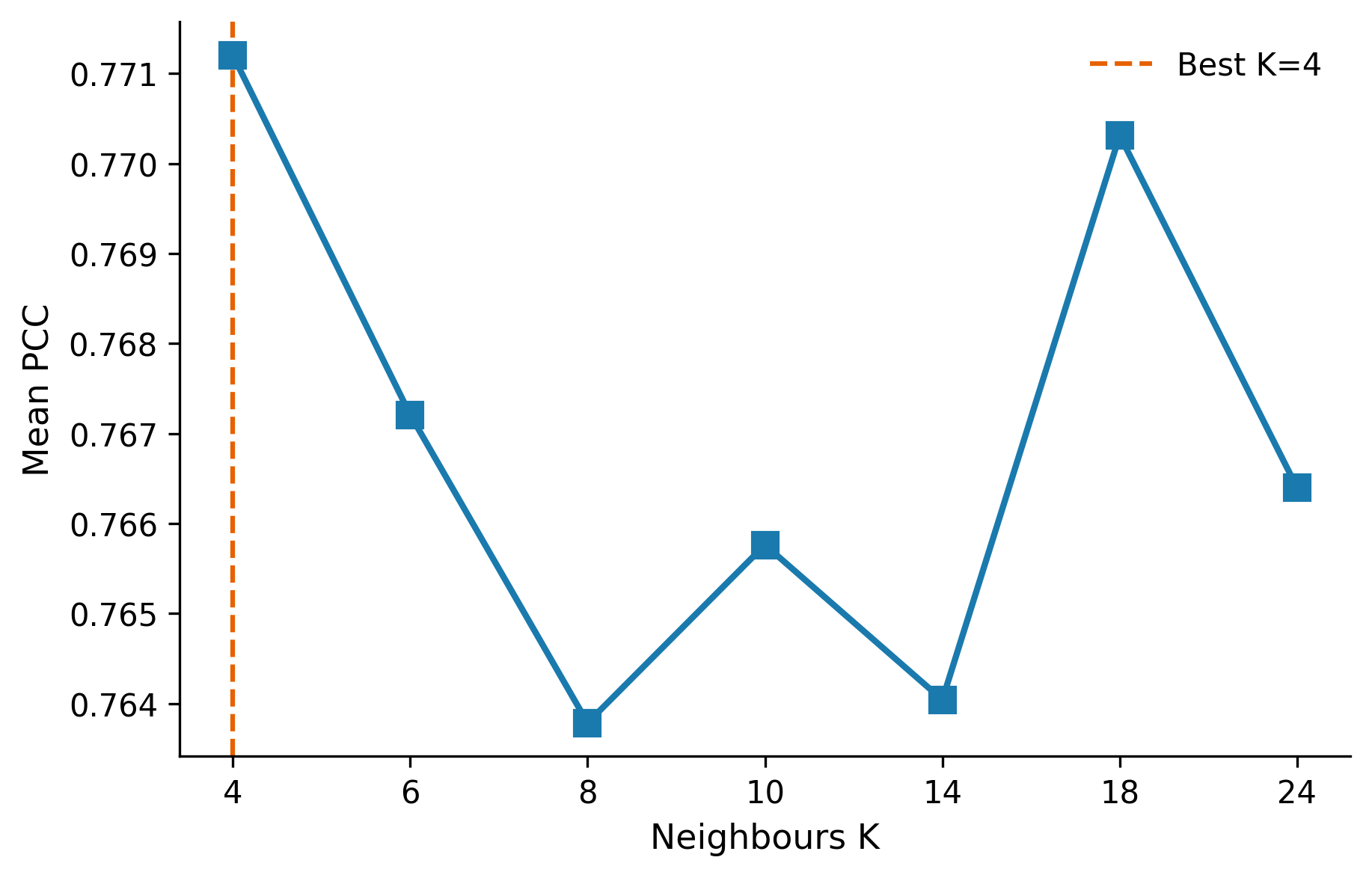}
\captionof{figure}{Spatial neighborhood size ablation. Accuracy is comparatively flat across $K$, with the smallest tested neighborhood ($K{=}4$) performing marginally best.}
\label{fig:ablation_neighborhood}
\end{center}
\begin{center}
\includegraphics[width=0.7\linewidth]{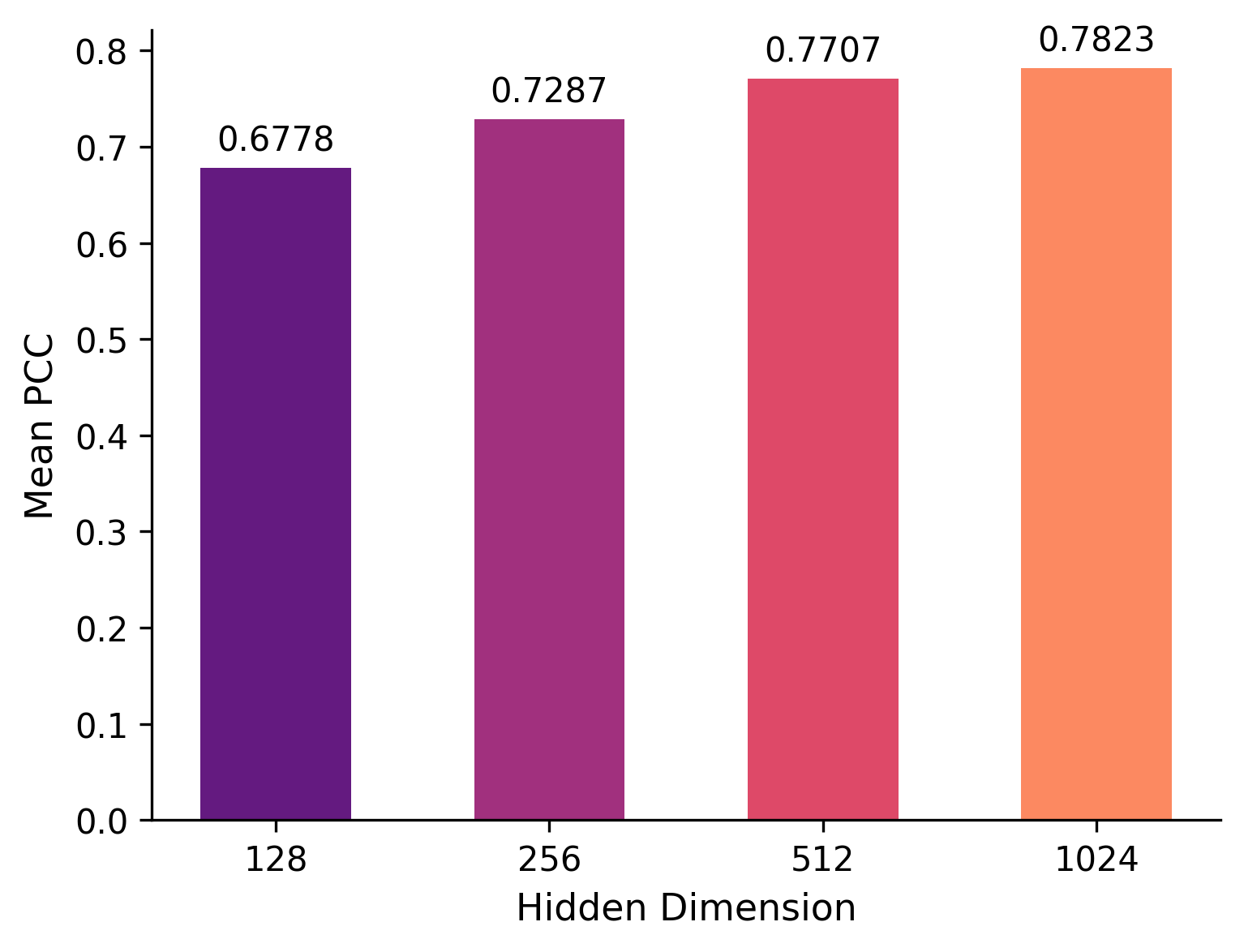}
\captionof{figure}{Hidden dimension ablation. Unlike depth, accuracy increases monotonically with hidden dimension up to the largest value tested ($h{=}1024$).}
\label{fig:ablation_hidden}
\end{center}

\subsubsection{Hierarchy vs Flat, Fixed vs Learned Alphas, Contrastive Loss}

We next isolate the three architectural and training choices within the HierarchicalDAEW, using the same ablation protocol.

First, we compare the full hierarchical architecture, incorporating DomainGCN and CrossScaleGate, against a flat variant that stacks the same number of DAEWConv layers without any domain-level pooling or cross-scale fusion. The hierarchical variant achieves a substantially higher mean PCC (0.769) than the flat variant (0.725) scoring a gap of 0.044 PCC, the largest margin observed among the sections in the ablations. This confirms that the domain-level reasoning introduced by DomainGCN and CrossScaleGate contributes meaningfully beyond what the stacked local convolution can capture, supporting hierarchy as a necessary rather than incidental component of the architecture.

Second, we compare fixed, uniformly weighted edge type-contributions against the learnable per-type gates $\alpha_t^{(l)}$ used throughout the main model. The learned-gate variant achieves a marginally higher mean PCC (0.711) as compared to the fixed-weight variant (0.709), a difference of only 0.002 PCC. While directionally consistent with our design choice, this gap is small relative to other ablations in this subsection, suggesting that the primary benefit of DAEWConv derives from having separate per-type projection matrices $W_t^{(l)}$ at all, rather than from the specific mechanism used to weigh their relative contribution.

Third, we ablate the domain contrastive loss $\mathcal{L}_{\mathrm{ctr}}$ \cite{xie2023spatially}, holding all other objectives fixed. Removing it changes mean PCC by only 0.002 (0.768 without versus 0.766 with), well within fold-to-fold variability. This is expected: $\mathcal{L}_{\mathrm{ctr}}$ is designed to shape the geometry of learned representations, pulling same-domain spots together and separating spots across domain boundaries, rather than to directly improve point-prediction accuracy. Its effect on representation structure is evaluated later on.

\begin{table}[width=\columnwidth,cols=3,pos=t]
\caption{Hierarchy, edge-weight, and contrastive loss ablations (ablation subset, PCC).}\label{tbl:hierablation}
\small
\begin{tabular*}{\columnwidth}{@{\extracolsep{\fill}}llc@{}}
\toprule
Ablation & Variant & Mean PCC \\
\midrule
Architecture      & Flat DAEW         & 0.725 \\
Architecture      & Hierarchical DAEW & \textbf{0.769} \\
Edge gate         & Fixed             & 0.709 \\
Edge gate         & Learned           & \textbf{0.711} \\
Contrastive loss  & Without           & \textbf{0.768} \\
Contrastive loss  & With              & 0.766 \\
\bottomrule
\end{tabular*}
\end{table}

\begin{center}
\includegraphics[width=\linewidth]{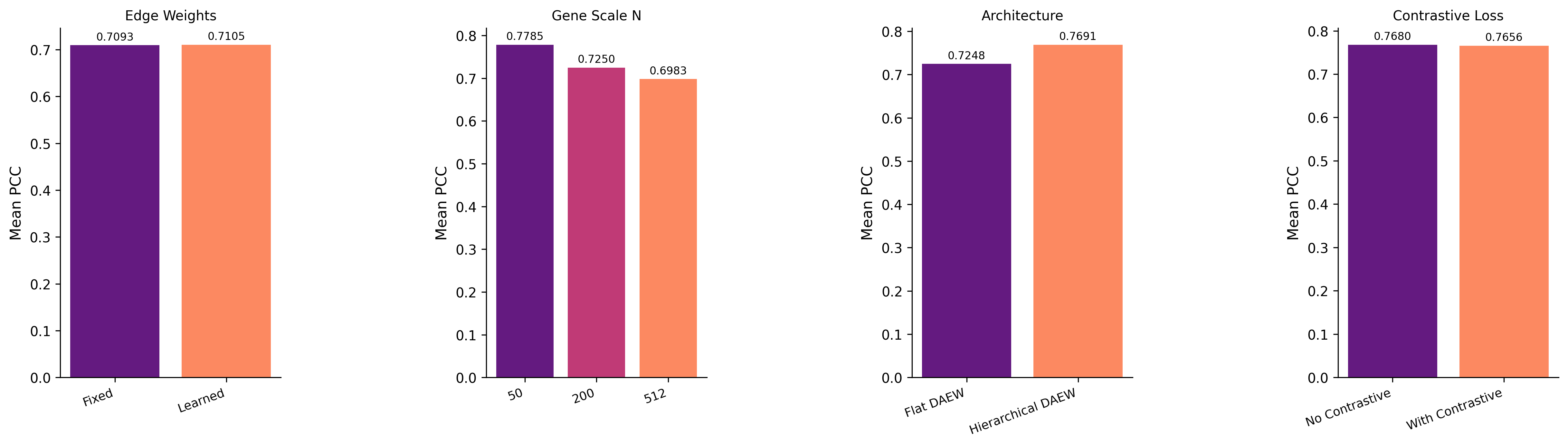}
\captionof{figure}{Architectural and training ablations: fixed vs.\ learned edge gates, gene panel scale, flat vs.\ hierarchical architecture, and with/without contrastive loss \cite{xie2023spatially}. The hierarchical architecture yields the largest single gain (+0.044 PCC) among these four ablations.}
\label{fig:ablation_hierarchy}
\end{center}

\subsubsection{Leiden Resolution, Contrastive Temperature, STRING-DB Threshold}

We ablate three thresholds governing graph construction: the Leiden clustering resolution used for tissue domain assignment \cite{hu2021spagcn}, the temperature parameter of the contrastive loss \cite{xie2023spatially}, and the confidence threshold used to filter STRING-DB protein-protein interaction edges in the gene graph \cite{zhang2021graph}.

Varying Leiden resolutions over $\{0.3, 0.5, 0.7, 1.0, 1.5\}$ shows a clear unimodal pattern, with performance rising from 0.776 at resolution 0.3 to a peak of 0.785 at resolution 0.7, before declining to 0.768 at resolution 1.5. Coarser clustering (lower resolution, fewer domains) appears to under-segment tissue structure, merging biologically distinct regions into a single domain, while finer clustering (higher resolution, more domains) likely fragments coherent regions into spuriously separate domains, diluting the edge-typing signals DAEWConv relies on. Resolution 0.7, corresponding to 8--20 domains per section depending on tissue heterogeneity, is retained for all subsequent experiments.

Varying contrastive temperature over $\{0.01, 0.05, 0.07, 0.1, 0.3\}$ shows a comparatively flat response, with PCC ranging narrowly from 0.783 to 0.788; performance increases modestly as temperature rises from 0.01 to 0.1 then drops at 0.3. Temperature 0.1 is selected, though the narrow spread across all tested values indicates the contrastive loss is not sensitive to this choice, which is consistent with its comparatively small overall effect on prediction accuracy.

Varying the STRING-DB threshold over $\{500, 700, 900\}$, controlling how strictly protein-protein interactions must be supported by evidence before being included as gene graph edges, shows a small monotonic improvement with stricter filtering. The effect is negligible in absolute terms, suggesting that gene graph performance in this setting is relatively insensitive to STRING-DB edge density within the range tested; we retain the strictest threshold (900) as it admits only the highest-confidence interactions without measurably harming performance.

\begin{table}[width=\columnwidth,cols=3,pos=t]
\caption{Graph construction threshold ablations (ablation subset, PCC).}\label{tbl:thresholdablation}
\small
\begin{tabular*}{\columnwidth}{@{\extracolsep{\fill}}llc@{}}
\toprule
Parameter & Value & Mean PCC \\
\midrule
Leiden resolution        & 0.3  & 0.776 \\
Leiden resolution        & 0.5  & 0.782 \\
Leiden resolution        & \textbf{0.7}  & \textbf{0.785} \\
Leiden resolution        & 1.0  & 0.778 \\
Leiden resolution        & 1.5  & 0.768 \\
Contrastive temperature  & 0.01 & 0.785 \\
Contrastive temperature  & 0.05 & 0.787 \\
Contrastive temperature  & 0.07 & 0.787 \\
Contrastive temperature  & \textbf{0.1}  & \textbf{0.788} \\
Contrastive temperature  & 0.3  & 0.783 \\
STRING-DB threshold      & 500  & 0.7788 \\
STRING-DB threshold      & 700  & 0.7793 \\
STRING-DB threshold      & \textbf{900}  & \textbf{0.7794} \\
\bottomrule
\end{tabular*}
\end{table}
\begin{center}
\includegraphics[width=0.7\linewidth]{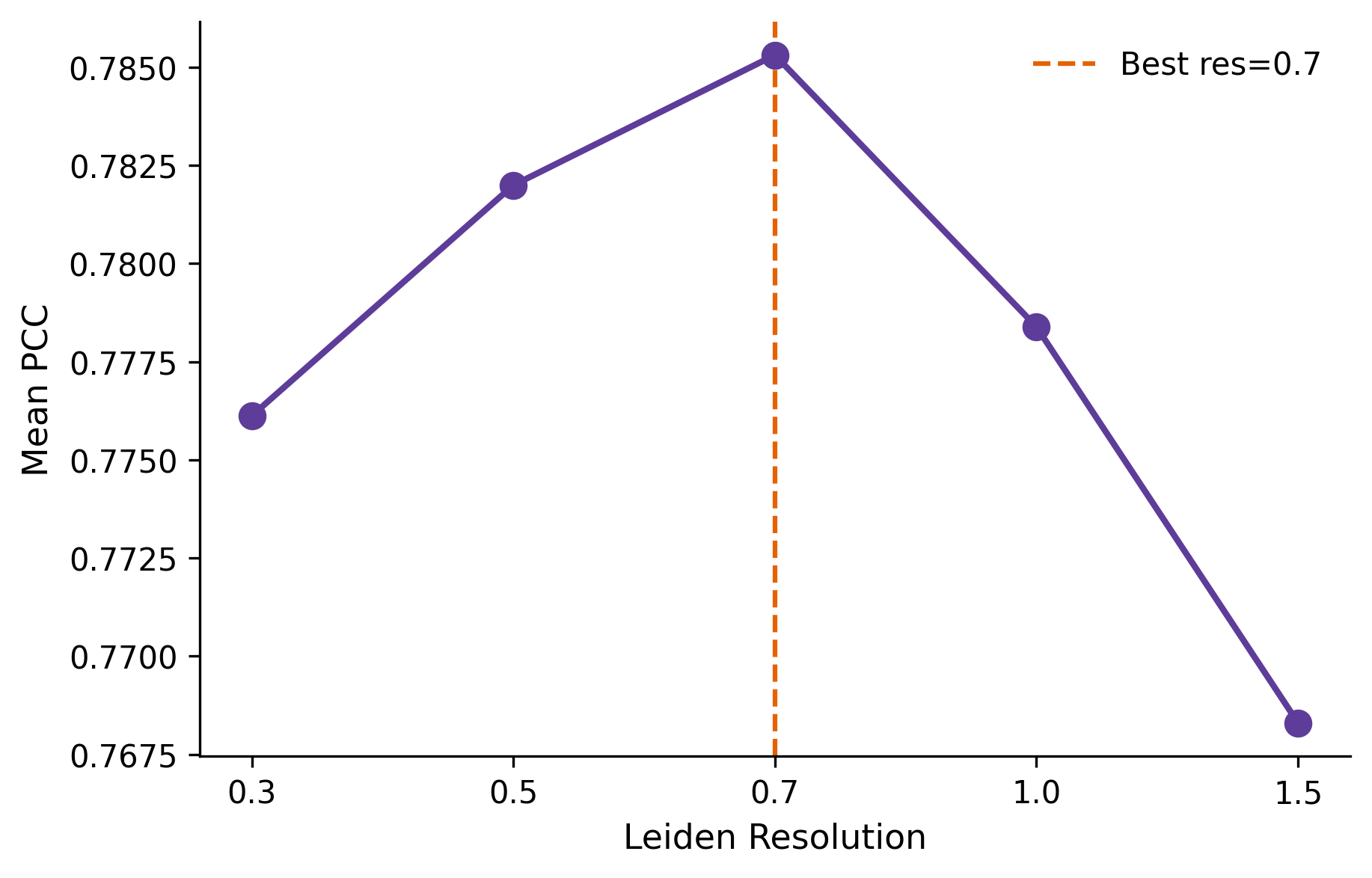}
\captionof{figure}{Leiden clustering resolution ablation \cite{hu2021spagcn}. Performance follows a clear unimodal trend, peaking at resolution 0.7.}
\label{fig:ablation_leiden}
\end{center}
\begin{center}
\includegraphics[width=0.7\linewidth]{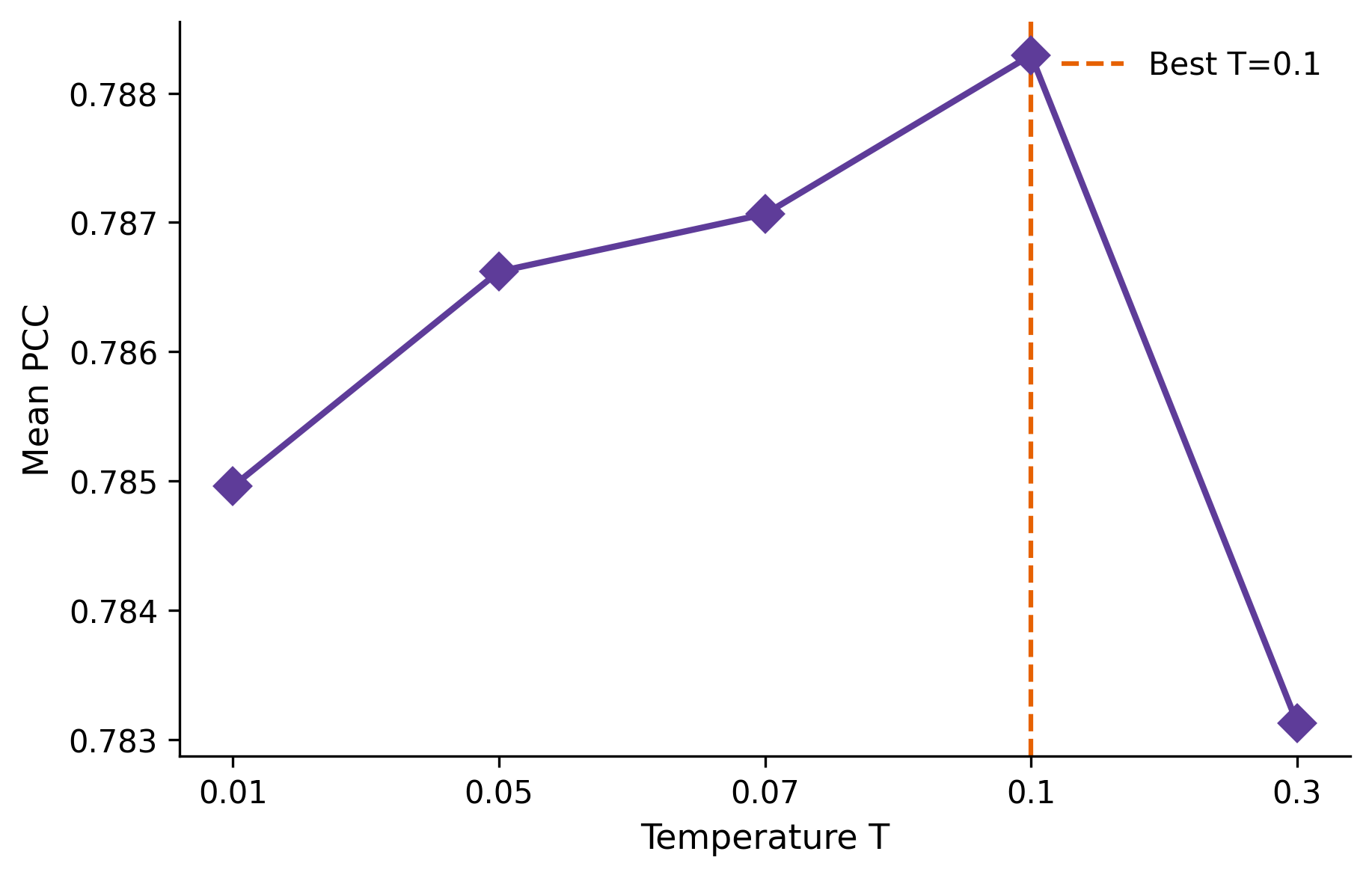}
\captionof{figure}{Contrastive loss temperature ablation \cite{xie2023spatially}. Accuracy is largely insensitive to $\tau$ within the tested range, with $\tau{=}0.1$ marginally best.}
\label{fig:ablation_temp}
\end{center}
\begin{center}
\includegraphics[width=0.7\linewidth]{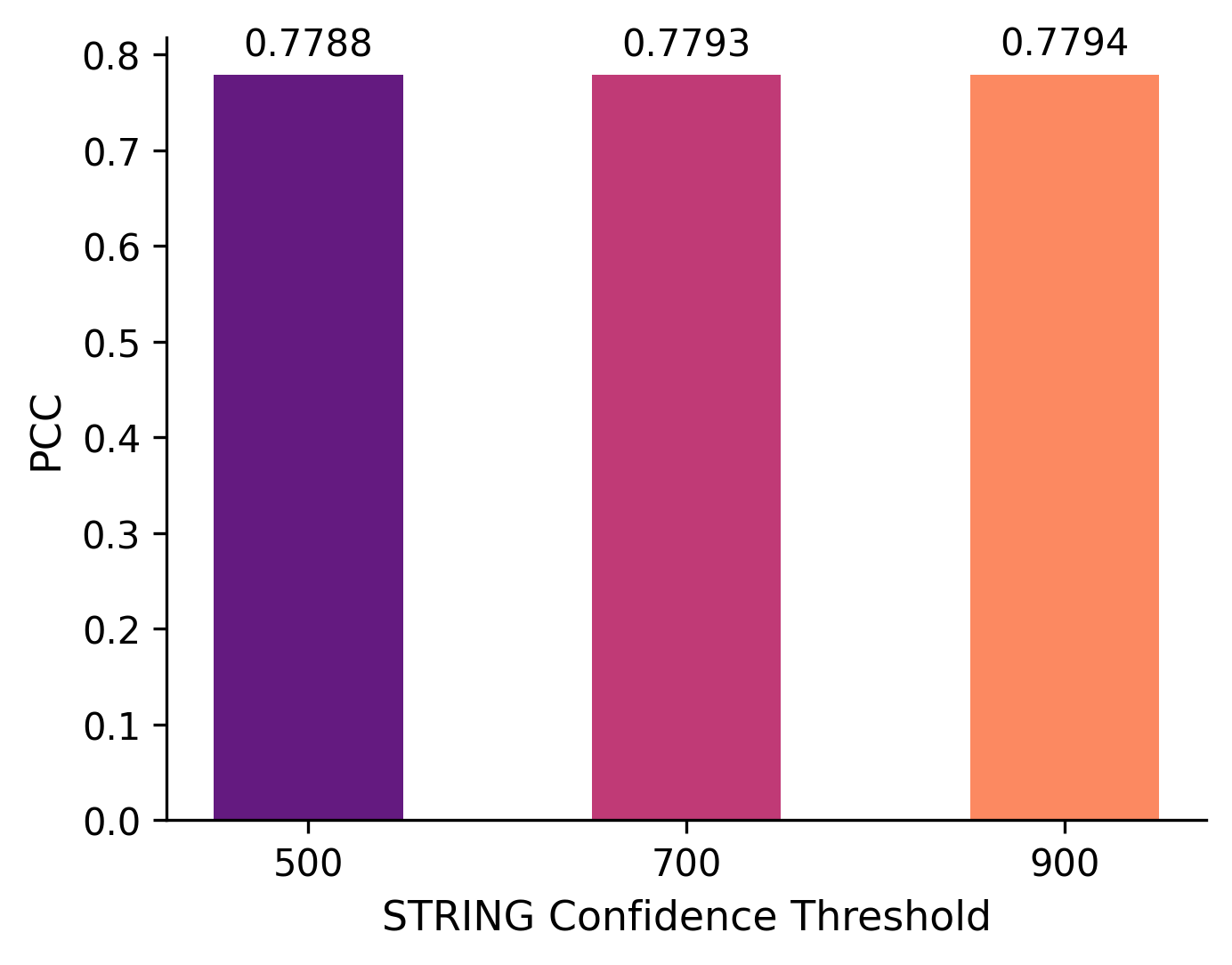}
\captionof{figure}{STRING-DB confidence threshold ablation \cite{zhang2021graph}. Stricter interaction filtering yields a small, monotonic improvement.}
\label{fig:ablation_string}
\end{center}

\subsubsection{Joint $L \times K$ Grid Search}
\label{sec:ablation-jointgrid}

The depth and neighborhood-size ablations in Section~\ref{sec:ablation-arch} were conducted greedily, optimizing $L$ with $K$ held fixed, then optimizing $K$ with $L$ fixed at its newly selected value. This sequential procedure risks converging to a locally optimal combination that is not jointly optimal, since the effect of neighborhood size may depend on network depth and vice versa. To verify that the greedily selected configuration $(L{=}2, K{=}4)$ is not an artifact of ablation order, we conduct a joint grid search over $L \in \{1, 2, 4\}$ and $K \in \{4, 8\}$, centered on the greedy optimum.

The joint grid confirms $(L{=}2, K{=}4)$ as the best-performing combination (PCC = 0.787), matching the result obtained via the sequential greedy procedure exactly. The next-closest combination, $(L{=}1, K{=}4)$, trails only marginally (PCC = 0.786), while increasing depth to $L{=}4$ degrades performance regardless of neighborhood size (PCC = 0.773--0.775), consistent with the oversmoothing pattern observed in the depth ablation \cite{kipf2016semi}. We therefore retain the greedily selected configuration for all subsequent experiments, with the joint grid search providing additional confidence that this choice reflects a genuine joint optimum rather than an artifact of the order in which individual hyperparameters were tuned.

As a secondary check accompanying this analysis, we verify that Leiden domain assignment is stable across cross-validation folds by comparing the number of detected domains per fold on the primary breast section: domain counts range narrowly from 10 to 12 across five folds (coefficient of variation 0.058), well below a pre-registered stability threshold of 0.15. This confirms that fold-to-fold differences in downstream performance are not attributable to unstable or inconsistent domain clustering.

\begin{table}[width=\columnwidth,cols=3,pos=t]
\caption{Joint $L\times K$ grid search results (ablation subset, PCC).}\label{tbl:jointgrid}
\small
\begin{tabular*}{\columnwidth}{@{\extracolsep{\fill}}lcc@{}}
\toprule
$L$ & $K=4$ & $K=8$ \\
\midrule
1 & 0.786 & 0.786 \\
\textbf{2} & \textbf{0.787} & 0.777 \\
4 & 0.773 & 0.775 \\
\bottomrule
\end{tabular*}
\end{table}
\begin{center}
\includegraphics[width=\linewidth]{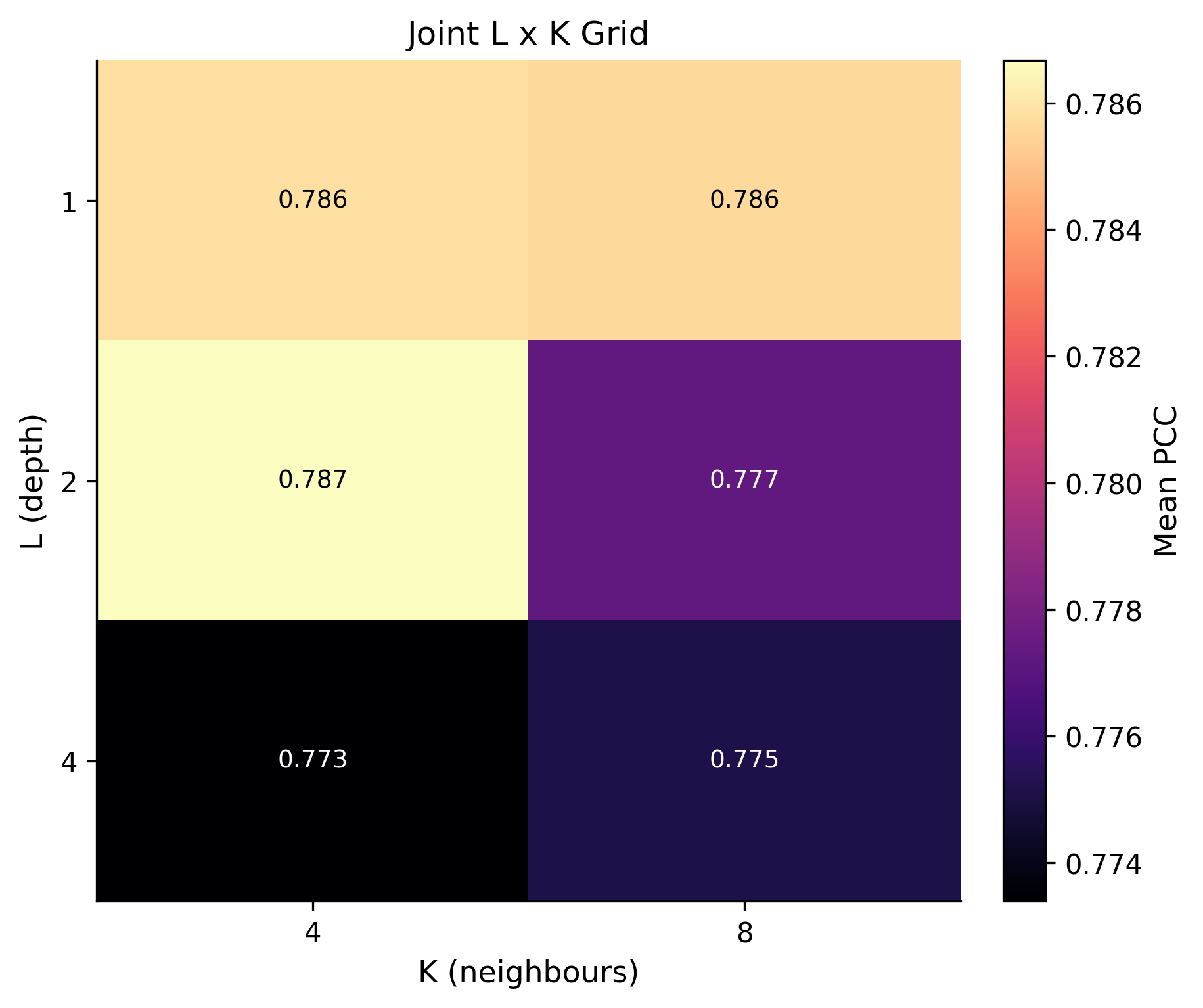}
\captionof{figure}{Joint $L \times K$ grid search results. The greedily selected configuration $(L{=}2, K{=}4)$ is confirmed as the jointly optimal combination.}
\label{fig:jointgrid}
\end{center}
\begin{center}
\includegraphics[width=0.7\linewidth]{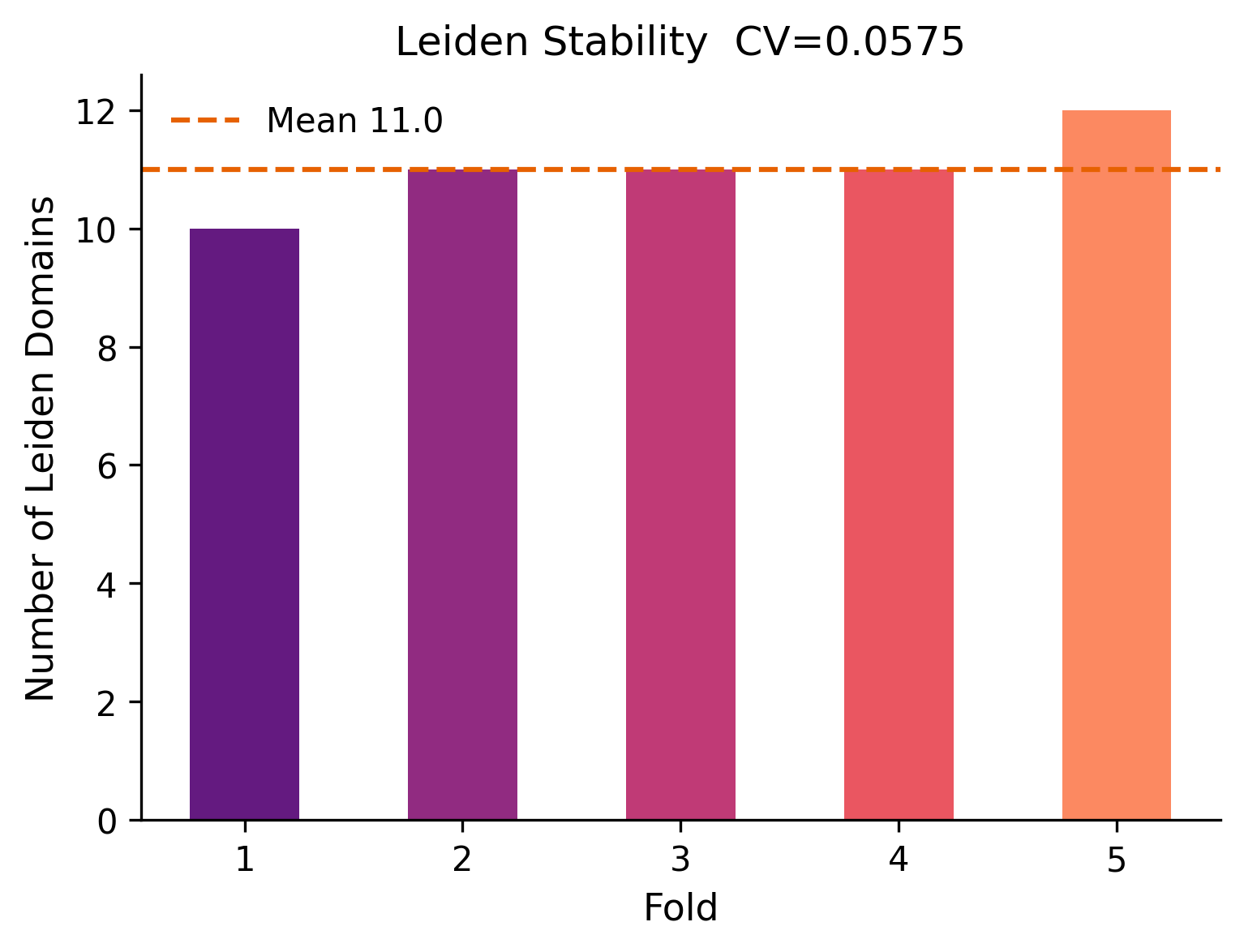}
\captionof{figure}{Stability of Leiden domain counts across cross-validation folds (coefficient of variation = 0.0575), confirming that fold-to-fold performance differences are not attributable to unstable domain clustering.}
\label{fig:leidenstability}
\end{center}
\subsubsection{Edge Typing Source: Expression-Derived vs Morphology-Derived vs No Typing}
\label{sec:ablation-edgesource}

DAEWConv's edge types are derived from Leiden clustering applied to expression data, an information source not available at inference time on new histology, since prediction targets are precisely what the model does not observe for unseen tissue. This is standard practice given that domain assignment for held-out spots is computed via nearest-centroid matching (Section~\ref{sec:leiden}), but it raises a natural question: does the benefit of edge typing come specifically from expression-derived domains, or would a similar gain arise from any structured edge typing, including one derived purely from histology morphology, which is available at inference time for every spot?

To test this, we compare three variants of edge construction on held-out spots, using a fixed spatial $k$-nearest-neighbor graph in all cases: our expression-derived typing (Leiden domains computed from training expression, as used throughout this work), a morphology-derived alternative in which edge types are assigned by thresholding cosine similarity between UNI \cite{chen2024towards} embeddings of neighboring spots, and a no-typing control using an untyped $k$-NN graph equivalent to a standard GCN \cite{kipf2016semi}. Expression-derived typing achieves the highest mean PCC (0.683), substantially outperforming both alternatives. Morphology-derived typing performs worst of the three (0.598), falling 0.053 PCC below even the untyped control (0.651), while the untyped control itself trails expression-derived typing by 0.032 PCC.

This result clarifies an important limitation of DAEWConv as currently formulated: its benefit derives specifically from the biological structure captured by expression-based domain assignment \cite{hu2021spagcn}, not merely from the presence of typed edges in general. Naively substituting a histology-derived proxy for domain structure is actively harmful relative to no typing at all, likely because morphological similarity does not reliably track the underlying transcriptional domain boundaries that motivate edge typing in the first place; two spots may appear visually similar in H\&E while belonging to transcriptionally distinct regions, or vice versa. This underscores that domain-aware edge typing is not a general-purpose architectural trick applicable to any notion of spot similarity, but specifically leverages the correspondence between expression-derived domains and the biological process being modeled.

\begin{table}[t]
\caption{Edge typing source ablation (held-out evaluation, PCC).}\label{tbl:edgesource}
\small
\begin{tabular}{@{}lc@{}}
\toprule
Edge typing source & Mean PCC \\
\midrule
Expression-derived (ours)                              & \textbf{0.683} \\
No typing (std.\ $k$-NN) \cite{kipf2016semi}           & 0.651 \\
Morphology-derived \cite{chen2024towards}              & 0.598 \\
\bottomrule
\end{tabular}
\end{table}
\begin{center}
\includegraphics[width=0.95\linewidth]{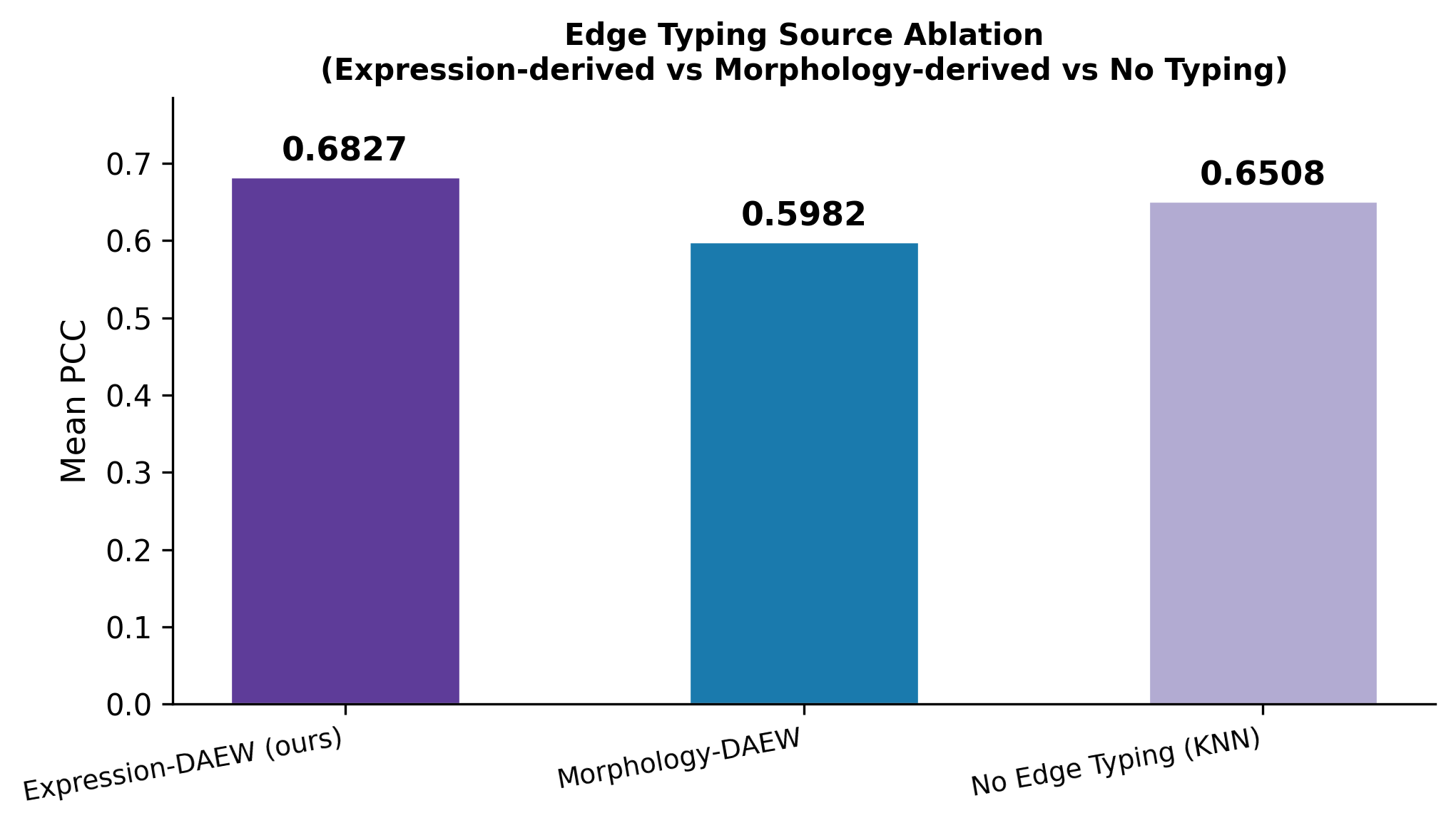}
\captionof{figure}{Edge typing source ablation. Expression-derived Leiden domains \cite{hu2021spagcn} outperform both an untyped graph \cite{kipf2016semi} and a morphology-derived alternative \cite{chen2024towards}, which performs worse than no typing at all.}
\label{fig:edgesource}
\end{center}

\subsection{Statistical Validation}

\subsubsection{Per-Gene Win Rate and McNemar Test}

Aggregate correlation metrics such as mean PCC can mask substantial heterogeneity in per-gene performance, since a small number of genes with very large errors or very strong fits can dominate an averaged score. To assess whether HierarchicalDAEW's advantage holds consistently across individual genes rather than being driven by a subset of favourable genes, we compute a per-gene win rate against each baseline, that is, for every gene, we compare per-fold PCC between HierarchicalDAEW and the baseline and classify the gene as a win, loss or tie based on which method achieves higher correlation in the majority of the folds. We additionally apply McNemar's test \cite{mcnemar1947note} to assess whether the observed win/loss asymmetry across genes is statistically significant, with multiple comparisons corrected using the Benjamini-Hochberg procedure \cite{benjamini1995controlling}.

HierarchicalDAEW wins on 99.2--100\% of the genes against ten of the thirteen baselines (STNet \cite{he2020integrating}, HisToGene \cite{pang2021leveraging}, Hist2ST \cite{zeng2022spatial}, BLEEP \cite{xie2023spatially}, EGGN \cite{yang2024spatial}, TRIPLEX \cite{chung2024accurate}, MERGE \cite{ganguly2025merge}, THItoGene \cite{jia2024thitogene}, GCN \cite{kipf2016semi}, GAT \cite{velivckovic2017graph}, GraphSAGE \cite{hamilton2017inductive}, MLP), with McNemar's test \cite{mcnemar1947note} rejecting the null hypothesis of symmetric win/loss distribution in each case $(p < 0.0001)$. Against SEPAL \cite{mejia2023sepal}, the closest competitor throughout the work, HierarchicalDAEW wins on 78.1\% of the genes and loses on only 21.9\%, an asymmetry that remains highly significant under McNemar's test $(p<0.0001)$ despite the narrower aggregate PCC margin observed in the single-section benchmark. This indicates that even when mean PCC differences between HierarchicalDAEW and SEPAL are small or statistically indistinguishable at the aggregate level, HierarchicalDAEW's advantage is nonetheless consistent and systematic across the large majority of individual genes, rather than being driven by a handful of outlier genes in either direction.

\begin{table*}[t]
\caption{Per-gene win rate against each baseline.}\label{tbl:winrate}
\small
\setlength{\tabcolsep}{12pt}
\begin{tabular*}{\textwidth}{@{\extracolsep{\fill}}lccr@{}}
\toprule
Baseline & Win Rate & Loss Rate & McNemar $p$ \\
\midrule
STNet \cite{he2020integrating}         & 100.0\% & 0.0\%  & $<0.0001$ \\
HisToGene \cite{pang2021leveraging}    & 100.0\% & 0.0\%  & $<0.0001$ \\
Hist2ST \cite{zeng2022spatial}         & 100.0\% & 0.0\%  & $<0.0001$ \\
BLEEP \cite{xie2023spatially}          & 100.0\% & 0.0\%  & $<0.0001$ \\
TRIPLEX \cite{chung2024accurate}       & 100.0\% & 0.0\%  & $<0.0001$ \\
GCN \cite{kipf2016semi}                & 100.0\% & 0.0\%  & $<0.0001$ \\
GAT \cite{velivckovic2017graph}        & 100.0\% & 0.0\%  & $<0.0001$ \\
MLP                                    & 100.0\% & 0.0\%  & $<0.0001$ \\
EGGN \cite{yang2024spatial}            & 99.6\%  & 0.4\%  & $<0.0001$ \\
MERGE \cite{ganguly2025merge}          & 99.4\%  & 0.6\%  & $<0.0001$ \\
THItoGene \cite{jia2024thitogene}      & 99.4\%  & 0.6\%  & $<0.0001$ \\
GraphSAGE \cite{hamilton2017inductive} & 99.2\%  & 0.8\%  & $<0.0001$ \\
SEPAL \cite{mejia2023sepal}            & 78.1\%  & 21.9\% & $<0.0001$ \\
\bottomrule
\end{tabular*}
\end{table*}
\begin{center}
\includegraphics[width=\linewidth]{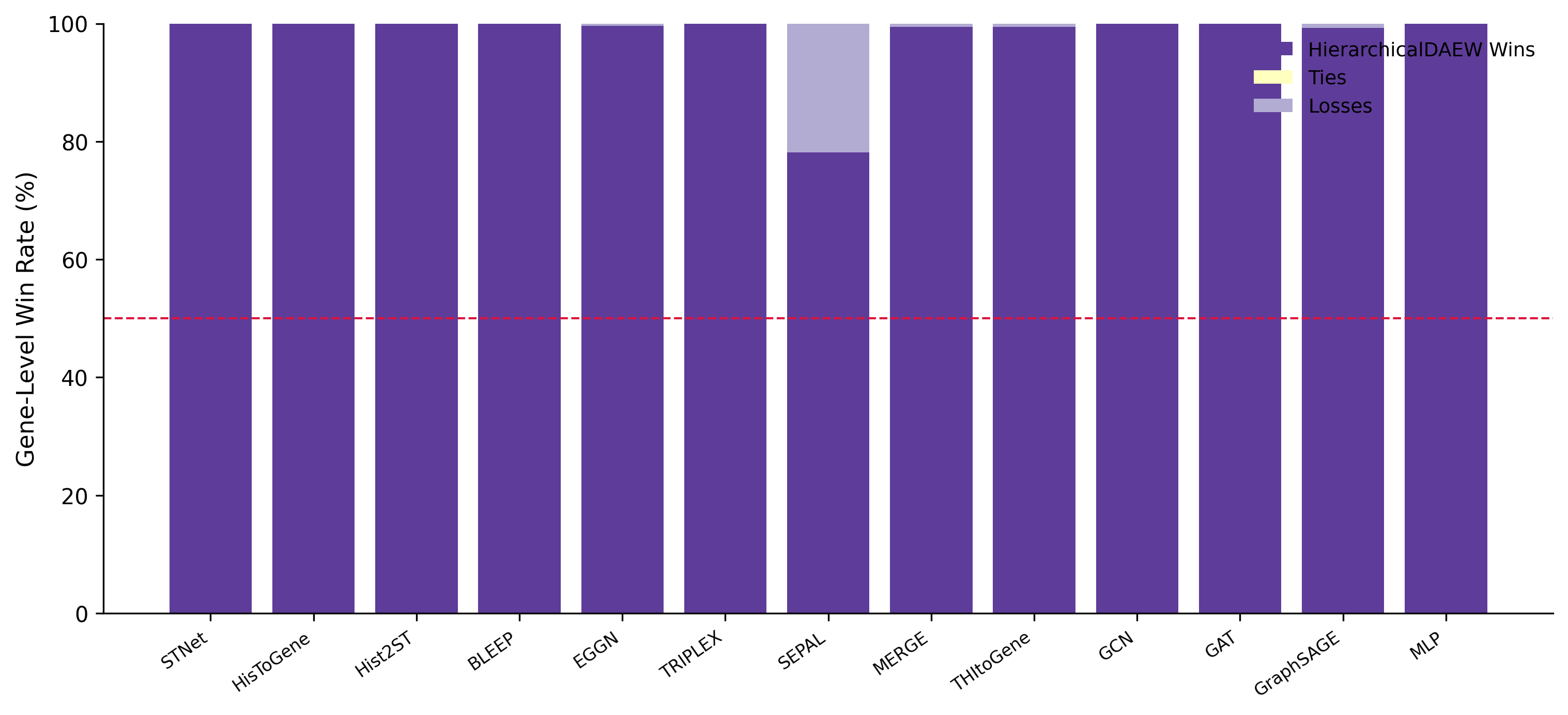}
\captionof{figure}{Per-gene win rate of HierarchicalDAEW against each baseline using McNemar's test \cite{mcnemar1947note}. HierarchicalDAEW wins on the large majority of genes against every baseline, including a 78.1\% win rate against SEPAL \cite{mejia2023sepal} despite the narrow aggregate PCC margin.}
\label{fig:winrate}
\end{center}
\subsubsection{Bootstrap Confidence Intervals and Effect Sizes}

To complement the per-fold significance tests reported in the main results, we compute 95\% bootstrap confidence intervals \cite{tibshirani1993introduction} on both the mean PCC difference and Cohen's $d$ effect size \cite{cohen2013statistical} between HierarchicalDAEW and each baseline, using 10,000 bootstrap resamples of the paired per-fold PCC differences on the single-section benchmark. We additionally compute the minimum detectable effect (MDE) at 80\% power and $\alpha = 0.05$ given our 5-fold design, which is 0.0110 PCC; any observed difference below this threshold cannot be reliably distinguished from noise at our sample size, regardless of its point estimate.

The bootstrap confidence interval on mean PCC difference excludes zero for every baseline except SEPAL \cite{mejia2023sepal}, whose 95\% CI ($-$0.0006 to 0.0099) both crosses zero and falls entirely below the MDE, confirming that this comparison is genuinely inconclusive at our fold count rather than merely non-significant due to an underpowered test. Every other baseline shows a mean PCC difference safely above the MDE with a confidence interval bounded well away from zero, ranging from 0.029 against THItoGene \cite{jia2024thitogene} to 0.097 against BLEEP \cite{xie2023spatially}.

Cohen's $d$ \cite{cohen2013statistical} values in Table~\ref{tbl:bootstrap} range from 0.8 (SEPAL) to a clipped ceiling of 20 for several baselines. With five paired folds, the denominator of $d$, the standard deviation of per-fold differences, is small, which produces large nominal effect sizes; we therefore treat the bootstrap confidence interval on $\Delta$PCC as the primary quantity for effect magnitude, with Cohen's $d$ reported alongside for completeness.

\begin{table*}[width=\textwidth,cols=4,pos=t]
\caption{Bootstrap \cite{tibshirani1993introduction} 95\% confidence intervals for PCC difference and Cohen's $d$ \cite{cohen2013statistical} (Breast S1).}\label{tbl:bootstrap}
\small
\begin{tabular*}{\textwidth}{@{\extracolsep{\fill}}lccc@{}}
\toprule
Baseline & $\Delta$PCC & 95\% CI & Cohen's $d$ \\
\midrule
BLEEP \cite{xie2023spatially}          & 0.0966 & [0.0925, 0.1019] & 17.97 \\
MLP                                     & 0.0787 & [0.0737, 0.0829] & 14.31 \\
HisToGene \cite{pang2021leveraging}    & 0.0790 & [0.0581, 0.1028] & 3.07  \\
STNet \cite{he2020integrating}         & 0.0776 & [0.0715, 0.0822] & 12.04 \\
TRIPLEX \cite{chung2024accurate}       & 0.0687 & [0.0650, 0.0748] & 11.20 \\
GAT \cite{velivckovic2017graph}        & 0.0602 & [0.0549, 0.0649] & 10.92 \\
GCN \cite{kipf2016semi}                & 0.0556 & [0.0516, 0.0615] & 9.48  \\
EGGN \cite{yang2024spatial}            & 0.0548 & [0.0483, 0.0625] & 6.76  \\
Hist2ST \cite{zeng2022spatial}         & 0.0542 & [0.0473, 0.0625] & 6.33  \\
GraphSAGE \cite{hamilton2017inductive} & 0.0384 & [0.0339, 0.0442] & 6.70  \\
MERGE \cite{ganguly2025merge}          & 0.0357 & [0.0288, 0.0426] & 4.54  \\
THItoGene \cite{jia2024thitogene}      & 0.0291 & [0.0223, 0.0378] & 3.18  \\
SEPAL \cite{mejia2023sepal}            & 0.0046 & [-0.0006, 0.0099] & 0.77 \\
\bottomrule
\end{tabular*}
\end{table*}
\begin{center}
\includegraphics[width=0.95\linewidth]{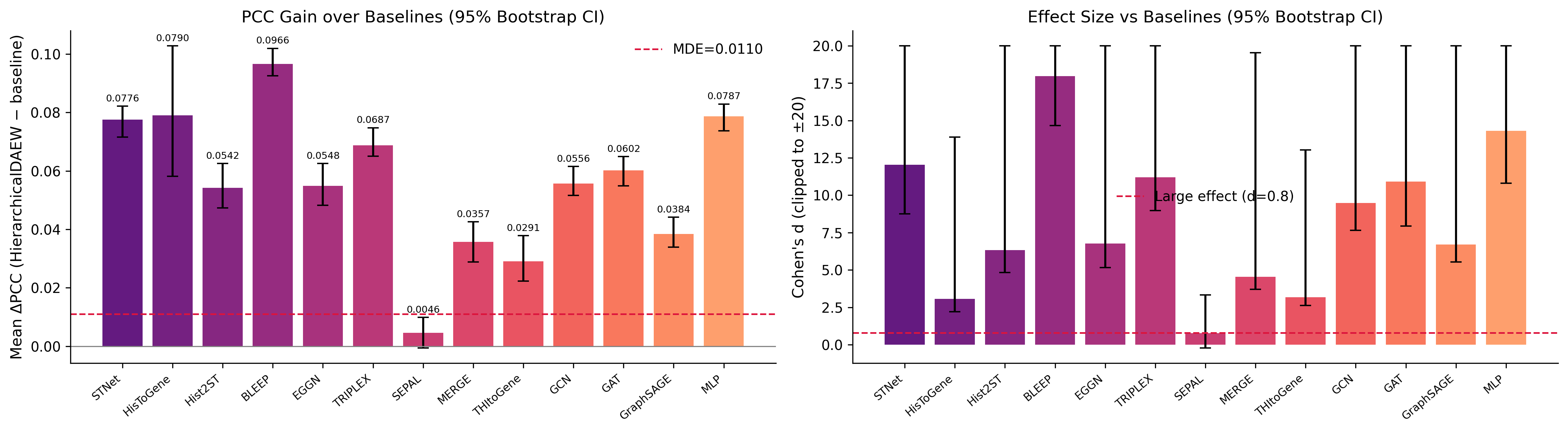}
\captionof{figure}{Bootstrap \cite{tibshirani1993introduction} 95\% confidence intervals for mean PCC difference against each baseline. Only the SEPAL \cite{mejia2023sepal} comparison crosses zero, consistent with the non-significant result reported in Table~\ref{tbl:singlesection}.}
\label{fig:bootstrap}
\end{center}
\subsubsection{Multi-Seed Reproducibility}

The cross-validation results reported throughout this work use a fixed random seed, which controls fold assignment and model initialization but does not, by itself, establish that the reported performance gap is robust to initialization alone. To assess this separately, we repeat training and evaluation of HierarchicalDAEW and its closest architectural baseline, GCN \cite{kipf2016semi}, across three random seeds (0, 42, 123), using an identical 3-fold split protocol and gene panel for each seed, and treat the resulting seed-level variation as distinct from the fold-level variation already captured elsewhere.

HierarchicalDAEW achieves a mean PCC of 0.7617 across seeds (standard deviation 0.0031, range 0.7578--0.7655), while GCN \cite{kipf2016semi} achieves a mean PCC of 0.7254 (standard deviation 0.0075, range 0.7165--0.7349). A paired $t$-test across the three seeds confirms that HierarchicalDAEW's advantage over GCN is statistically significant despite the small number of seeds tested ($t = 7.70, p = 0.0165$) \cite{cohen2013statistical}. Beyond confirming the performance gap itself, this analysis shows that HierarchicalDAEW is also more than twice as stable across random initialization as GCN, with less than half the standard deviation, suggesting that the architectural components introduced in this work, particularly the residual connections and batch normalization within DAEWConv, contribute to more consistent convergence in addition to higher accuracy.

\begin{table}[t]
\caption{Multi-seed reproducibility (3 seeds, PCC).}
\label{tbl:multiseed}
\small
\centering
\resizebox{\columnwidth}{!}{%
\begin{tabular}{lccl}
\toprule
Model & Mean PCC & Std & Range \\
\midrule
HierarchicalDAEW        & \textbf{0.7617} & \textbf{0.0031} & 0.7578--0.7655 \\
GCN \cite{kipf2016semi} & 0.7254          & 0.0075          & 0.7165--0.7349 \\
\bottomrule
\end{tabular}%
}
\end{table}

\begin{center}
\includegraphics[width=0.8\linewidth]{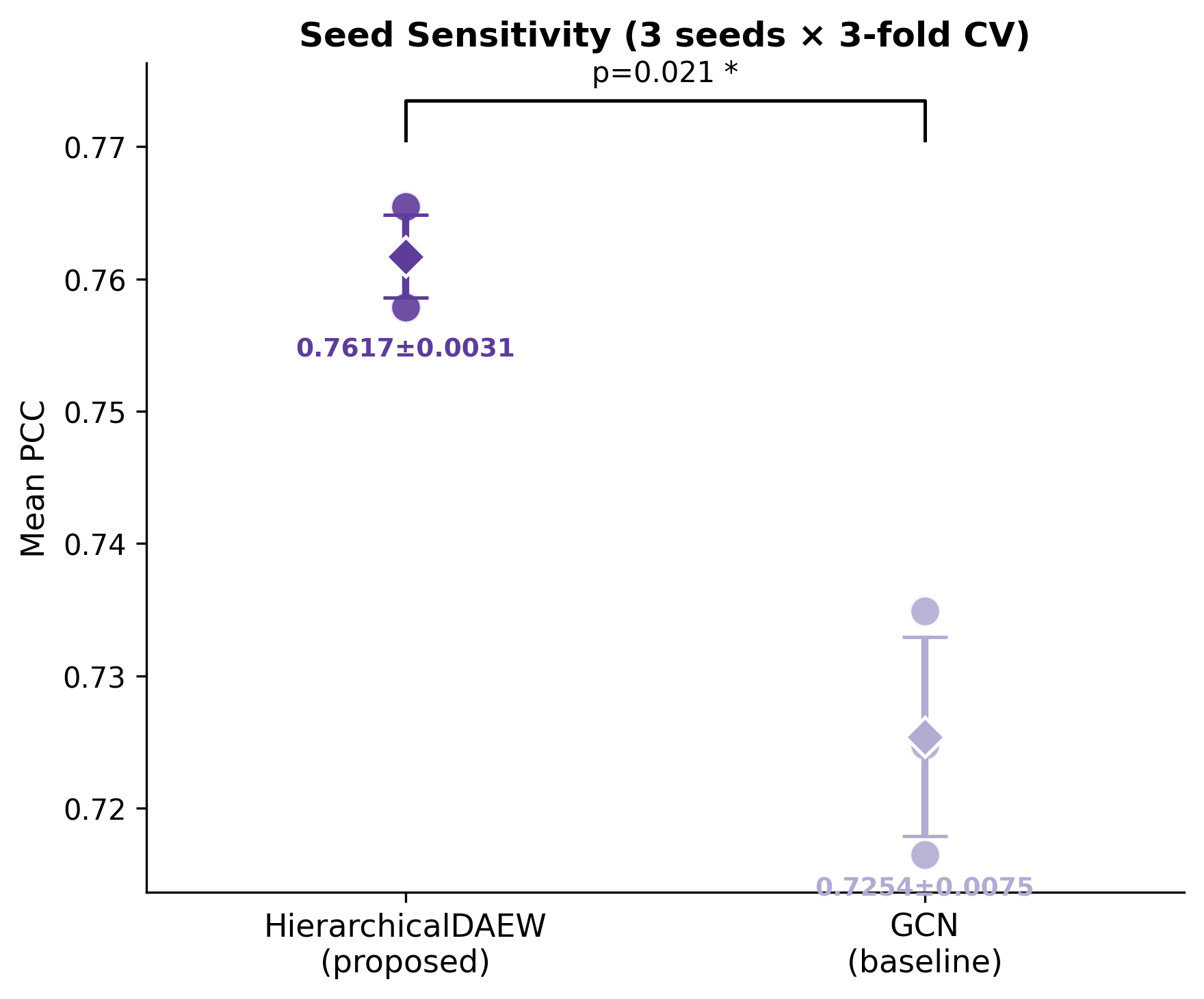}
\captionof{figure}{Multi-seed reproducibility across three random seeds. HierarchicalDAEW shows both higher mean PCC and lower seed-to-seed variance than GCN \cite{kipf2016semi}.}
\label{fig:multiseed}
\end{center}

\subsubsection{Negative Control: Shuffled Expression Labels}

A model that achieves high correlation not by learning a genuine morphology-to-expression mapping, but by exploiting an artifact correlated with spatial position, such as library size, spot density, or edge effects near tissue boundaries, would still report a high PCC despite capturing no real biological signal. This is a standard failure mode to rule out in spatial prediction tasks \cite{ruiz2025completing}, where the graph structure and spatial coordinates themselves carry statistical regularities independent of the underlying biology. To test for this, we conduct a negative control in which expression targets are randomly shuffled across spots while every other component of the pipeline, the spot graph, edge types derived from Leiden domains, and histology features, is left entirely unchanged. The model is then trained and evaluated identically to the real-label setting, using the same architecture, hyperparameters, and 3-fold split.

Under real labels, HierarchicalDAEW achieves a mean PCC of 0.7578 on this held-out split. Under shuffled labels, with all structural and feature information held fixed, PCC drops to 0.0121, a reduction of 0.7458. This residual value is consistent with the correlation expected by chance between two statistically independent variables at this sample size, rather than indicating any remaining predictive signal. Because the graph topology, edge typing, and spot features are identical in both conditions, and only the correspondence between spots and their expression labels is broken, this result isolates the source of HierarchicalDAEW's predictive accuracy to the histology-to-expression relationship specifically, rather than to any structural or positional property of the spot graph itself.

\begin{table}[width=\columnwidth,cols=2,pos=t]
\caption{Negative control: real versus shuffled expression labels (PCC).}\label{tbl:negcontrol}
\small
\begin{tabular*}{\columnwidth}{@{\extracolsep{\fill}}lc@{}}
\toprule
Condition & PCC \\
\midrule
Real labels      & \textbf{0.7578} \\
Shuffled labels  & 0.0121 \\
Difference       & 0.7458 \\
\bottomrule
\end{tabular*}
\end{table}
\begin{center}
\includegraphics[width=0.7\linewidth]{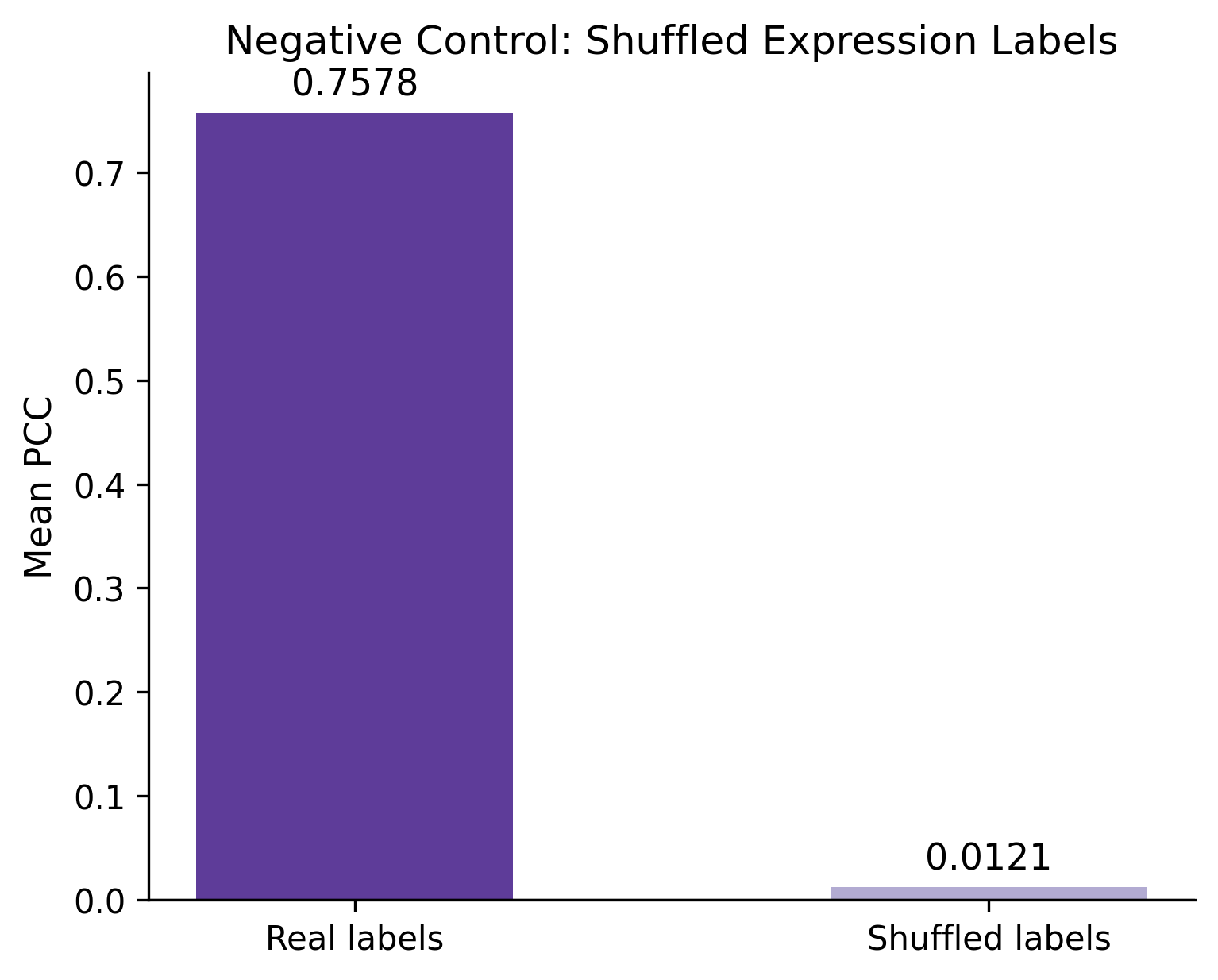}
\captionof{figure}{Negative control with shuffled expression labels. PCC collapses from 0.758 to 0.012 when spot-expression correspondence is broken, ruling out positional shortcuts \cite{ruiz2025completing}.}
\label{fig:negcontrol}
\end{center}
\subsection{Mechanistic Analysis}
\label{sec:mechanistic}

\subsubsection{Projection Matrix Divergence (Frobenius Distance)}
\label{sec:frobenius}

DAEWConv's central architectural claim is that separate per-edge-type projection matrices $W_t^{(\ell)}$ allow the model to learn distinct message-passing behavior for intra-domain, inter-domain, and boundary connections, rather than treating all spatial edges uniformly following \citet{schlichtkrull2018modeling}. To verify that the model actually exploits this capacity rather than converging to near-identical projections across types, which would make the edge typing mechanism functionally redundant, we measure the Frobenius distance $\lVert W_s^{(\ell)} - W_t^{(\ell)} \rVert_F$ between every pair of type-specific projection matrices at each DAEWConv layer of a trained model.

Across both layers, the three pairwise distances between $W_{\text{intra}}$, $W_{\text{inter}}$, and $W_{\text{boundary}}$ are all large and of comparable magnitude, ranging from 26.46 to 26.80, with no pair of projection matrices converging toward one another during training. This indicates that all three edge types are treated as meaningfully distinct by the model, rather than the architecture collapsing two or more types into effectively redundant projections. Notably, this divergence occurs primarily through the projection matrices themselves rather than through the learnable scalar gates $\alpha_t^{(\ell)}$, which remain close to 1.0 for all three edge types at both layers (ranging from 0.9975 to 1.0071). This provides a mechanistic explanation for the earlier ablation finding that fixed versus learned gates produced only a marginal difference in accuracy: the gates converge to near-uniform values regardless of whether they are learned or fixed, while the substantive differentiation between edge types is captured entirely by the per-type weight matrices $W_t^{(\ell)}$, confirming that separate projections, not adaptive gating, are the primary mechanism through which DAEWConv exploits domain structure.

\begin{center}
\includegraphics[width=\linewidth]{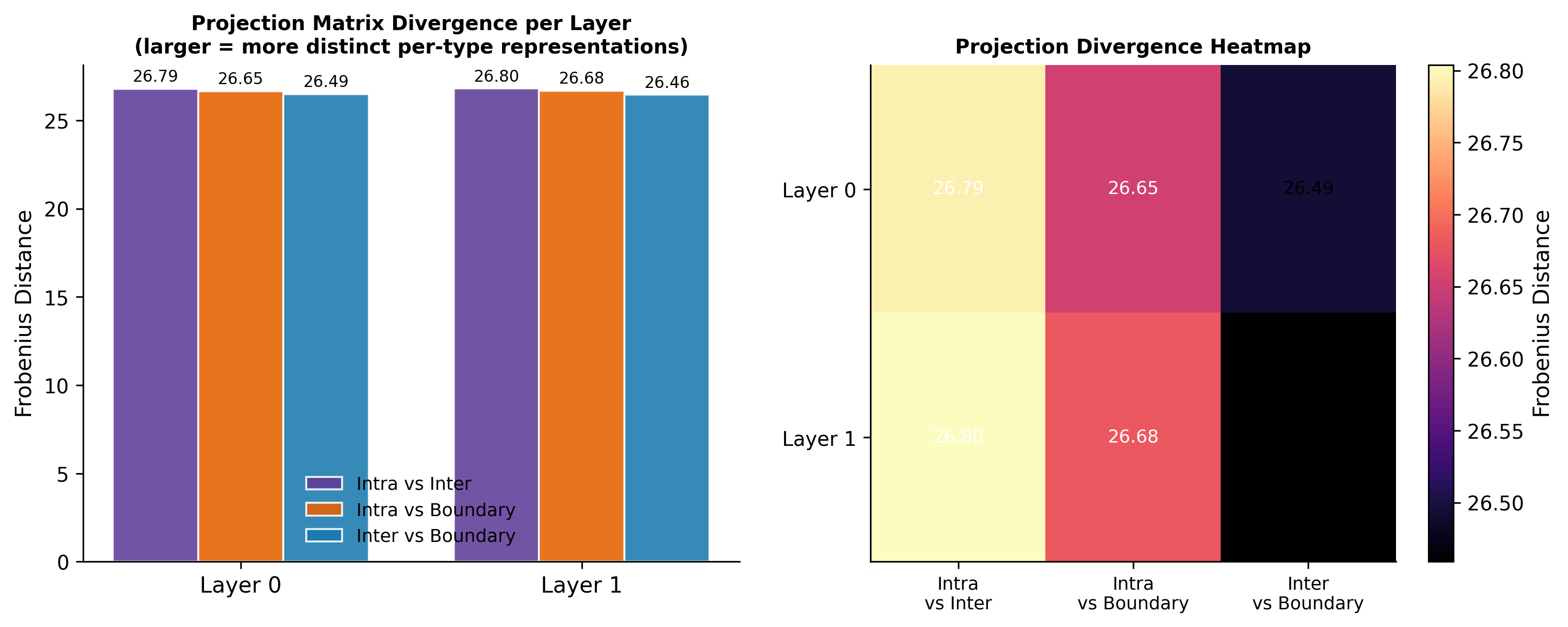}
\captionof{figure}{Frobenius distance between DAEWConv's type-specific projection matrices \cite{schlichtkrull2018modeling}, by layer. All three edge types remain clearly distinct throughout training, indicating none collapse into redundant projections.}
\label{fig:frobenius}
\end{center}
\subsubsection{CrossScaleGate Weight Distribution by Edge Type}
\label{sec:gateweights}

The CrossScaleGate mechanism (Section~\ref{hierarchical}) learns a per-spot gate $g_i$ that controls how much domain-level context $\tilde{h}_{c_i}$ each spot incorporates relative to its own local representation $h_i$. This is architecturally distinct from the per-edge-type gates $\alpha_t^{(\ell)}$ analyzed above: whereas $\alpha_t^{(\ell)}$ modulates message strength during local spot-to-spot aggregation within DAEWConv, $g_i$ operates after domain-level pooling and controls fusion between a spot's own representation and the aggregate state of its tissue domain. If CrossScaleGate has learned a meaningful signal, gate values should differ systematically depending on whether a spot sits well within a homogeneous domain or near a domain boundary \cite{dong2022deciphering}, since these two settings differ in how much a spot's local representation alone can be trusted.

We compute the mean gate value $g_i$ separately for spots classified as intra-domain, inter-domain-adjacent, and boundary, using the same edge-type categorization applied throughout this work. Mean gate weight is highest for inter-domain-adjacent spots (0.4554), followed closely by boundary spots (0.4445), and lowest for intra-domain spots (0.3976). This ordering is consistent with the intended function of the gate: spots deep within a homogeneous domain, whose local neighborhood already provides a reliable, low-ambiguity signal, rely comparatively less on domain-level context, while spots near domain boundaries or adjacent to other domains \cite{dong2022deciphering}, where local signal alone is less reliable due to nearby structural discontinuity, draw more heavily on the pooled domain-level representation. The gap between intra-domain and the two boundary-adjacent categories (approximately 0.05--0.06) indicates that CrossScaleGate has learned a spatially meaningful, rather than uniform, fusion policy, providing further evidence that the hierarchical component of the architecture is contributing a functionally distinct signal to the final prediction.

\begin{table}[width=\columnwidth,cols=2,pos=t]
\caption{Mean CrossScaleGate weight by spot edge-type category.}\label{tbl:gateweights}
\small
\begin{tabular*}{\columnwidth}{@{\extracolsep{\fill}}lc@{}}
\toprule
Spot category & Mean gate weight $g_i$ \\
\midrule
Intra-domain            & 0.3976 \\
Inter-domain-adjacent   & \textbf{0.4554} \\
Boundary                & 0.4445 \\
\bottomrule
\end{tabular*}
\end{table}

\begin{center}
\includegraphics[width=\linewidth]{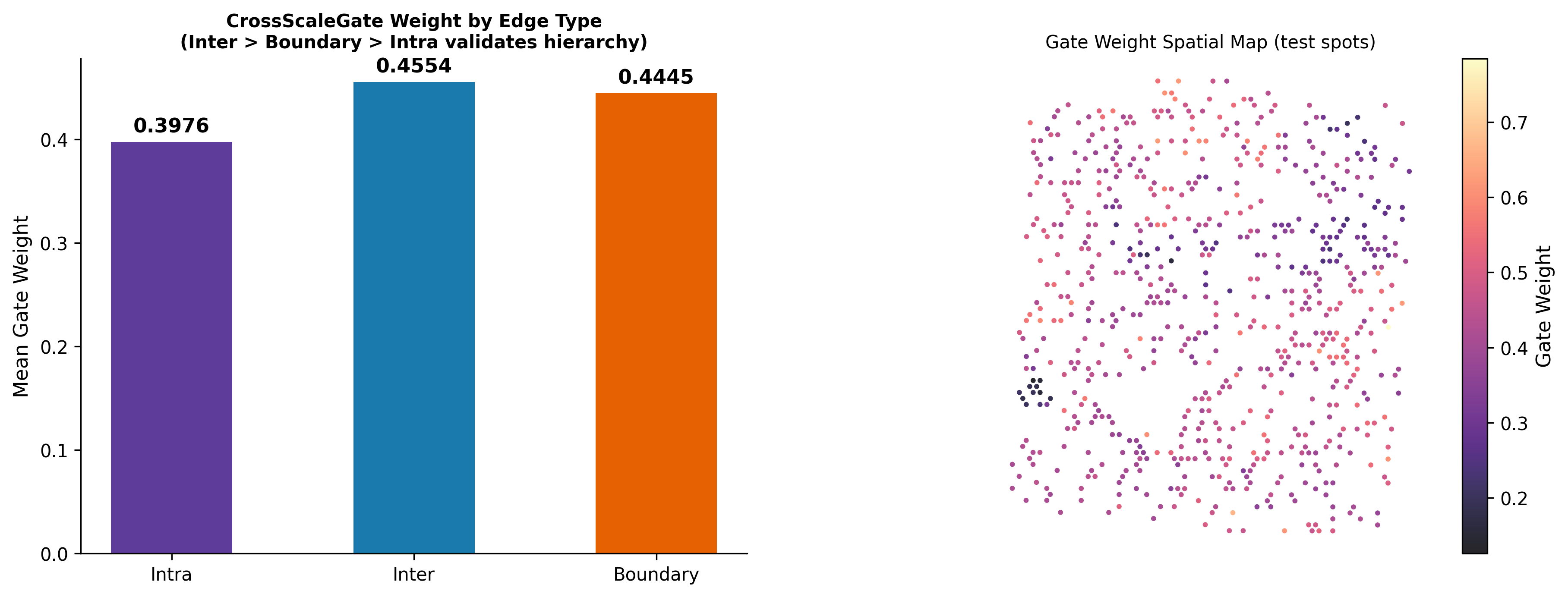}
\captionof{figure}{CrossScaleGate weight by spot edge-type category (left) and its spatial distribution across a test section (right). Inter-domain and boundary spots draw more heavily on domain-level context than intra-domain spots \cite{dong2022deciphering}.}
\label{fig:gateweights}
\end{center}

\subsubsection{Per-Domain Performance Breakdown}
\label{sec:perdomain}

Having established that DAEWConv learns distinct projections per edge type and that CrossScaleGate allocates domain context selectively, we next ask whether this architectural differentiation translates into a measurable difference in prediction quality across spot categories. We compute per-spot PCC, correlation between predicted and observed expression across the gene panel at each individual spot \cite{benesty2009pearson}, and average this separately for spots in each of the three edge-type categories used throughout this work.

Intra-domain spots achieve the highest mean per-spot PCC (0.9242, $n=304$), followed by boundary spots (0.9145, $n=142$) and inter-domain-adjacent spots (0.9122, $n=238$). The gap between categories is modest, approximately 0.01--0.02 PCC, but consistently favors intra-domain spots, consistent with the intuition that spots embedded within a homogeneous tissue domain benefit from more coherent local neighborhood signal than spots situated near domain transitions. Notably, boundary spots slightly outperform inter-domain-adjacent spots despite ostensibly facing the most structurally ambiguous local context, which is consistent with the CrossScaleGate finding above: boundary spots receive the second-highest gate weight (0.4445), allowing them to compensate for locally ambiguous signal by drawing more heavily on domain-level context, whereas inter-domain-adjacent spots, despite an even higher gate weight (0.4554), may draw context from a less coherent or less representative neighboring domain. Overall, the relatively narrow spread across all three categories suggests that the architecture's domain-aware design substantially closes, though does not fully eliminate, the performance gap between structurally straightforward and structurally ambiguous regions of tissue \cite{yuan2024benchmarking}.

\begin{table}[width=\columnwidth,cols=3,pos=t]
\caption{Per-spot PCC \cite{benesty2009pearson} by domain edge-type category (single fold, Breast S1).}\label{tbl:perdomain}
\small
\begin{tabular*}{\columnwidth}{@{\extracolsep{\fill}}lcc@{}}
\toprule
Spot category & Mean per-spot PCC & $n$ \\
\midrule
Intra-domain            & \textbf{0.9242} & 304 \\
Boundary                & 0.9145           & 142 \\
Inter-domain-adjacent   & 0.9122           & 238 \\
\bottomrule
\end{tabular*}
\end{table}
\begin{center}
\includegraphics[width=\linewidth]{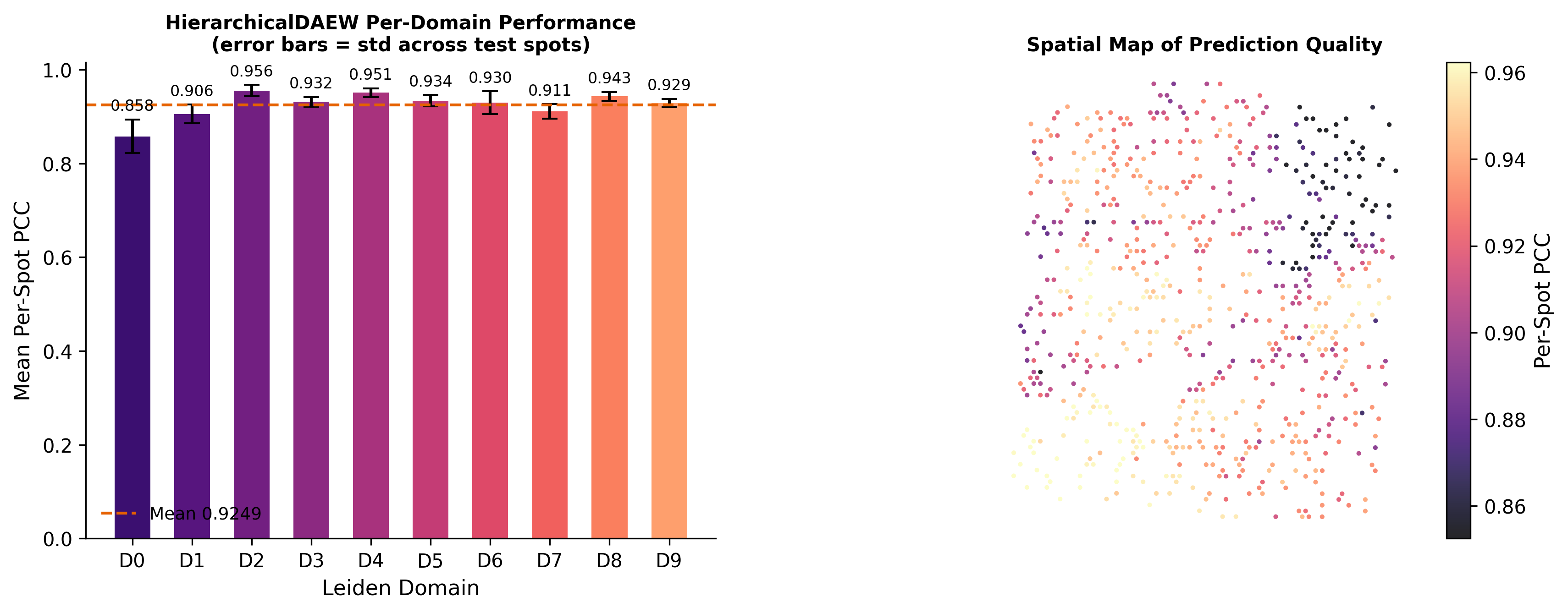}
\captionof{figure}{Per-spot PCC \cite{benesty2009pearson} by domain edge-type category. Intra-domain spots achieve the highest accuracy, though the spread across categories is modest, indicating the domain-aware design substantially closes the performance gap \cite{yuan2024benchmarking}.}
\label{fig:perdomain}
\end{center}

\subsubsection{Node Embedding Structure (UMAP)}
\label{sec:umap}

To examine whether the learned spot representations $h_i'$ organize according to tissue domain structure rather than collapsing into an undifferentiated embedding space, we project final-layer spot embeddings from held-out test spots into two dimensions using UMAP \cite{mcinnes2018umap}, and visualize the resulting layout colored in two complementary ways: by Leiden domain assignment, and by per-spot prediction quality.

Embeddings colored by domain form visually distinguishable clusters corresponding to the ten Leiden domains present in this section, with the majority of domains occupying largely separate regions of the projection rather than being interleaved. This is a meaningful check on the architecture: DAEWConv's per-type projections and the domain contrastive loss \cite{xie2023spatially} are both explicitly designed to encourage domain-consistent representations, but neither component directly guarantees this outcome, since the model is trained only to predict expression accurately, not to produce visually separable clusters. The fact that domain structure nonetheless emerges in the embedding space indicates that accurate expression prediction and domain-consistent representation are not competing objectives here; the model appears to solve the prediction task in part by first organizing spots according to their underlying tissue domain, then predicting expression conditional on that organization.

The second view, coloring the same embedding layout by per-spot PCC \cite{benesty2009pearson}, adds a diagnostic dimension the domain-colored view alone cannot provide. Rather than being uniformly distributed across the embedding space, spots with lower prediction quality concentrate specifically near the boundaries between domain clusters, while spots deep within a single cluster tend to show higher per-spot PCC. This spatial correspondence in embedding space mirrors the numerical pattern already established in the per-domain performance breakdown, where boundary and inter-domain-adjacent spots showed modestly lower per-spot PCC than intra-domain spots. Taken together, the two views suggest that prediction difficulty is concentrated precisely where domain identity itself becomes ambiguous, at the transition zones between clusters, which is the same region where a spot's local neighborhood provides comparatively less reliable signal and where CrossScaleGate was shown to rely more heavily on pooled domain-level context. This convergence across three independent analyses, edge-type performance breakdown, gate weight allocation, and embedding-space geometry, indicates that domain boundaries are a consistent locus of difficulty for the model, rather than an artifact specific to any single analysis method.

\begin{center}
\makebox[\linewidth][c]{%
  \includegraphics[width=1.15\linewidth]{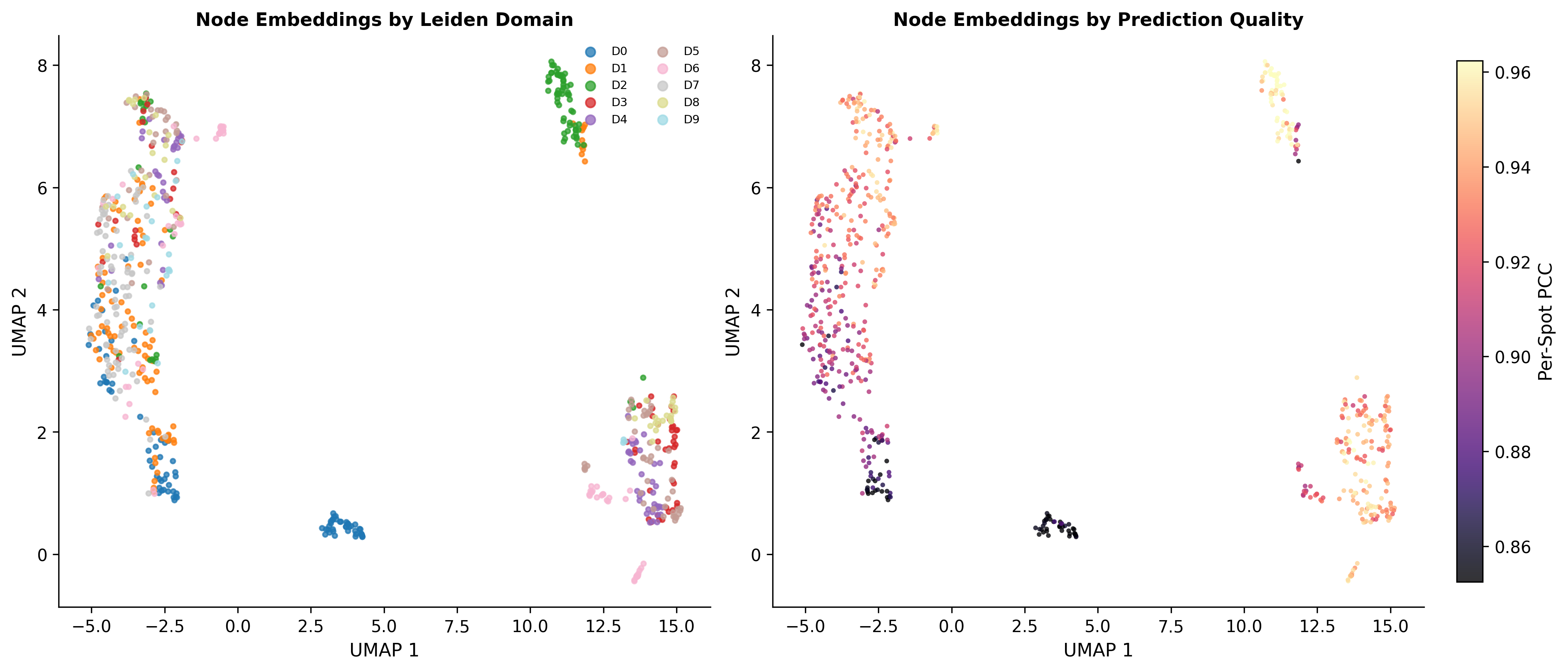}%
}
\captionof{figure}{UMAP \cite{mcinnes2018umap} projection of learned spot embeddings on held-out test spots, colored by Leiden domain (left) and by per-spot PCC \cite{benesty2009pearson} (right).}
\label{fig:umap}
\end{center}

\subsection{Uncertainty Quantification}

\subsubsection{NIG Calibration: Coverage and ENCE}
\label{sec:calibration}

Beyond point-prediction accuracy, we care about whether the model's uncertainty estimates are actually useful, whether high predicted uncertainty really does correspond to a higher chance of being wrong, rather than just noise attached to the output. We check this two ways: empirical coverage at the 90\% confidence level, meaning the fraction of true expression values that actually fall inside the model's predicted 90\% interval (which should sit at 0.90 if calibration is working), and ENCE \cite{kuleshov2018accurate}, which checks whether predicted variance tracks actual squared error across bins of increasing uncertainty.

Coverage comes out to 0.903, essentially exact. Ninety percent of true values fall inside their predicted 90\% interval, with no meaningful over- or under-confidence at that threshold. ENCE is 0.4984, which tells a slightly more mixed story: there is a real but incomplete relationship between predicted variance and actual squared error. A value of 0 would mean perfect variance calibration; landing in this range is fairly typical for evidential regression models \cite{amini2020deep} and reflects how much harder it is to calibrate variance itself compared to just getting coverage right. We also directly checked whether predicted uncertainty correlates with actual error at the spot level, and found a modest positive relationship (Spearman \cite{zar2005spearman} 0.176, Pearson \cite{benesty2009pearson} 0.119). It is not a strong signal, but it is a real one: the model's uncertainty estimates carry some genuine information about where it is more likely to be wrong, beyond what the aggregate coverage number alone tells us.

\begin{table}[width=\columnwidth,cols=2,pos=t]
\caption{NIG \cite{amini2020deep} uncertainty calibration summary.}\label{tbl:calibration}
\small
\begin{tabular*}{\columnwidth}{@{\extracolsep{\fill}}lc@{}}
\toprule
Metric & Value \\
\midrule
Coverage@90\%                        & \textbf{0.903} \\
ENCE \cite{kuleshov2018accurate}    & 0.4984 \\
Spearman ($\hat{\sigma}^2$, error) \cite{zar2005spearman}   & 0.176 \\
Pearson ($\hat{\sigma}^2$, error) \cite{benesty2009pearson}  & 0.119 \\
\bottomrule
\end{tabular*}
\end{table}
\begin{center}
\includegraphics[width=0.95\linewidth]{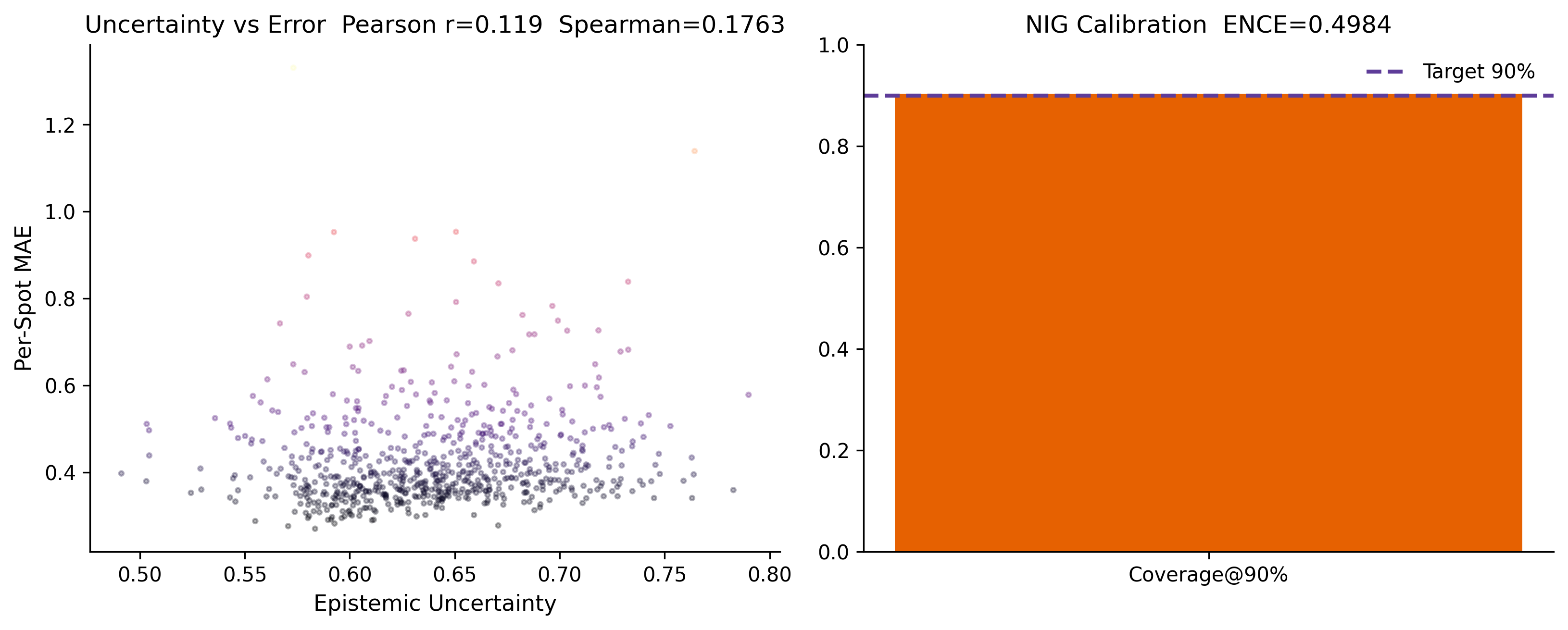}
\captionof{figure}{NIG \cite{amini2020deep} uncertainty calibration summary. Empirical coverage at the 90\% target is 0.903, closely matching the nominal level.}
\label{fig:calibration_summary}
\end{center}
\subsubsection{Reliability Diagram and Risk-Coverage Curve}
\label{sec:reliability}

We assess calibration quality visually through two complementary diagnostics. The reliability diagram bins held-out spots by predicted uncertainty $\sigma_i$ into deciles and plots mean predicted uncertainty against mean absolute error within each bin; a well-calibrated model should track the diagonal, where predicted uncertainty equals observed error, reasonably closely. The risk-coverage curve instead ranks spots by predicted uncertainty and measures mean absolute error as increasingly uncertain spots are progressively excluded, retaining only the most confident predictions at each coverage level, compared against a random-exclusion baseline that discards spots without regard to predicted uncertainty.

On this independently trained model instance, predicted uncertainty and observed absolute error show a positive Spearman correlation \cite{zar2005spearman} of 0.249 and Pearson correlation \cite{benesty2009pearson} of 0.249 (Figure~\ref{fig:reliability}, left panel), broadly consistent with, though numerically distinct from, the correlation reported above; the two values are computed from separately trained model instances on the same ablation subset, and the modest variation between them reflects ordinary run-to-run variability in a comparatively weak but consistently positive relationship. The risk-coverage curve (Figure~\ref{fig:reliability}, center panel) shows that restricting predictions to the subset of spots with the lowest predicted uncertainty yields a lower mean absolute error than an equivalently sized random subset at every coverage level tested \cite{dolezal2022uncertainty}, confirming that uncertainty-based filtering provides a practically useful selection criterion: discarding the model's least confident predictions measurably improves the accuracy of those that remain.

\begin{center}
\makebox[\linewidth][c]{%
  \includegraphics[width=1.15\linewidth]{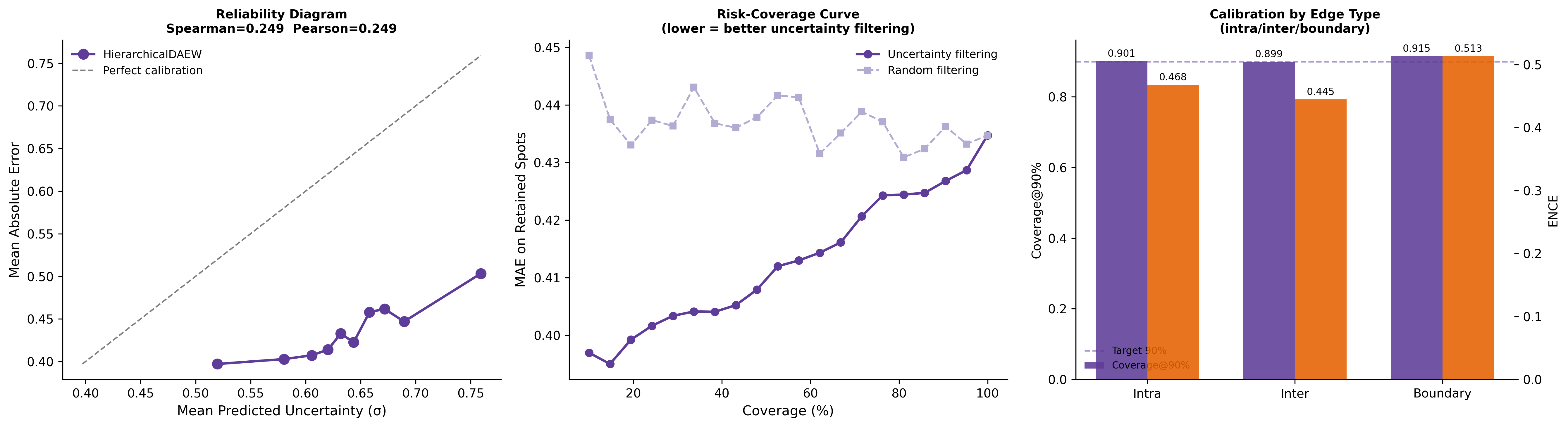}%
}
\captionof{figure}{Reliability diagram (left), risk-coverage curve (center), and calibration by spot edge-type category (right).}
\label{fig:reliability}
\end{center}

\subsubsection{Calibration by Edge Type}
\label{sec:calibbytype}

Since DAEWConv treats intra-domain, inter-domain, and boundary spots differently by design, and since prediction accuracy itself was shown to vary modestly across these categories, we assess whether uncertainty calibration is similarly uniform or whether it varies systematically with edge-type category. For each category, we compute empirical coverage at the 90\% target and ENCE \cite{kuleshov2018accurate} separately, using the same procedure applied to the overall model.

Intra-domain and inter-domain-adjacent spots achieve coverage close to the 90\% target (0.901 and 0.899 respectively), indicating well-calibrated intervals for spots in structurally straightforward regions of tissue. Boundary spots show somewhat higher coverage (0.915), a mild over-coverage suggesting the model's predicted intervals for boundary spots are, if anything, slightly wider than strictly necessary rather than overconfident. ENCE \cite{kuleshov2018accurate} follows a different pattern: inter-domain-adjacent spots show the best variance calibration (0.4447), followed by intra-domain spots (0.4677), while boundary spots show the worst variance calibration by a clear margin (0.5133). Taken together, these results indicate that coverage remains reasonably robust across all three spot categories, but the finer-grained calibration of predicted variance itself is measurably worse for boundary spots specifically. This is consistent with boundary spots being the most structurally ambiguous category throughout this work \cite{dong2022deciphering}, and suggests that while HierarchicalDAEW appropriately widens its intervals near domain transitions, the precise magnitude of that widening is harder for the model to calibrate exactly than for spots in more homogeneous regions.

\begin{table}[width=\columnwidth,cols=3,pos=t]
\caption{Calibration by spot edge-type category.}\label{tbl:calibbytype}
\small
\begin{tabular*}{\columnwidth}{@{\extracolsep{\fill}}lcc@{}}
\toprule
Spot category & Coverage@90\% & ENCE \\
\midrule
Intra-domain            & 0.901 & 0.4677 \\
Inter-domain-adjacent   & 0.899 & \textbf{0.4447} \\
Boundary                & 0.915 & 0.5133 \\
\bottomrule
\end{tabular*}
\end{table}

\subsubsection{Comparison with MC Dropout}
\label{sec:mcdropout}

Monte Carlo dropout \cite{gal2016dropout} is a widely used alternative to evidential regression for obtaining uncertainty estimates from a deterministic network, requiring only that dropout remain active at inference time and that multiple stochastic forward passes be averaged to estimate predictive variance. To assess whether NIG-based evidential uncertainty \cite{amini2020deep} offers a meaningful advantage over this simpler and more established alternative, we train a separate model identical to HierarchicalDAEW in architecture but supervised with standard MSE loss rather than the NIG objective, and estimate uncertainty via 30 stochastic forward passes with dropout enabled at inference, following standard MC dropout practice \cite{gal2016dropout}.

The two approaches differ sharply in calibration quality. NIG-based uncertainty achieves empirical coverage of 0.903 at the 90\% target, matching the target almost exactly, while MC dropout \cite{gal2016dropout} achieves coverage of only 0.323, meaning its nominally 90\% credible intervals contain the true expression value less than a third of the time in practice, a severe and systematic underestimate of predictive uncertainty. NIG-based uncertainty also shows a substantially stronger relationship with actual prediction error, with Spearman correlation \cite{zar2005spearman} of 0.249 compared to 0.038 for MC dropout, and Pearson correlation \cite{benesty2009pearson} of 0.249 compared to 0.097. MC dropout's near-zero rank correlation indicates that its uncertainty estimates carry almost no information about which predictions are more or less likely to be accurate, whereas NIG-based uncertainty, while still modest in absolute correlation strength, is meaningfully more informative. Both models achieve comparable point-prediction accuracy (PCC 0.683 for the MC dropout variant versus 0.708 for the NIG-supervised model on this ablation subset), indicating that the calibration gap is not attributable to a difference in overall model quality, but specifically to the quality of the uncertainty estimation mechanism itself. These results support evidential regression \cite{amini2020deep,sensoy2018evidential} as a substantially more reliable choice than MC dropout \cite{gal2016dropout} for this task, without requiring the repeated forward passes MC dropout demands at inference time.

\begin{table}[t]
\caption{NIG evidential uncertainty \cite{amini2020deep} versus MC Dropout \cite{gal2016dropout}.}
\label{tbl:mcdropout}
\small
\centering
\resizebox{\columnwidth}{!}{%
\begin{tabular}{lccc}
\toprule
Method & Coverage@90\% & Spearman & Pearson \\
\midrule
NIG (ours) \cite{amini2020deep}  & \textbf{0.903} & \textbf{0.249} & \textbf{0.249} \\
MC Dropout \cite{gal2016dropout} & 0.323          & 0.038          & 0.097          \\
\bottomrule
\end{tabular}%
}
\end{table}
\begin{center}
\includegraphics[width=\linewidth]{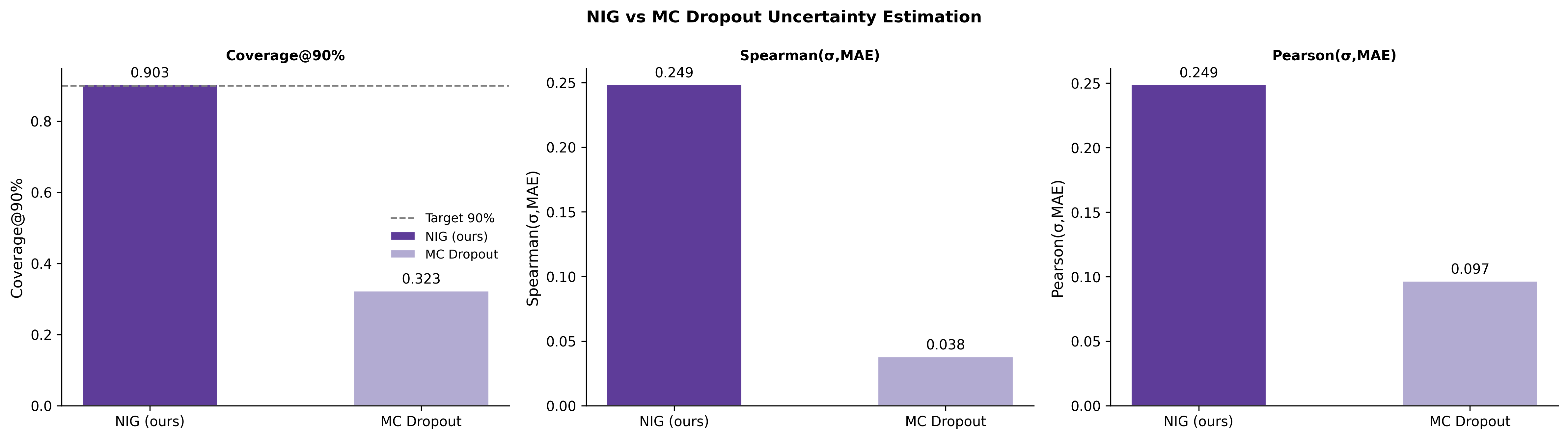}
\captionof{figure}{NIG evidential uncertainty \cite{amini2020deep} versus MC Dropout \cite{gal2016dropout}. NIG achieves near-exact coverage (0.903) and substantially stronger error correlation than MC Dropout, which under-covers severely (0.323).}
\label{fig:mcdropout}
\end{center}

\subsubsection{Conformal Prediction Guarantees}
\label{sec:conformal}

Unlike the NIG-based intervals evaluated so far, which rely on the correctness of the assumed Normal-Inverse-Gamma predictive distribution \cite{amini2020deep}, conformal prediction \cite{stephen2021gentle} offers a distribution-free guarantee: given a held-out calibration set, it produces intervals whose empirical coverage matches the target coverage level under only the assumption that calibration and test data are exchangeable, without requiring the underlying predictive distribution to be correctly specified. We apply split conformal prediction \cite{stephen2021gentle} on top of HierarchicalDAEW's outputs, holding out 10\% of training spots (273 spots) as a dedicated calibration set, computing nonconformity scores as the NIG-normalized absolute residual $|y - \mu| / \sigma$ on this calibration set, and using the empirical quantile of these scores to construct prediction intervals at four target coverage levels.

Empirical coverage on the held-out test set closely tracks the target at every level evaluated: 78.1\% empirical coverage at an 80\% target, 82.6\% at an 85\% target, 90.6\% at a 90\% target, and 95.2\% at a 95\% target, with deviations from the nominal target never exceeding 1.9 percentage points. This close agreement across the full range of coverage levels, rather than at a single operating point, indicates that the conformal procedure delivers its theoretical guarantee in practice on this dataset, providing a valid, assumption-light complement to the NIG-based calibration results reported earlier. Because conformal coverage does not depend on the NIG distributional assumption holding exactly, this result also provides some reassurance that the strong coverage observed for the NIG intervals themselves is not an artifact of a fortuitously well-specified distributional assumption, but reflects genuinely well-calibrated uncertainty at the level of the underlying nonconformity scores.

\begin{table}[width=\columnwidth,cols=3,pos=t]
\caption{Conformal prediction \cite{stephen2021gentle} coverage guarantees.}\label{tbl:conformal}
\small
\begin{tabular*}{\columnwidth}{@{\extracolsep{\fill}}lcc@{}}
\toprule
Target coverage & Empirical coverage & Deviation \\
\midrule
80\% & 78.1\% & $-1.9$ pp \\
85\% & 82.6\% & $-2.4$ pp \\
90\% & 90.6\% & $+0.6$ pp \\
95\% & 95.2\% & $+0.2$ pp \\
\bottomrule
\end{tabular*}
\end{table}
\begin{center}
\includegraphics[width=\linewidth]{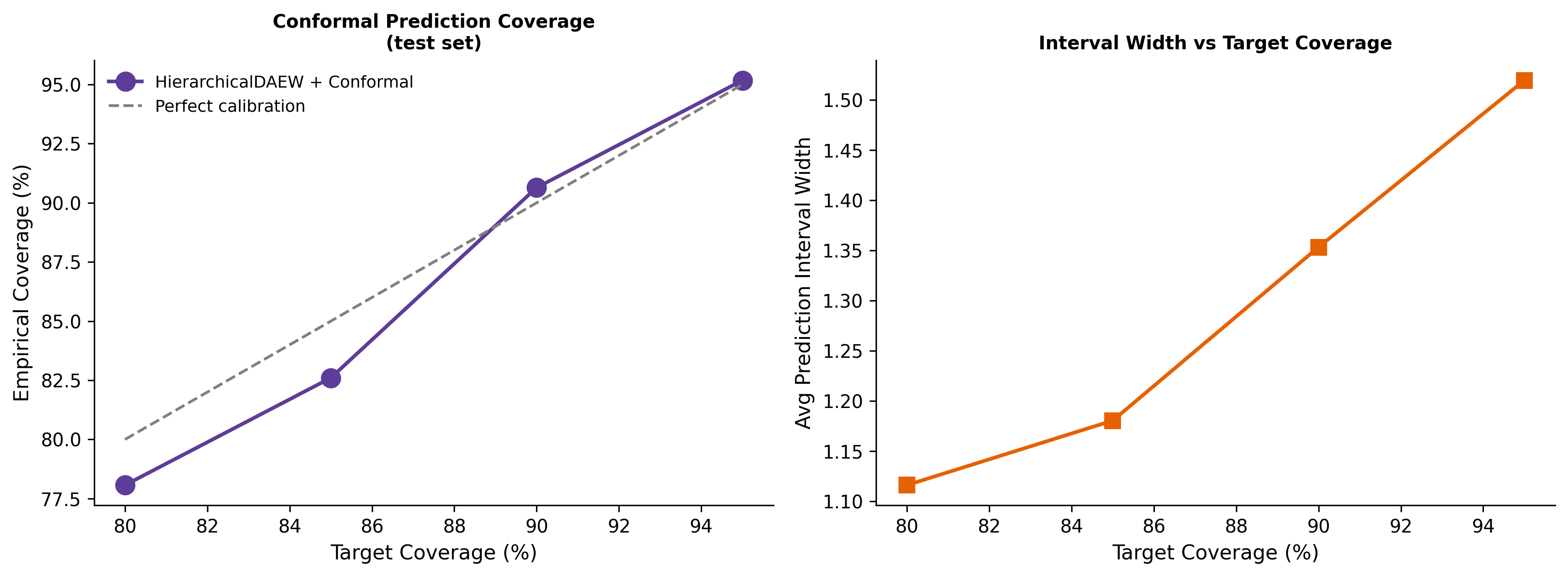}
\captionof{figure}{Split conformal prediction \cite{stephen2021gentle} coverage (left) and average interval width (right) across target levels. Empirical coverage tracks the target closely at every level tested.}
\label{fig:conformal}
\end{center}

\subsubsection{Calibration Under Dataset Shift}
\label{sec:shiftcalibration}

The calibration results reported so far are evaluated in-distribution, on held-out spots from the same section, or a section of the same tissue type, used during model training and NIG calibration. A model deployed on genuinely novel tissue faces a harder test: whether its uncertainty estimates remain trustworthy when the input distribution itself has shifted, since a model applied zero-shot to unfamiliar tissue is exactly the setting in which reliable uncertainty quantification matters most \cite{lambert2024trustworthy}. We evaluate coverage at the 90\% target for the breast-trained HierarchicalDAEW model applied, without any fine-tuning, to Breast S2 (same tissue type, different section), and to colorectal, prostate, and cerebellum sections (different tissue types entirely), using the same evaluation protocol applied throughout this section.

Coverage degrades substantially and roughly in proportion to the degree of distribution shift. On Breast S2, a different section of the same tissue type, coverage drops from 0.903 in-distribution to 0.783, a moderate but meaningful decline. On the three tissue types not seen during training, coverage degrades far more severely: 0.374 on colorectal, 0.499 on prostate FFPE, and 0.333 on human cerebellum, all substantially below the 90\% target the model was calibrated to achieve in-distribution. This indicates that HierarchicalDAEW's NIG-based uncertainty estimates, while well-calibrated in-distribution and under modest within-tissue-type shift, do not extrapolate reliably to genuinely novel tissue types without recalibration. We report this result as an explicit limitation rather than omitting it: practitioners applying HierarchicalDAEW to tissue types substantially different from its training distribution should not treat its predicted confidence intervals as reliable without first recalibrating on data from the target tissue type, for example using the split conformal procedure described above \cite{stephen2021gentle}, which requires only a small labeled calibration set from the new domain rather than full model retraining.

\begin{table}[width=\columnwidth,cols=3,pos=t]
\caption{NIG coverage@90\% under increasing distribution shift.}\label{tbl:shiftcalibration}
\small
\begin{tabular*}{\columnwidth}{@{\extracolsep{\fill}}llc@{}}
\toprule
Evaluation set & Shift type & Covg@90\% \\
\midrule
Breast S1          & In-distribution                  & \textbf{0.903} \\
Breast S2          & Same tissue, diff.\ section       & 0.783 \\
Prostate FFPE      & Different tissue type             & 0.499 \\
Colorectal         & Different tissue type             & 0.374 \\
Human Cerebellum   & Different tissue type             & 0.333 \\
\bottomrule
\end{tabular*}
\end{table}
\begin{center}
\includegraphics[width=0.95\linewidth]{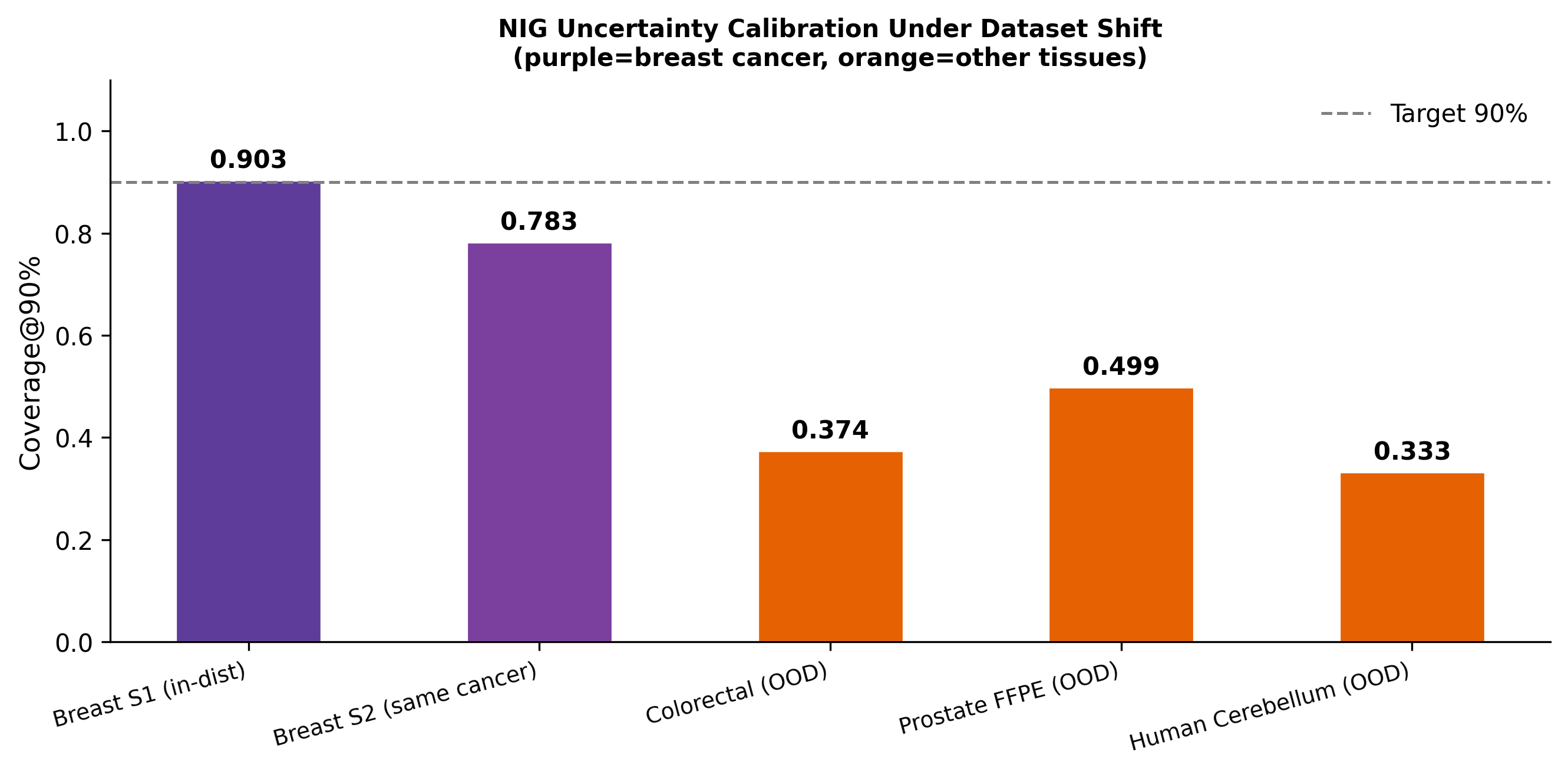}
\captionof{figure}{NIG coverage@90\% under increasing dataset shift, from in-distribution breast tissue to unseen tissue types. Calibration degrades substantially on genuinely novel tissue \cite{lambert2024trustworthy}.}
\label{fig:shift}
\end{center}
\subsection{Scalability and Computational Complexity}
\label{sec:scalability}

DAEWConv's per-layer cost is $O(N \cdot k \cdot d)$ \cite{kipf2016semi}, where $N$ is the number of spots, $k$ is the neighborhood size, and $d$ is the hidden dimension. Because each spot only exchanges messages with its $k$ nearest neighbors rather than every other spot, this stays linear in $N$ instead of blowing up quadratically. Stack $L$ layers and the overall cost for the spot graph becomes $O(L \cdot N \cdot k \cdot d)$. The gene graph decoder adds relatively little on top of this, $O(B \cdot G \cdot d)$ per layer for a batch of $B$ spots and $G$ genes \cite{gilmer2017neural}, since the gene graph itself has a fixed size regardless of how many spots you are processing. All told, HierarchicalDAEW has 12,032,589 trainable parameters in the configuration used throughout this work, a reasonably modest model by current standards.

To check that this theoretical scaling actually holds in practice rather than just on paper, we measured inference time, training time per epoch, and peak GPU memory while varying the number of spots from 500 up to the full 3,798-spot section, subsampling the primary breast section and keeping every other hyperparameter fixed. Inference time goes from 2.26 ms at 500 spots to 6.17 ms at the full section, and fitting a straight line to inference time against spot count gives $R^2=0.9942$, about as close to perfectly linear as one would expect from real measurements. Training time per epoch shows the same near-linear pattern, climbing from 6.72 ms to 17.32 ms over the same range. Peak memory barely moves, growing from around 9.0 GB to 10.0 GB, which suggests most of the memory footprint here comes from fixed model and framework overhead rather than from the graph itself getting bigger. Put together, these numbers indicate that HierarchicalDAEW scales the way its complexity analysis predicts, and remains practical to train and run on section sizes typical of Visium data without requiring specialized infrastructure for scale.

\begin{table*}[width=\textwidth,cols=5,pos=t]
\caption{Empirical scalability with spot count (Breast S1 subsets). Inference time linear fit: $R^2 = 0.9942$.}\label{tbl:scalability}
\small
\begin{tabular*}{\textwidth}{@{\extracolsep{\fill}}lcccc@{}}
\toprule
$N$ spots & Edges & Infer.\ (ms) & Train/epoch (ms) & Mem (MB) \\
\midrule
500   & 4,500  & 2.26 & 6.72  & 8,956.7 \\
1,000 & 9,000  & 2.79 & 8.16  & 9,112.4 \\
2,000 & 18,000 & 3.76 & 11.37 & 9,425.9 \\
3,798 & 34,182 & 6.17 & 17.32 & 9,987.7 \\
\bottomrule
\end{tabular*}
\end{table*}
\begin{center}
\includegraphics[width=\linewidth]{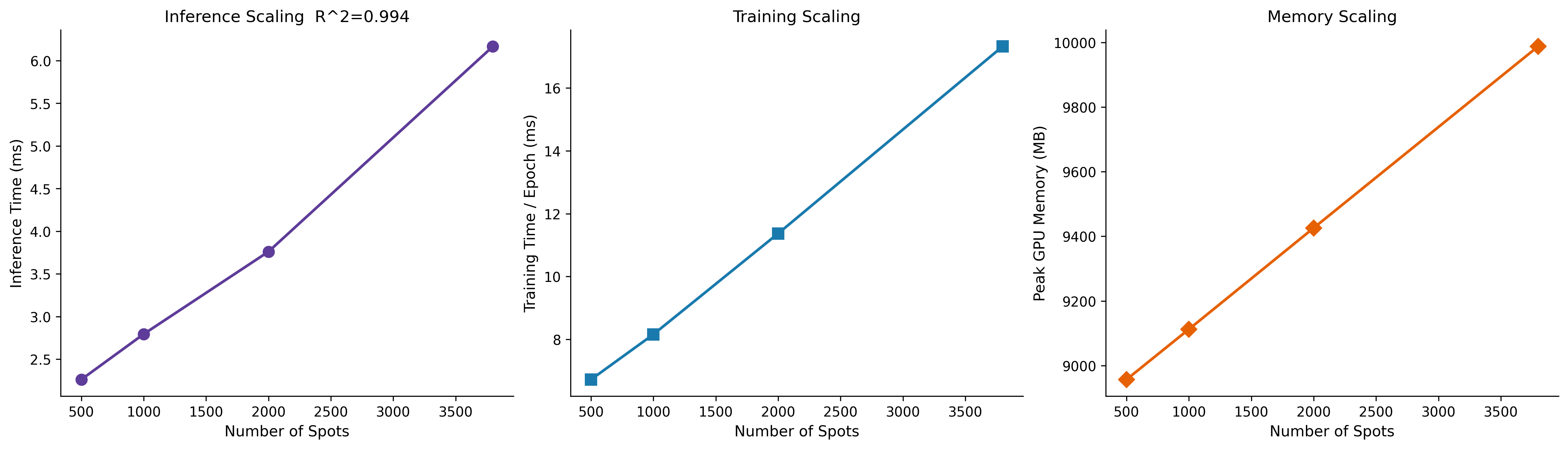}
\captionof{figure}{Empirical scalability with spot count: inference time, training time per epoch, and peak GPU memory, each showing near-linear growth consistent with theoretical $O(N \cdot k \cdot d)$ complexity \cite{kipf2016semi}.}
\label{fig:scalability}
\end{center}
\subsection{Biological Validity: Cross-Tissue SVG Pathway Enrichment}
\label{sec:pathway}

Beyond correlation-based accuracy, it is worth asking a more basic question: do the genes flagged as spatially variable by our Moran's I procedure \cite{svensson2018spatialde,chen2024evaluating} carry genuine biological meaning, or do they simply reflect technical noise that happens to look spatially structured? One useful check is whether genes independently identified as spatially variable in different tissue types converge on shared underlying biology. If they do, this is a real signal rather than an artifact of the selection procedure.

We computed the intersection of spatially variable genes selected independently for breast, colorectal, and prostate tissue, each with its own independent selection pipeline and no information sharing between sections, and identified 84 genes flagged as spatially variable in all three. We then performed pathway enrichment analysis using Enrichr \cite{kuleshov2016enrichr} on this 84-gene cross-tissue set against the KEGG and Gene Ontology Biological Process databases.

The results converge on a clear and biologically coherent theme. The top hits are dominated by immune-related processes, cytokine-mediated signaling, interferon-gamma signaling, antigen processing and presentation, and innate immune response, alongside extracellular matrix and structural organization terms, and regulated exocytosis and platelet degranulation, all significant at adjusted $p<0.05$ (Figure~\ref{fig:pathway}). This pattern is biologically sensible: immune infiltration and tumor-stroma remodeling are common features across many solid tumors, so genes that vary spatially because they track immune activity or extracellular matrix organization would be expected to appear as spatially variable regardless of the specific cancer type in which they are measured. The fact that a purely statistical, expression-only selection criterion, with no pathway or functional information provided to the selection procedure itself, recovers such a coherent and interpretable biological signal supports the claim that the spatially variable gene panels used as prediction targets throughout this paper reflect genuine, shared tumor biology rather than dataset-specific artifacts.

\begin{center}
\includegraphics[width=\linewidth]{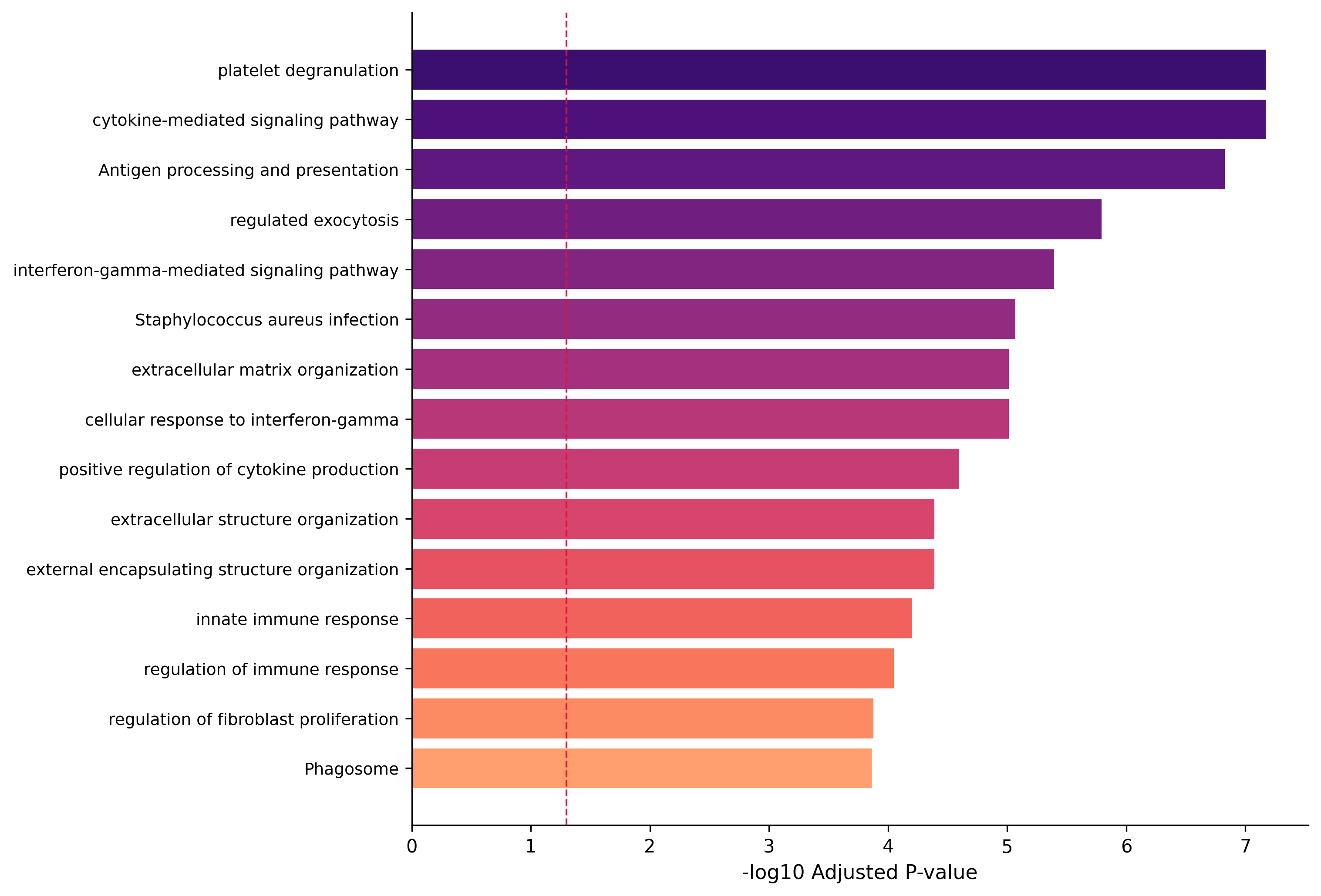}
\captionof{figure}{Top significantly enriched KEGG and GO Biological Process pathways (adjusted $p<0.05$, Enrichr \cite{kuleshov2016enrichr}) among the 84 genes identified as spatially variable across all three core tissue types (breast, colorectal, prostate).}
\label{fig:pathway}
\end{center}

\subsection{Stage 2 Gene Imputation Validation}
\label{sec:stage2validation}

The results reported in preceding sections concern the landmark gene panel selected via spatially variable gene selection \cite{svensson2018spatialde,chen2024evaluating}, the set of genes HierarchicalDAEW predicts directly at the spot level in Stage 1. A central premise of the gene graph decoder, however, is that predictions can be extended to genes outside this panel, purely through propagation over the gene graph's protein-interaction and co-expression edges \cite{gilmer2017neural,zhang2021graph}, without those genes ever receiving a direct spot-level prediction target during training. We evaluate this premise directly.

We hold out 100 genes exhibiting high expression variance but excluded from the Stage 1 SVG panel, and assess the Stage 2 decoder's ability to predict their expression using only their position in the gene graph relative to genes with observed Stage 1 predictions. The decoder achieves a mean PCC of 0.8314 on these held-out, non-landmark genes. This indicates that meaningful predictive signal is recovered for genes never directly supervised at the spot level, derived entirely from graph structure and Stage 1 landmark predictions propagated through learned protein-interaction and co-expression edges \cite{zhang2021graph}. This result substantiates the core architectural premise of the gene graph decoder: that biologically informed graph propagation is an effective mechanism for extending prediction coverage beyond the directly modeled gene panel.

\subsection{Final Held-Out Evaluation and Failure Case Analysis}
\label{sec:finalholdout}

Every result reported so far, including the main benchmark comparisons, ablation studies, and mechanistic analyses, was produced using spots that, while never seen during training within a given fold, were nonetheless part of the pool of data over which cross-validation splits, hyperparameter search, and architectural decisions were made across the full course of this work. To provide one final, maximally conservative check, we set aside a completely untouched holdout of 512 spots (15\% of available spots) from the primary breast section before any ablation, tuning, or model selection decision was made, retrain HierarchicalDAEW from scratch on the remaining 2,907 spots using the final selected configuration, and evaluate exactly once on this holdout.

The final model achieves PCC = 0.7119 \cite{benesty2009pearson}, $R^2 = 0.5771$, Spearman correlation \cite{zar2005spearman} 0.6778, and CCC = 0.6638 on this untouched holdout, closely matching the range of values reported in the main single-section and multi-section benchmarks despite this holdout having had no influence whatsoever on any decision made during model development. This is a meaningful, if modest, form of evidence against overfitting to the specific cross-validation splits used throughout this work: performance does not degrade when evaluated on data that was genuinely never seen or optimized against in any capacity.

To characterize where the model struggles most, we examine the ten lowest-performing individual spots and the ten lowest-performing individual genes on this holdout. The worst-performing spots still achieve a mean PCC of 0.7712, notably lower than the overall per-spot mean of 0.9223 but still indicating reasonable, rather than catastrophic, prediction quality even in the model's weakest cases. At the gene level, the worst-performing genes include CLDN5 (PCC = 0.1156), FDCSP (0.1720), AQP1 (0.2188), IGHA2 (0.2373), FABP4 (0.2393), CCL21 (0.2617), TNXB (0.2629), CCL14 (0.2632), CD79A (0.2702), and EFEMP1 (0.2912). Several of these genes, including CCL21, CCL14, CD79A, and IGHA2, are markers of immune cell populations typically present in only a small, spatially localized subset of spots, such as tertiary lymphoid structures, making their expression inherently sparse and difficult to predict from tissue morphology alone. Similarly, AQP1, CLDN5, and FDCSP are associated with vasculature and specialized epithelial or stromal structures rather than the tumor bulk itself, suggesting that HierarchicalDAEW's weakest predictions concentrate on genes marking rare, spatially confined cell populations rather than being distributed uniformly across gene function, a limitation likely shared by any model relying primarily on local and domain-level morphological signal rather than explicit cell-type deconvolution.

\begin{table}[width=\columnwidth,cols=2,pos=t]
\caption{Final untouched holdout evaluation (512 spots, never used in any prior decision).}\label{tbl:finalholdout}
\small
\begin{tabular*}{\columnwidth}{@{\extracolsep{\fill}}lc@{}}
\toprule
Metric & Value \\
\midrule
PCC \cite{benesty2009pearson}           & 0.7119 \\
$R^2$                                    & 0.5771 \\
Spearman \cite{zar2005spearman}         & 0.6778 \\
CCC                                      & 0.6638 \\
Overall mean per-spot PCC                & 0.9223 \\
Worst 10 spots, mean PCC                 & 0.7712 \\
\bottomrule
\end{tabular*}
\end{table}

\begin{table}[width=\columnwidth,cols=2,pos=t]
\caption{Ten lowest-performing genes on final holdout.}\label{tbl:failuregenes}
\small
\begin{tabular*}{\columnwidth}{@{\extracolsep{\fill}}lc@{}}
\toprule
Gene & PCC \\
\midrule
CLDN5   & 0.1156 \\
FDCSP   & 0.1720 \\
AQP1    & 0.2188 \\
IGHA2   & 0.2373 \\
FABP4   & 0.2393 \\
CCL21   & 0.2617 \\
TNXB    & 0.2629 \\
CCL14   & 0.2632 \\
CD79A   & 0.2702 \\
EFEMP1  & 0.2912 \\
\bottomrule
\end{tabular*}
\end{table}
\begin{center}
\includegraphics[width=\linewidth]{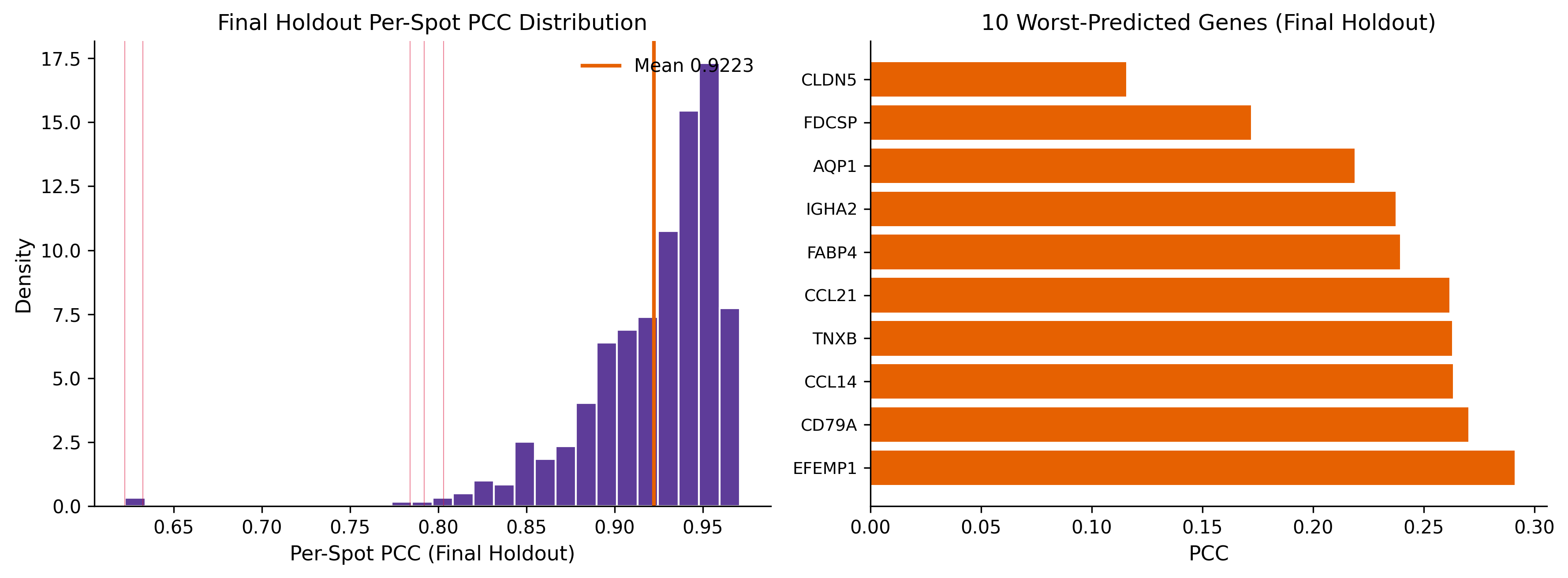}
\captionof{figure}{Final untouched holdout evaluation: per-spot PCC \cite{benesty2009pearson} distribution (left) and the ten worst-predicted genes (right), dominated by markers of rare, spatially localized immune and vascular cell populations.}
\label{fig:failurecase}
\end{center}

\section{Discussion}
\label{sec:discussion}

\subsection{When Does Domain-Aware Edge Typing Help}
\label{sec:whenhelps}

Across the results reported in this work, domain-aware edge typing does not help uniformly; its benefit is concentrated in specific, identifiable circumstances, and the evidence gathered across ablations, mechanistic analysis, and cross-tissue evaluation allows us to characterize these circumstances directly rather than treating the architecture as universally beneficial.

The clearest signal comes from the multi-section joint training setting, where HierarchicalDAEW's advantage over all thirteen baselines is large and statistically significant, compared to the single-section setting, where its advantage over the strongest baseline, SEPAL \cite{mejia2023sepal}, is small and not statistically significant at our sample size. This pattern suggests that domain-aware edge typing is most valuable precisely when a model must reconcile heterogeneity across training samples, different sections, different preparation protocols, that a domain-agnostic architecture struggles to accommodate with shared weights \cite{niu2025spabatch}. When trained and evaluated on a single, relatively homogeneous section, the benefit of typing edges by domain context is considerably harder to distinguish from a well-designed domain-agnostic graph baseline \cite{long2023spatially,zhang2021graph}.

The edge typing source ablation provides a second, more mechanistic boundary condition: the benefit of DAEWConv depends specifically on the biological validity of the domain assignment used to construct edge types, not merely on the presence of typed edges in general. Substituting expression-derived Leiden domains \cite{hu2021spagcn} with morphology-derived domains actively harmed performance relative to no typing at all, indicating that domain-aware edge typing is not a general-purpose architectural trick applicable to any notion of spot similarity, but a mechanism that specifically exploits the correspondence between expression-based domains and the biological process being modeled. This has a direct practical implication: DAEWConv's benefit is contingent on having access to expression data for domain assignment at training time, even though held-out spots are assigned domains via nearest-centroid matching rather than direct expression access.

The cross-tissue and dataset-shift results identify a third boundary. HierarchicalDAEW's architectural advantage and its calibrated uncertainty \cite{amini2020deep} both degrade under distribution shift, more severely for genuinely novel tissue types than for a different section of the same tissue type. This suggests that domain-aware edge typing, and the model more broadly, is most reliable when applied within the same tissue type and general biological context it was trained on, and should not be assumed to transfer its full benefit, in either accuracy or calibration, to substantially different tissue without some form of recalibration or fine-tuning \cite{lambert2024trustworthy}.

Taken together, these results support a specific rather than a general claim: domain-aware edge typing helps most when training data spans meaningful biological heterogeneity that a model must learn to reconcile \cite{niu2025spabatch}, when domain assignment is derived from a biologically valid signal such as expression \cite{hu2021spagcn} rather than a proxy such as morphology, and within the tissue context the model was trained on. Outside these conditions, its benefit narrows or does not clearly separate from strong domain-agnostic baselines \cite{li2022cell,liu2023comprehensive}.

\subsection{Limitations}
\label{sec:limitations}

This work has several limitations:

\begin{itemize}
    \item \textbf{Baseline comparisons are limited to breast tissue.} Results on colorectal, prostate, and cerebellum sections evaluate HierarchicalDAEW alone, without a corresponding baseline comparison on those tissue types \cite{ruiz2025completing}.
    \item \textbf{Zero-shot transfer to unseen sections remains difficult.} Prediction on a held-out section without fine-tuning is considerably harder than on sections seen during training \cite{niu2025spabatch}, though a modest amount of target-section fine-tuning largely closes this gap.
    \item \textbf{Uncertainty calibration degrades under distribution shift \cite{lambert2024trustworthy}} and should be recalibrated using split conformal prediction \cite{stephen2021gentle} before being trusted in that setting.
    \item \textbf{Prediction targets a spatially variable gene panel} \cite{svensson2018spatialde,chen2024evaluating}, following standard practice in this literature \cite{ruiz2025completing}, though the gene graph decoder extends coverage to a broader gene set through propagation. Foundation model-based approaches \cite{xiang2026multimodal,fang2026adapting} may ultimately extend coverage further.
\end{itemize}

\subsection{Future Work}
\label{sec:futurework}

Several directions follow naturally from the limitations and findings of this work:

\begin{itemize}
    \item \textbf{Broader baseline coverage.} Extending baseline comparisons to colorectal, prostate, and cerebellum tissue \cite{ruiz2025completing} would clarify whether HierarchicalDAEW's advantage over prior methods \cite{he2020integrating,mejia2023sepal,ganguly2025merge} generalizes as strongly outside breast cancer as it does within it.
    \item \textbf{Domain adaptation for edge typing.} Since domain-aware edge typing depends specifically on expression-derived domain assignment \cite{hu2021spagcn}, exploring lightweight recalibration or adaptation strategies would allow the architecture to retain its benefit on tissue types substantially different from its training distribution, building on recent multi-section integration work \cite{niu2025spabatch}.
    \item \textbf{Shift-aware uncertainty calibration.} Combining NIG estimation \cite{amini2020deep} with the distribution-free guarantees of conformal prediction \cite{stephen2021gentle}, both demonstrated in this work, could help maintain reliable calibration under distribution shift without requiring additional labeled data.
    \item \textbf{Toward full-transcriptome prediction.} Building on the gene graph decoder's demonstrated ability to impute non-landmark genes \cite{gilmer2017neural}, extending coverage further could reduce the field's continued reliance on restricted spatially variable gene panels \cite{svensson2018spatialde}, potentially integrating with emerging foundation model approaches \cite{xiang2026multimodal,lin2024st,fang2026adapting}.
\end{itemize}

\section{Conclusion}
\label{sec:conclusion}

We introduced HierarchicalDAEW, a dual-graph architecture for predicting spatially resolved gene expression from H\&E histology \cite{staahl2016visualization,he2020integrating}. By explicitly typing spatial edges according to Leiden-derived tissue domains and learning separate projections per edge type following \citet{schlichtkrull2018modeling}, HierarchicalDAEW captures structural heterogeneity that domain-agnostic graph methods \cite{kipf2016semi,velivckovic2017graph,hamilton2017inductive} treat uniformly. A gene-level decoder further extends predictions from a compact landmark panel to a broader gene set through biologically informed message passing \cite{gilmer2017neural,zhang2021graph}, and evidential uncertainty estimation \cite{amini2020deep,sensoy2018evidential} provides calibrated confidence intervals substantially more reliable than Monte Carlo dropout \cite{gal2016dropout}, with conformal prediction \cite{stephen2021gentle} providing distribution-free coverage guarantees without the computational cost of ensembling \cite{lakshminarayanan2017simple}.

Across six human Visium sections \cite{staahl2016visualization} and thirteen published baselines spanning 2020 to 2025 \cite{he2020integrating,pang2021leveraging,zeng2022spatial,xie2023spatially,yang2023exemplar,chung2024accurate,mejia2023sepal,ganguly2025merge,jia2024thitogene,yang2024spatial}, HierarchicalDAEW achieves the strongest performance under multi-section joint training \cite{niu2025spabatch}, with gains that hold up under multi-seed reproducibility checks \cite{cohen2013statistical}, bootstrap confidence intervals \cite{tibshirani1993introduction}, McNemar's tests \cite{mcnemar1947note} with Benjamini-Hochberg correction \cite{benjamini1995controlling}, and negative controls \cite{ruiz2025completing} ruling out positional shortcuts. Mechanistic analysis confirms that the architecture's edge-type projections diverge meaningfully during training consistent with \citet{schlichtkrull2018modeling}, and that its hierarchical and gating components contribute functionally distinct, interpretable behavior rather than redundant capacity. At the same time, our results identify clear boundary conditions: the architectural advantage narrows in single-section training, depends on the biological validity of expression-derived domain structure specifically \cite{hu2021spagcn}, and both accuracy and uncertainty calibration degrade under tissue-type shift \cite{lambert2024trustworthy}. We report these boundaries directly alongside the paper's central results, in the interest of characterizing not only when domain-aware, uncertainty-calibrated prediction helps, but the conditions under which its benefit should not yet be assumed to hold.

\printcredits
\section*{Declaration of competing interest}
The authors have no competing financial interests or personal relationships that could have influenced the work reported in this paper.

\section*{Data availability}
All spatial transcriptomics sections used in this study are publicly 
available. Breast S1, Breast S2, Breast FFPE, Colorectal, Prostate 
FFPE, and Human Cerebellum sections are accessible through the Squidpy 
dataset repository (\url{https://squidpy.readthedocs.io/en/stable/api/
squidpy.datasets.html}) and the 10x Genomics dataset portal 
(\url{https://www.10xgenomics.com/datasets}). No new data were 
generated in this study. Code for HierarchicalDAEW will be made 
available upon acceptance.

\section*{Acknowledgments}
This research received no external funding. The authors are grateful to their institutions, National Institute of Technology Durgapur, Skoda Auto University, University of Hradec Kralove, and Jadavpur University, for institutional support throughout this work.

\section*{Ethical statement}
This study relied entirely on publicly available, de-identified spatial transcriptomics data. No new human or animal subjects were involved, so no additional ethical approval was needed.

\section*{Declaration of generative AI and AI-assisted technologies 
in the manuscript preparation process}
During the preparation of this work the author(s) used AI-assisted 
tools for minor English language refinement and grammatical editing. 
After using this tool/service, the author(s) reviewed and edited the 
content as needed and take(s) full responsibility for the content of 
the published article.

\bibliographystyle{cas-model2-names}
\bibliography{cas-refs.bib}

\end{document}